\documentclass[letterpaper]{article} %
\usepackage{aaai25}  %
\usepackage{times}  %
\usepackage{helvet}  %
\usepackage{courier}  %
\usepackage{multirow}
\usepackage{booktabs}
\usepackage[hyphens]{url}  %
\usepackage{graphicx} %
\usepackage{natbib}  %
\usepackage{caption} %
\nocopyright
\usepackage{algorithm}
\usepackage{algorithmic}

\usepackage{newfloat}
\usepackage{listings}
\DeclareCaptionStyle{ruled}{labelfont=normalfont,labelsep=colon,strut=off} %
\floatstyle{ruled}
\newfloat{listing}{tb}{lst}{}
\floatname{listing}{Listing}
\usepackage{graphicx}
\usepackage{xcolor}
\usepackage{multirow}
\usepackage{rotating}
\usepackage{multicol}
\usepackage{booktabs}
\usepackage{amsfonts}
\usepackage{amssymb}
\usepackage{pifont}
\usepackage{cleveref}
\Crefname{figure}{Figures}{Figures}

\usepackage{nicefrac}       %
\usepackage{xcolor}  
\usepackage{subfigure}
\usepackage{makecell}
\newcommand{\name}{StructFact}
\newcommand{\cmark}{\ding{51}}%

\title{Reasoning Factual Knowledge in Structured Data with Large Language Models}
\author{
    Sirui Huang\textsuperscript{\rm 1,\rm 2}\equalcontrib,
    Yanggan Gu\textsuperscript{\rm 3,\rm 4}\equalcontrib,
    Xuming Hu\textsuperscript{\rm 3}\thanks{Corresponding author: xuminghu@hkust-gz.edu.cn},
Zhonghao Li\textsuperscript{\rm 3},
    Qing Li\textsuperscript{\rm 1},
    Guandong Xu\textsuperscript{\rm 2,5}\thanks{Corresponding author: guandong.xu@uts.edu.au}
}
\affiliations{
    \textsuperscript{\rm 1}Hong Kong Polytechnic University, China\\
    \textsuperscript{\rm 2}University Technology Sydney, Australia\\
    \textsuperscript{\rm 3}The Hong Kong University of Science and Technology (Guangzhou), China\\
    \textsuperscript{\rm 4}Soochow University, China\\
    \textsuperscript{\rm 5}The Education University of Hong Kong, China\\
}

\usepackage{bibentry}

\begin{document}

\maketitle

\begin{abstract}
Large language models (LLMs) have made remarkable progress in various natural language processing tasks as a benefit of their capability to comprehend and reason with factual knowledge. However, a significant amount of factual knowledge is stored in structured data, which possesses unique characteristics that differ from the unstructured texts used for pretraining. This difference can introduce imperceptible inference parameter deviations, posing challenges for LLMs in effectively utilizing and reasoning with structured data to accurately infer factual knowledge. To this end, we propose a benchmark named \name\, to evaluate the structural reasoning capabilities of LLMs in inferring factual knowledge. \name\ comprises 8,340 factual questions encompassing various tasks, domains, timelines, and regions. This benchmark allows us to investigate the capability of LLMs across five factual tasks derived from the unique characteristics of structural facts. Extensive experiments on a set of LLMs with different training strategies reveal the limitations of current LLMs in inferring factual knowledge from structured data. We present this benchmark as a compass to navigate the strengths and weaknesses of LLMs in reasoning with structured data for knowledge-sensitive tasks, and to encourage advancements in related real-world applications. Please find our code at https://github.com/EganGu/StructFact.

\end{abstract}

\section{Introduction}
Large Language Models (LLMs) have revolutionized various downstream natural language processing (NLP) tasks with their impressive capabilities to comprehend and reason on textual data. Previous studies have demonstrated that factual knowledge can be stored within LLMs as a knowledge base, serving knowledge-sensitive tasks such as fact-checking and question-answering~\shortcite{kojima2022large, tirumala2022memorization, hu2023large}. Compared to the traditional method of retrieving knowledge from knowledge bases, reasoning factual knowledge with LLMs can introduce difficult-to-correct errors due to deviations in inference parameters~\shortcite{tablemeetsllm}. Additionally, LLMs are pretrained on serialized data, overlooking the structural nature of factual knowledge storage (e.g., tables, lists)~\shortcite{hu2023large,cui2024tabular}. Therefore, effectively using structured data to infer factual knowledge with LLMs remains challenging.

Compared to unstructured data, certain unique characteristics of structured data affect the ability of LLMs to understand and reason about factual knowledge~\shortcite{fang2024large}. These characteristics include: (1) \textit{Heterogeneity}. structured data consists of diverse types (e.g., texts, numerics, dates). Misunderstandings or biases of any type can lead to inaccuracies in the facts. (2) \textit{Topological Interdependencies}. Most LLMs are based on the Transformer architecture~\shortcite{vaswani2017attention} and are trained with a next-word prediction loss objective, primarily designed to process continuous text data. Extracting relevant interdependencies from complex topological structures is a significant challenge for LLMs in understanding and reasoning about facts. (3) \textit{Order Invariance}. A key assumption in pretraining is that the order of words significantly impacts their semantics~\shortcite{chen2024hytrel}. However, in structured data, the permutation of entities (e.g., rows in tables) does not alter the underlying factual knowledge. (4) \textit{Sparsity}. To maintain the same performance in sparse structured data (e.g., missing values or incomplete descriptions) as in data-rich scenarios, LLMs need to accurately utilize the general knowledge learned during pretraining and avoid non-factual imputations. (5) \textit{Lack of Prior Knowledge}. Structured data holds domain-specific knowledge not exposed during pretraining, challenging the accurate application of general reasoning to downstream tasks without distortion~\shortcite{colon2021combining,zhao2023investigating,li2024sheetcopilot}. These characteristics of structured data impact the ability of LLMs to reason about factual issues, limiting their real-world applications, especially in high-risk domains such as healthcare and finance. To enable LLMs to effectively utilize knowledge embedded in structured data and enhance reliable reasoning, it is essential to examine their capabilities based on the specific characteristics of structured data.

In light of these characteristics, we analyze the reasoning capabilities of LLMs on structured data from the perspective of five factual tasks: Arithmetic Calculation, Spatiotemporal Cognition, Multi-hop Reasoning, Composition Understanding, Combining Structured and Unstructured. We develop \name, a benchmark comprising 8,340 questions that involve questions and corresponding structured evidence in various data types, knowledge domains, timeliness, and regions. Additionally, we categorized these questions into five factual tasks and provided fine-grained difficulty annotations based on the specific focus of each task to facilitate a multifaceted analysis of the reasoning capabilities of LLMs. 

\begin{figure*}[!t]
  \centering
  \includegraphics[scale=0.85]{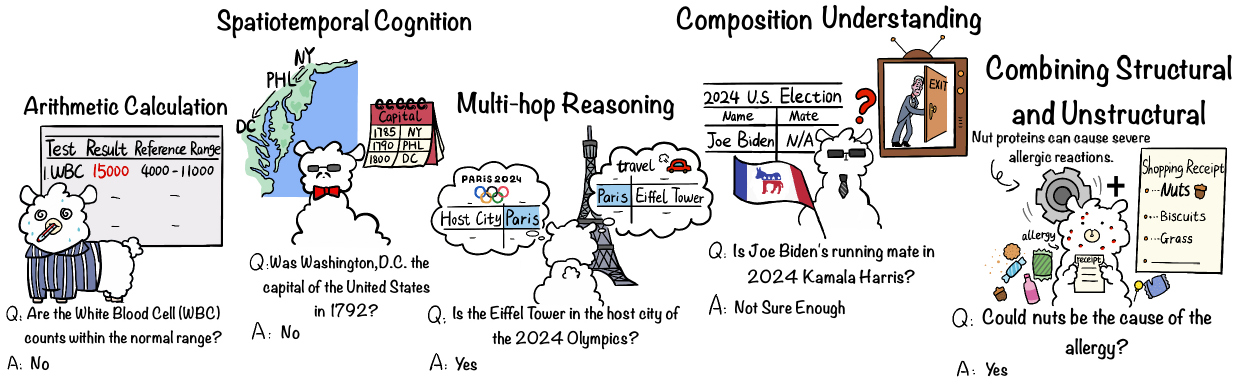}
  \vspace{-8mm}
  \caption{\name\ aims at evaluating the reasoning ability of LLMs over structured factual knowledge across five tasks.}
  \label{fig:structfacts}
  \vspace{-5mm}
\end{figure*}

Through explorations with the \name\ benchmark, we explore whether a range of large language models can understand and reason on the factual knowledge stored in structured data. For instance, in tasks that rely on \textit{heterogenous} data, LLMs heavily depend on the order of information to achieve understanding; in the situation where data is \textit{sparse}, LLMs face challenges in accurately utilizing their knowledge base to comprehend and reason over complex structure. We aim for \name\ to serve as a guiding tool in exploring the boundaries of LLMs in knowledge-sensitive tasks involving structural facts, while also advancing their practical applications in real-world scenarios.

\begin{figure*}[!t]
  \centering
  \includegraphics[scale=0.43]{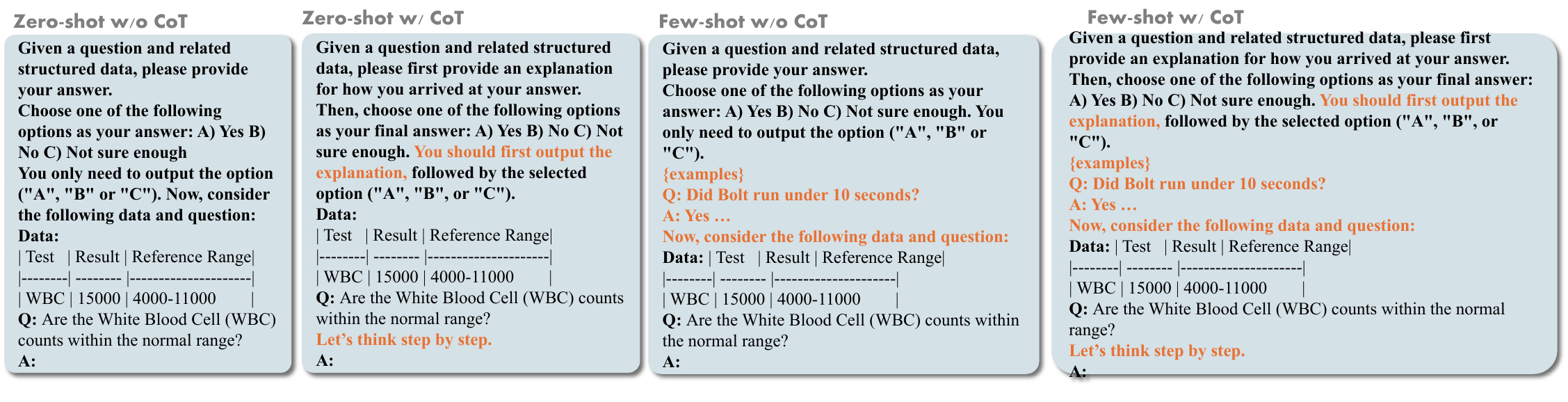}
  \vspace{-4mm}
  \caption{Prompts used in different settings (the main differences between each with zero-shot w/o CoT are marked in orange).}
  \label{fig:prompts_main_res}
\end{figure*}

\section{Data Construction}
To thoroughly assess the ability of LLMs to reason factual knowledge from structural data, we have designed five tasks aimed at investigating different research questions and meticulously selected questions and corresponding structured evidence from diverse sources. We compare our \name\ with other public datasets containing structured knowledge in Appendix \ref{sec:apdx_structural_datasets}.

\subsection{Tasks}
\begin{itemize}
    \item \textbf{Arithmetic Calculation}. Given the substantial amount of numerical facts stored in structural data, such as the health report in Figure \ref{fig:structfacts}, LLMs are required to perform arithmetic calculations when reasoning over such \textit{heterogeneous} data~\shortcite{ibmstructured,amzstructured}. %
    Models such as Graph Neural Networks (GNNs) seamlessly handle arithmetic calculations by inferring arithmetic rules from numerical patterns through their structural architecture, while LLMs are based on the transformer architecture which is designed for unstructured data. This presents the question of \textit{whether LLMs are capable of capturing and memorizing arithmetic rules}. To answer this question, our \name\ benchmark includes factual questions from simple numerical matching to difficult computational analysis. 
    \item \textbf{Spatiotemporal Cognition}. Spatiotemporal information in factual knowledge can be presented in diverse formats. Temporal data span dates and time periods, while spatial data include geographic coordinates (i.e., latitude and longitude), city names, and country names. This \textit{heterogeneity} of structured spatiotemporal data poses a challenge for LLMs, which are required to precisely interpret and align these diverse formats. For example, to answer the query about the capital of the U.S. in 1792, as shown in Figure \ref{fig:structfacts}, LLMs have to reason about the specific periods during which each city served as the capital. Consequently, a crucial research question arises: \textit{Can LLMs cognize spatiotemporal knowledge for factual reasoning?} To assess the capabilities of LLMs in spatiotemporal cognition, we incorporate factual knowledge related to temporal, spatial, and spatiotemporal entities.
    \item \textbf{Multi-hop Reasoning}. Factual knowledge in structural data involves entities dispersed across multiple sources~\shortcite{li2023chain}. In Figure \ref{fig:structfacts}, the query from the tourist llama involves structured knowledge about the Olympics and travel guides. However, language models typically generate answers by gathering factual knowledge separately, thereby overlooking the \textit{topological interdependencies}~\shortcite{10.1145/3581783.3611964}. Moreover, when gathering data from multiple sources, models should recognize the \textit{order invariance} of structural data. Unlike textual data, which is order-dependent, the order of entities within a similar topological structure should not affect the inherent factual knowledge. To explore \textit{how LLMs recognize and combine factual knowledge from multiple sources of structured data}, we include questions where knowledge for verification may be dispersed across multiple discontinuous structured sources, compromising different tables, lists, and pages.
    \item \textbf{Composition Understanding}. Reasoning about factual knowledge in structural data suffers from \textit{sparsity} due to missing values or incomplete descriptions. 
    Additionally, LLMs are expected to accurately reason through sparse information without misinterpreting \textit{topological interdependencies}. For instance, to answer the question in Figure \ref{fig:structfacts}, LLMs have to comprehend the header ``2024 U.S. Election'', which spans multiple columns with a missing value. This arise the question of \textit{whether LLMs can accurately understand the composition of structural data and reason out factual knowledge}. To this end, our \name\ includes a variety of intricate compositions, encompassing missing values, compositions within complex structures, and compositions with incomplete descriptions.
    \item \textbf{Combining Structured and Unstructured}. Given the \textit{sparsity} and \textit{lack of prior knowledge} of the domain-specific information in structural data, LLMs needs to fully leverage the factual knowledge learned from textual contexts. The knowledge presented in unstructured data (e.g., table captions) often provides an important context for understanding the information in structural data. Moreover, general knowledge acquired from pretraining texts can aid in inferring domain-specific knowledge within structural data. In Figure \ref{fig:structfacts}, general knowledge about nut proteins helps identify the potential cause of the allergy from the shopping receipt.
    Therefore, we are also interested in \textit{whether LLMs can accurately combine factual knowledge from unstructured contexts with reasoning over structured data}. To this end, our \name\ includes questions that require factuality verification spanning both structured and unstructured evidence.
\end{itemize}

\subsection{Data Collections and Statistics} 
To collect factual knowledge expressed in various ways, our \name\ benchmark sources from a range of public datasets: Table Fact Verification (TFV) datasets FEVEROUS ~\shortcite{FEVEROUS} and TabFact ~\shortcite{TabFact}, Table-to-Text (ToT) dataset ToTTo~\shortcite{totto}, and Table Question Answering (TQA) dataset SQA ~\shortcite{sqa}. All of these source datasets are built based on factual knowledge from Wikipedia. Every claim within the TFV datasets is labeled by its factuality and accompanied by structural evidence. Texts in the ToT dataset are sampled as factual claims along with their source tables as the corresponding structural evidence. Additionally, we filtered open-ended questions with one correct answer from the SQA~\shortcite{sqa} dataset as positive samples, including a supporting table as the evidence. Previous studies show that multi-choice questions are effective for inducing factual knowledge from LLMs~\shortcite{kadavath2022language,lin2022teaching}. We further asked the annotators to rewrite the original claims, and open-questions with answers into yes-or-no questions while preserving their factuality and task definition. For example, the open-question \textit{``Which country received 0 gold medals in the Los Angeles 1984 Olympics?''} with the answer \textit{``Colombia''} is rewritten as \textit{``Is Colombia one of the countries that didn't win any gold medals in the Los Angeles 1984 Olympics?''}. The annotators are paid according to the quality and quantity of their annotations. In our \name, factual questions from various data sources are labeled as Yes, No, or Not Sure Enough, which indicates whether the question is factual (Fact.), non-factual (Non-fact.), or if there is insufficient information (NEI) to provide an answer, respectively.

Our \name\ collects 8,340 questions based on factual knowledge as evidence in tables, lists, and sentences. We annotated the collected questions with the five aforementioned factual tasks to evaluate the capability of LLMs' reasoning over factual knowledge from structural data. Specifically, for each question, we had 12 undergraduate students proficient in English to select a task label from \textbf{Arithmetic Calculation}, \textbf{Spatiotemporal Cognition}, \textbf{Multi-hop Reasoning}, \textbf{Composition Understanding}, and \textbf{Combining Structured and Unstructured}, and further provide detailed annotations for the questions within each task to better control the complexity of reasoning. The statistical information of \name\ is detailed in Table \ref{tab:datasets_stats}. It is important to note that our benchmark dataset is derived from real-world data, which inherently introduces challenges related to label imbalance. 
To mitigate this issue in our analysis, we report accuracy and weighted F1 scores in our main results, while also providing balanced accuracy and F1 scores in Appendix \ref{sec:apdx_sup_res}.

To ensure the quality of annotation, we implemented a three-phase verification process. We divided the 12 annotators into three groups of five to cross-validate a random sample of 500 questions annotated by another group. First, we calculated the Cohen's Kappa score to quantify the consistency between each pair of annotation groups, resulting in an average score of 0.94. Second, we calculated the Fleiss' Kappa score to evaluate the consistency across multiple groups of annotators, yielding a final score of 0.80. Both scores indicate high-quality annotations. Finally, an author randomly chooses 10 questions from each of the five tasks and carefully examines the annotations.

\begin{table}[ht]
\vspace{-2mm}
  \centering
  \resizebox{\linewidth}{!}{
  \begin{tabular}{ccccc}
    \hline
    \multirow{2}{*}{Tasks} &\multicolumn{4}{c}{Distribution}\\
    \cmidrule{2-5}
    &Fact.&Non-Fact.&NEI&Overall \\
    \midrule
    Arithmetic Calc.&1,331 & 623 & 65 & 2,019 \\
    Spatiotemporal Cogn.&1,474 & 1,590 & 96 & 3,160 \\
    Multi-hop Reas.&1,183 & 293 & 54 & 1,530 \\
    Composition Und. &78 & 53 & 3 & 134 \\
    Struct. and Unstruct. &1,133 & 322 & 42 & 1,497\\
    \midrule
    Total& 5,199 & 2,881 & 260 & 8,340 \\
    \hline 
  \end{tabular}
  }
  \vspace{-2mm}
  \caption{The statistics of our \name\ benchmark.}
  \vspace{-5mm}\label{tab:datasets_stats}
\end{table}

\begin{table*}[!htbp]
  \centering
  \resizebox{\linewidth}{!}{
    \begin{tabular}{lccccccccccccc}
      \toprule
      \multirow{2}{*}{Methods} & \multicolumn{2}{c}{Zero-shot w/o CoT} & \multicolumn{2}{c}{Zero-shot w/ CoT} & \multicolumn{2}{c}{Few-shot w/o CoT} & \multicolumn{2}{c}{Few-shot w/ CoT} & \multicolumn{2}{c}{Overall} \\
      \cmidrule(lr){2-3} \cmidrule(lr){4-5} \cmidrule(lr){6-7} \cmidrule(lr){8-9} \cmidrule(lr){10-11}
      & Acc.  & F1 & Acc.   & F1 & Acc.   & F1 & Acc.  & F1 & Acc.   & F1 \\ 
      \midrule
Qwen2-7B & 31.82 & 40.39 & 49.40 & 52.65 & 45.39 & 50.85 & 54.80 & 57.52 & 45.35 & 50.35 \\
LLaMA-3-8B & 29.72 & 35.37 & 27.65 & 36.48 & 32.13 & 39.17 & 55.64 & 55.25 & 36.28 & 41.57 \\
Gemma-2-9B & 22.67 & 26.69 & 42.76 & 48.17 & 27.03 & 34.08 & \textbf{61.14} & 60.11 & 38.40 & 42.26 \\

\midrule
Qwen2-7B Instruct & 47.85 & 53.80 & 41.27 & 48.90 & 44.88 & 51.42 & 41.01 & 48.96 & 43.75 & 50.77\\
LLaMA-3-8B Instruct & \textbf{62.92} & 58.79 & 43.01 & 49.64 &\textbf{63.39} & \underline{60.23} & 45.43 & 53.58 & 53.69 & 55.56 \\
Gemma-2-9B It & 43.53 & 49.89 & 41.08 & 49.84 & 44.81 & 51.60 & 43.03 & 51.82 & 43.11 & 50.79\\
GLM-4-9B Chat & 52.56 &56.09 & 42.58 & 51.29 & 52.97 & 56.52 & 47.10 & 54.74 & 48.80 & 54.66 \\
Mistral-7B Instruct & 50.90 & 53.80 & 37.33 & 46.19 & \underline{58.78} & 59.45 & 43.80 & 51.88 & 47.70 & 52.83\\
\midrule
GPT-4o-mini & 60.80 & \underline{63.67} & \textbf{54.20} & \textbf{60.71} & 55.06 & 60.10 & \underline{56.35} & \textbf{62.20} & \textbf{56.60} & \textbf{61.67} \\
GPT-4-turbo & \underline{62.50} & \textbf{64.58} & \underline{53.31} & \underline{60.20} & 56.01 & \textbf{60.99} & 53.18 & \underline{60.19} & \underline{56.25} & \underline{61.49} \\
      \bottomrule
    \end{tabular}
}
\caption{Performance of 10 LLMs on the \name\ benchmark using various prompts.}
\vspace{-5mm}
\label{tab:main_res}
\end{table*}

\section{Main Results}
To investigate the factual reasoning capabilities of LLMs on structured data, we conduct experiments with \name\ on 10 LLMs trained via pretraining, instruction tuning, and reinforcement learning with human feedback (RLHF). Detailed descriptions of the employed prompting strategies and selected LLMs can be found in Figure \ref{fig:prompts_main_res}, Appendix \ref{sec:apdx_llms_intro} and \ref{sec:apdx_prompts}. To address the issue of imbalanced labels, we evaluate using weighted accuracy and the F1 score, where the weights correspond to the label distribution ratio. Considering the order bias inherent in LLMs~\shortcite{pezeshkpour-hruschka-2024-large,zheng2024large,judging2024zheng}, we run each model three times with different order of options in Figure \ref{fig:prompts_main_res}, and report the average result.

\subsubsection{Different Prompts}\label{sec:main_res}
In Table \ref{tab:main_res}, we adhere to the input formats used in previous studies~\shortcite{singha2023tabular,tablemeetsllm,wangchain}, where factual questions from \name\ are combined with corresponding structured data and fed into these LLMs, prompting the models to answer the questions based on the provided data, as shown in Figure \ref{fig:prompts_main_res}. %
Based on the results in Table \ref{tab:main_res}, we conclude the following findings.

\begin{itemize}
    \item From the overall standpoint, models with instruction tuning exhibit superior results compared to the pretrained models. The results obtained by LLaMA-3-8B Instruct, Gemma-2-9B, and Qwen2-7B Instruct outperform their corresponding pretrained models, with an average F1 score improvement of 7.65\%.
    \item Performance improves as the number of model parameters increases. Although instruction-tuned 7B models achieve accuracy levels comparable to GPT models, they exhibit lower F1 scores. Meanwhile, we found that GPT models tend to be more cautious, as evidenced by a significant proportion of NEI responses. Please refer to the Appendix \ref{sec:apdx_sup_res} for detailed distributions of responses. 
    \item Both the Chain of Thought (CoT)~\shortcite{wei2022chain} and few-shot strategies effectively guide pre-trained models in utilizing their factual knowledge. In a zero-shot setting without CoT, the performance of pre-trained models falls below random guessing (with a probability of 33\%); incorporating few-shot learning and CoT results in an average F1 score improvement of 23.47\%.
    \item The CoT strategy has even negative impact on instruction-tuned models, and few-shot examples yield limited improvements. More complex prompting strategies (see Appendix \ref{sec:apdx_sup_res}) also result in modest gains in instruction-tuned models. This indicates that prompt optimization has limited effectiveness for these LLMs, likely because instruction tuning enables the models to understand tasks with the simplest prompts.
\end{itemize}

\subsubsection{Different Tasks}
\begin{table*}[!htbp]
\label{main-result}
\centering
\resizebox{\textwidth}{!}{
\begin{tabular}{lcccccccccccc}
\toprule
\multirow{2}{*}{Methods} & \multicolumn{2}{c}{Arithmetic Calc.} & \multicolumn{2}{c}{Spatiotemporal Cogn.}& \multicolumn{2}{c}{Multi-hop Reas.} & \multicolumn{2}{c}{Composition Und.}  & \multicolumn{2}{c}{Struct. \& Unstruct.} \\

\cmidrule(lr){2-3} \cmidrule(lr){4-5} \cmidrule(lr){6-7} \cmidrule(lr){8-9} \cmidrule(lr){10-11}
& Acc. & F1 & Acc. & F1 & Acc. & F1 & Acc. & F1 & Acc. & F1 \\ \midrule

Qwen2-7B & 28.30 & 37.56 & 28.24 & 34.88 & 34.05 & 45.05 & 38.31 & 46.75 & 41.26 & 51.18 \\
LLaMA-3-8B & 28.48 & 34.60 & 28.36 & 32.25 & 29.61 & 37.54 & 34.08 & 39.11 & 34.00 & 41.85 \\
Gemma-2-9B & 15.98 & 21.06 & 21.45 & 23.63 & 30.55 & 35.86 & 25.87 & 29.89 & 25.92 & 32.05 \\

\midrule
Qwen2-7B Instruct & 54.58 & 57.26 & 40.52 & 47.06 & 47.91 & 56.28 & 57.46 & 61.59 & 53.33 & 60.98  \\
LLaMA-3-8B Instruct & 62.28 & 57.10 & 54.78 & 50.95 & \textbf{70.61} & \textbf{68.37} & 60.94 & 57.68 & \textbf{73.28} & \textbf{70.89}  \\
Gemma-2-9B It & 51.36 & 53.92 & 33.03 & 39.15 & 44.73 & 54.42 & 59.95 & 62.54 & 52.46 & 60.96  \\
GLM-4-9B Chat & 59.27 & 59.21 & 46.70 & 51.04 & 50.67 & 56.50 & 63.93 & 63.85 & 56.80 & 62.59 \\
Mistral-7B Instruct & 55.37 & 56.07 & 43.44 & 46.15 & 52.03 & 58.18 & 54.98 & 55.21 & 59.07 & 63.91 \\

\midrule
GPT-4o-mini & \underline{62.52} & \underline{63.23} & \underline{60.13} & \underline{63.33} & 58.04 & \underline{63.72} & \underline{67.42} & \underline{67.96} & 62.10 & 66.88 \\
GPT-4-turbo & \textbf{62.86} & \textbf{63.28} & \textbf{62.98} & \textbf{65.09} & \underline{58.36} & 62.53 & \textbf{68.91} & \textbf{69.30} & \underline{64.64} & \underline{68.99} \\
\midrule
Overall & 48.10 & 50.33 & 41.96 & 45.35 & 47.66 & 53.85 & 53.19 & 55.39 & 52.29 & 58.03\\ 
\bottomrule
\end{tabular}
}
\vspace{-2mm}
\caption{Performance of 10 LLMs on the \name\ benchmark across five factual tasks under the zero-shot w/o CoT setting.}
\vspace{-5mm}
\label{tab:tasks}
\end{table*}

\begin{figure}[!tbp]
  \centering
  \includegraphics[scale=0.5]{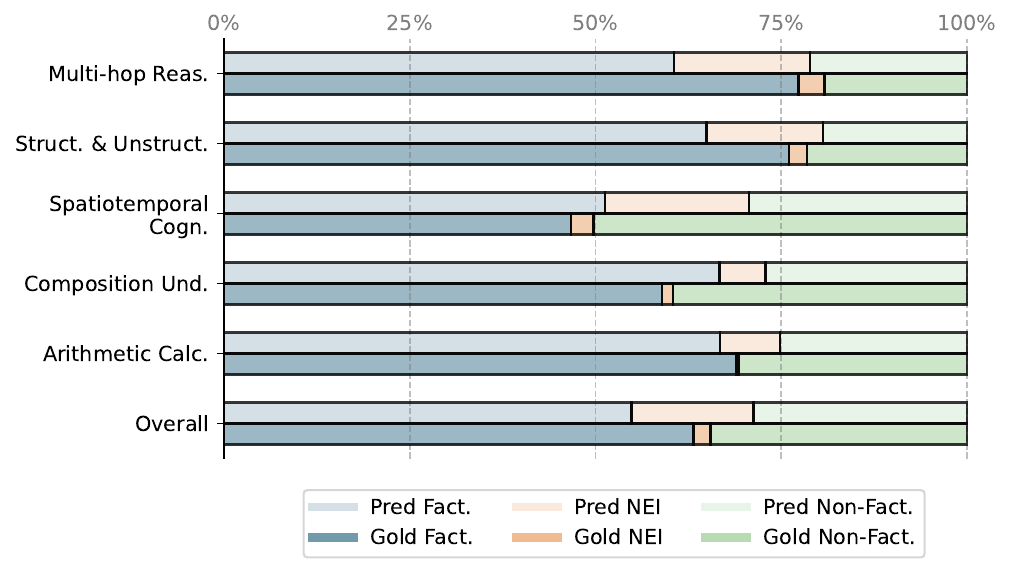}
  \vspace{-7mm}
  \caption{The distribution of three types of responses across five tasks, averaged across 10 LLMs.}
  \label{fig:distribution}
  \vspace{-5mm}
\end{figure}

We further analyze the reasoning performance of the selected 10 LLMs on structured data from the standpoint of various factual tasks. Table \ref{tab:tasks} presents the performances, we have the following observations.
\begin{itemize}
    \item LLMs exhibit inferior performance on spatiotemporal cognition and arithmetic calculation, with average weighted F1 scores of 45.35\% and 50.33\%, respectively. This underperformance is attributed to their limitations in handling the \textit{heterogeneous} data within structures.
    \item Among the five factual tasks, LLMs perform relatively well on those addressing the \textit{sparsity} issue in structured data, particularly in Composition Understanding and Combining Structured and Unstructured data. Further case studies reveal that LLMs effectively utilize pretraining or domain-specific knowledge to generate accurate responses in these tasks.
    \item We further analyze the distribution of three labels in Figure \ref{fig:distribution} and observed that the proportions of predicted NEI labels are generally higher than those of the gold labels across the five tasks, the proportions of factual and non-factual responses vary between different tasks. This indicates that, akin to human behavior, LLMs demonstrate caution when accepting or rejecting factual queries.
\end{itemize}

\section{Analysis}
Building on the main results, we delve deeper by conducting a series of in-depth analyses from various perspectives to evaluate the LLMs' capabilities in completing the five factual tasks on structured data, using GPT-4o-mini as the representative model. Further analysis of other LLMs is provided in the Appendix \ref{sec:apdx_sup_res}.

\subsection{Resilience to Evidence}\label{sec:context_resil}
We first aim to investigate whether the ability of LLMs to answer factual questions is influenced by the presence of structured data as evidence. In this context, ``evidence'' refers to the structured data that corresponds to the questions in the prompts, as illustrated in Figure \ref{fig:prompts_main_res}. We categorize the model's resilience to evidence into three levels, ranging from stringent to adaptable: (i) efficiently understanding and reasoning with the provided structured data as evidence, (ii) adapting to irrelevant interventions in the structure of the evidence data, and (iii) accurately recalling prior general knowledge without the support of structured data. We expect LLMs to sustain strong performance across these three levels, showcasing remarkable resilience.

\begin{figure}[!t]
  \centering
  \vspace{-10mm}
  \includegraphics[scale=0.28]{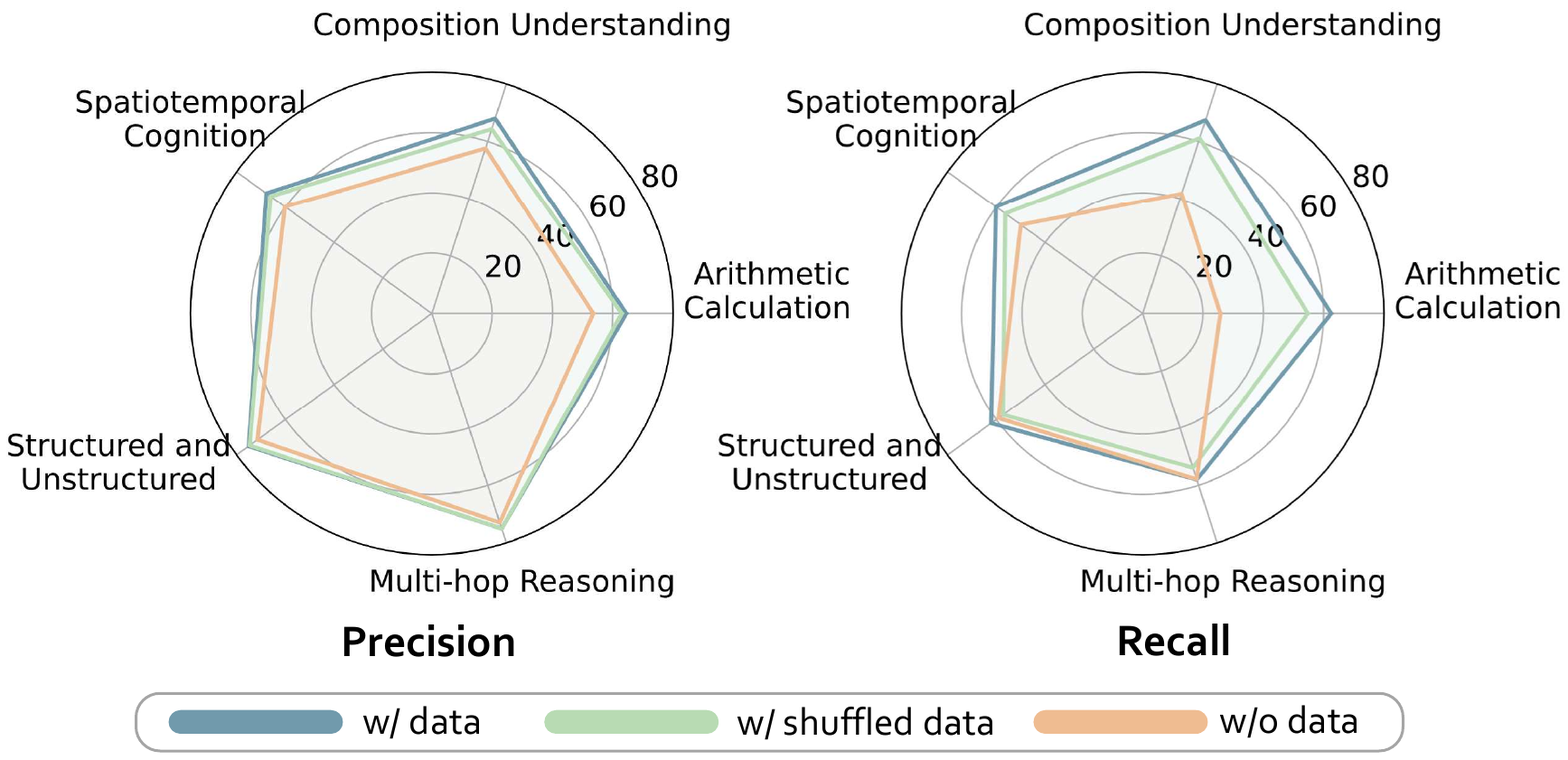}
  \vspace{-12mm}
  \caption{Performance of GPT-4o-mini under different settings of structured evidence.}
  \label{fig:radar-resilience}
  \vspace{-4mm}
\end{figure}

\begin{figure}[!t]
  \centering
  \includegraphics[scale=0.2]{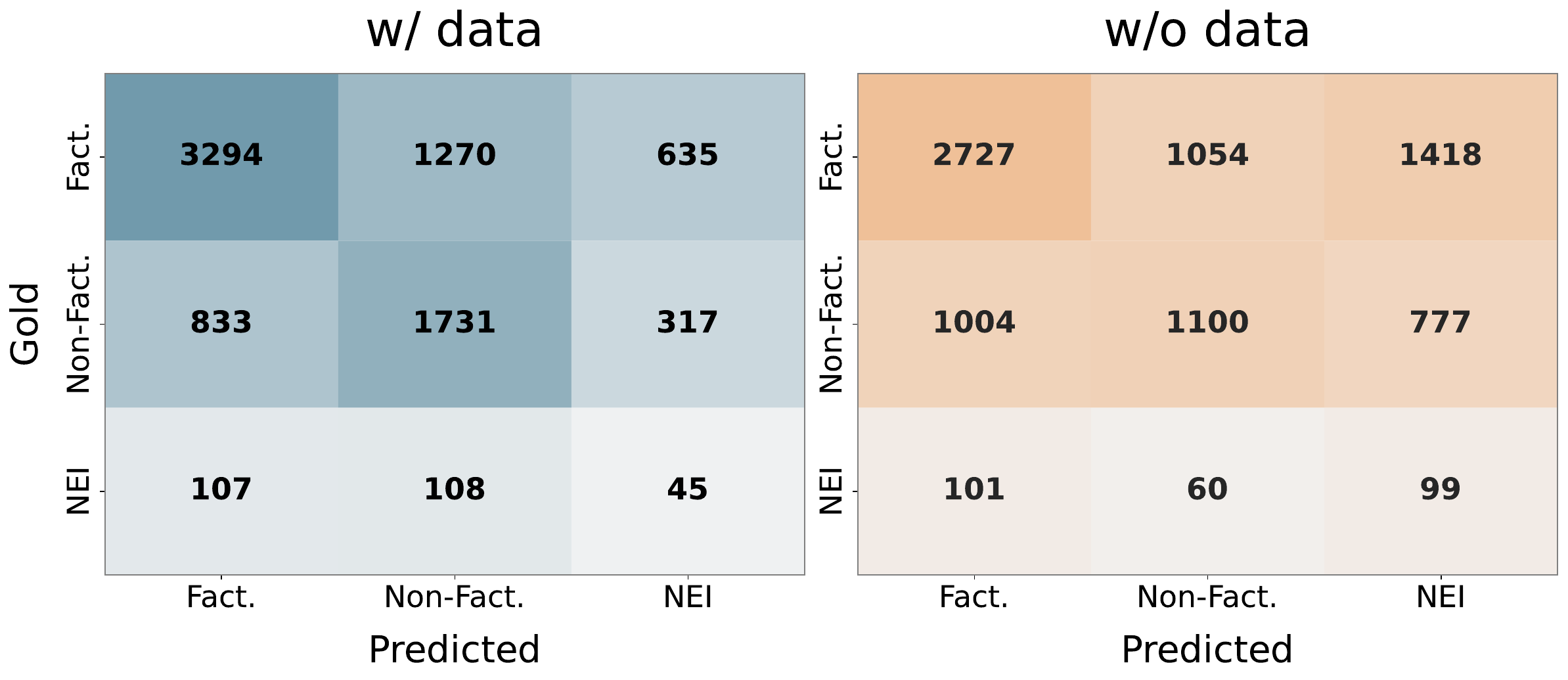}
  \vspace{-4mm}
  \caption{Confusion matrices of GPT-4o-mini's performance under the settings w/ and w/o structured data as evidence.}
  \label{fig:cm}
  \vspace{-5mm}
\end{figure}

To this end, we assess the performance of GPT-4o-mini under three distinct conditions: (i) with structured data provided as corresponding evidence for the factual questions (denoted as ``w/ data'' in Figure \ref{fig:radar-resilience}), (ii) with the structured data shuffled (denoted as ``w/ shuffled data'' in Figure \ref{fig:radar-resilience}), and (iii) without any structured data as evidence (denoted as ``w/o data'' in Figure \ref{fig:radar-resilience}). The first condition aligns with the zero-shot without the CoT setting in the main results (Table \ref{tab:main_res}). In the second condition, we exploit the \textit{order invariance} property of structured data to introduce semantically irrelevant interventions by shuffling the rows and columns in tables and the elements in lists in our \name\ benchmark. For the third condition, since all factual questions in our framework are supported by structured data from Wikipedia, we anticipate that the LLM will rely on its pretraining knowledge to effectively handle scenarios where evidence is absent. Details of the prompting strategies used in this analysis can be found in Appendix \ref{sec:apdx_prompts}.

As depicted in Figure \ref{fig:radar-resilience}, we assess the resilience to evidence of LLMs across five factual tasks under these three conditions. Notably, the LLM shows only a marginal decrease in performance when transitioning from original structured data to shuffled structured data, demonstrating its strong adaptability to the \textit{order invariance} properties of structured data. This marginal decrease is particularly evident in Arithmetic Calculation and Spatiotemporal Cognition, where recall drops by 7.74\% and 3.83\%, respectively. This indicates that LLMs place more reliance on the sequence of information when reasoning over \textit{heterogeneous} data. The evidence-absent scenario presents a more pronounced decline, with representative drops of 19.04\% and 29.02\% in Composition Understanding and Arithmetic Calculation, respectively. The dramatic drop in Arithmetic Calculation is attributed to the loss of numerical information in structured data. The decline in Composition Understanding highlights that the LLM struggles to grasp complex structures and effectively utilize general knowledge. The complete scores of Figure \ref{fig:radar-resilience} are provided in Appendix \ref{sec:apdx_sup_res}.

We further analyze the performance decline from the evidence-rich scenario (w/ data) to the evidence-absence scenario (w/o data) by comparing the confusion matrices in Figure \ref{fig:cm}. This decline can be attributed to the increased number of misclassifications across all three classes, i.e., Fact., Non-Fact., and NEI. In particular, the notable increase in misclassification of facts as NEI indicates that without structured data as evidence, the LLM models struggle to recognize the completeness or relevance of the facts, leading to uncertainty and a higher tendency to classify facts as NEI.

\subsection{Fine-grained Studies of Different Tasks}
In this section, we conduct fine-grained analyses of GPT-4o-mini's reasoning on structured data across the five tasks.

\begin{figure}[!t]
  \centering
  \begin{minipage}[t]{0.48\linewidth} %
    \subfigure[Arithmetic Calculation]{
        \label{fig:arithmetic}
        \includegraphics[width=\linewidth]{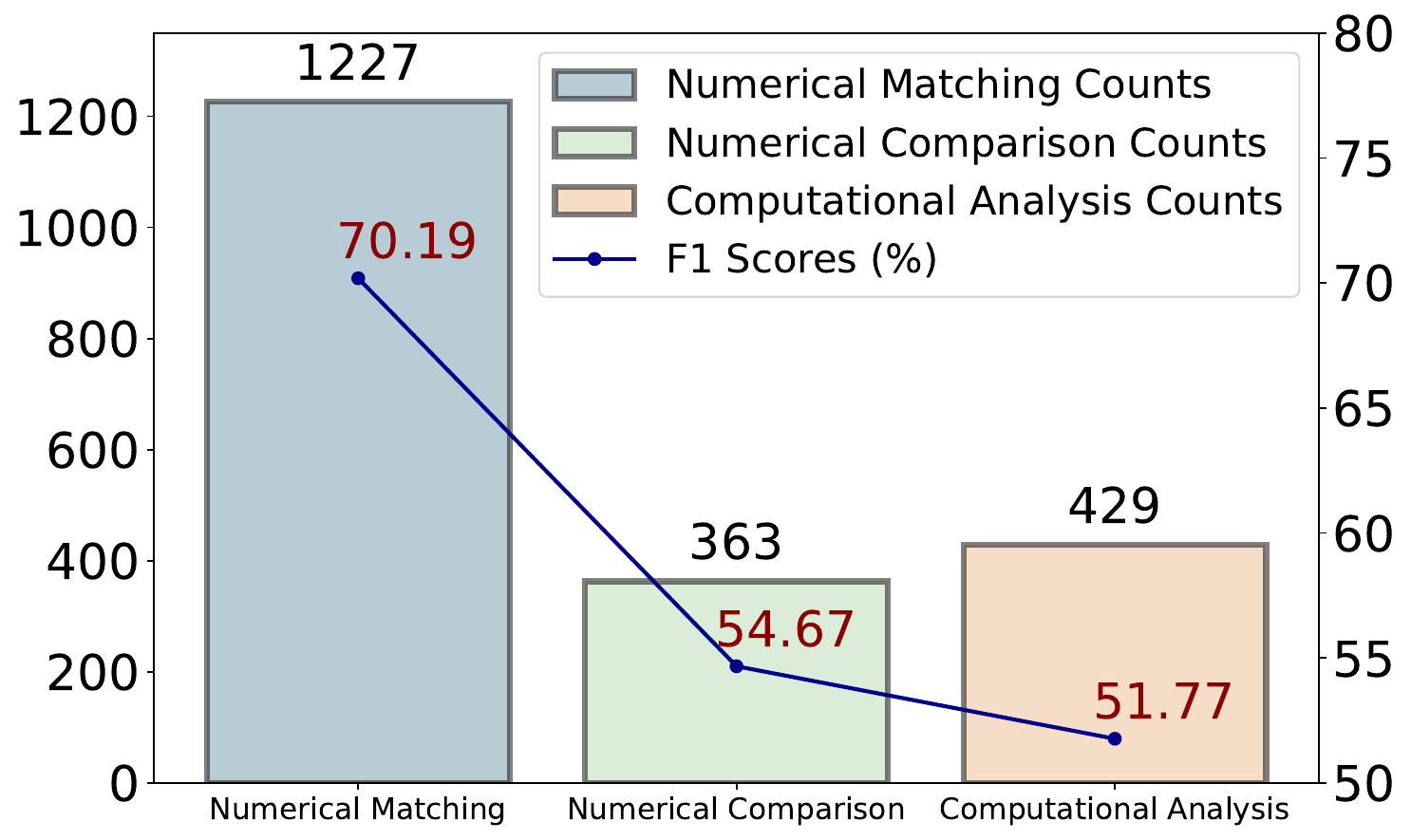}
    }
  \end{minipage}
  \hfill %
  \begin{minipage}[t]{0.48\linewidth} %
    \subfigure[Multi-hop Reasoning]{
        \label{fig:multihop}
        \includegraphics[width=\linewidth]{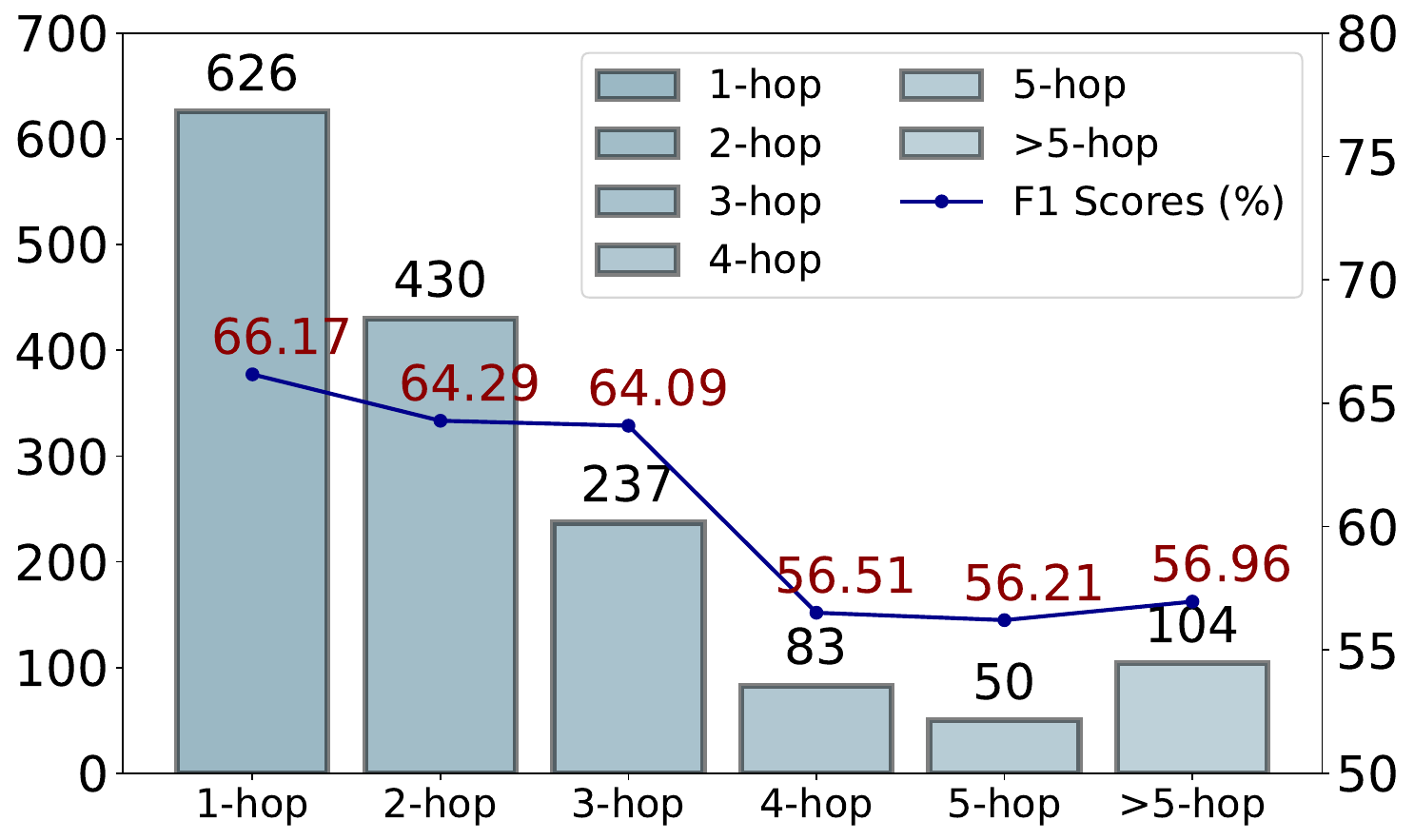}
    }  
  \end{minipage}
  \vspace{-1mm} %
  \\
  \begin{minipage}[t]{0.48\linewidth} %
    \subfigure[Composition Understanding]{
        \label{fig:composition}
        \includegraphics[width=\linewidth]{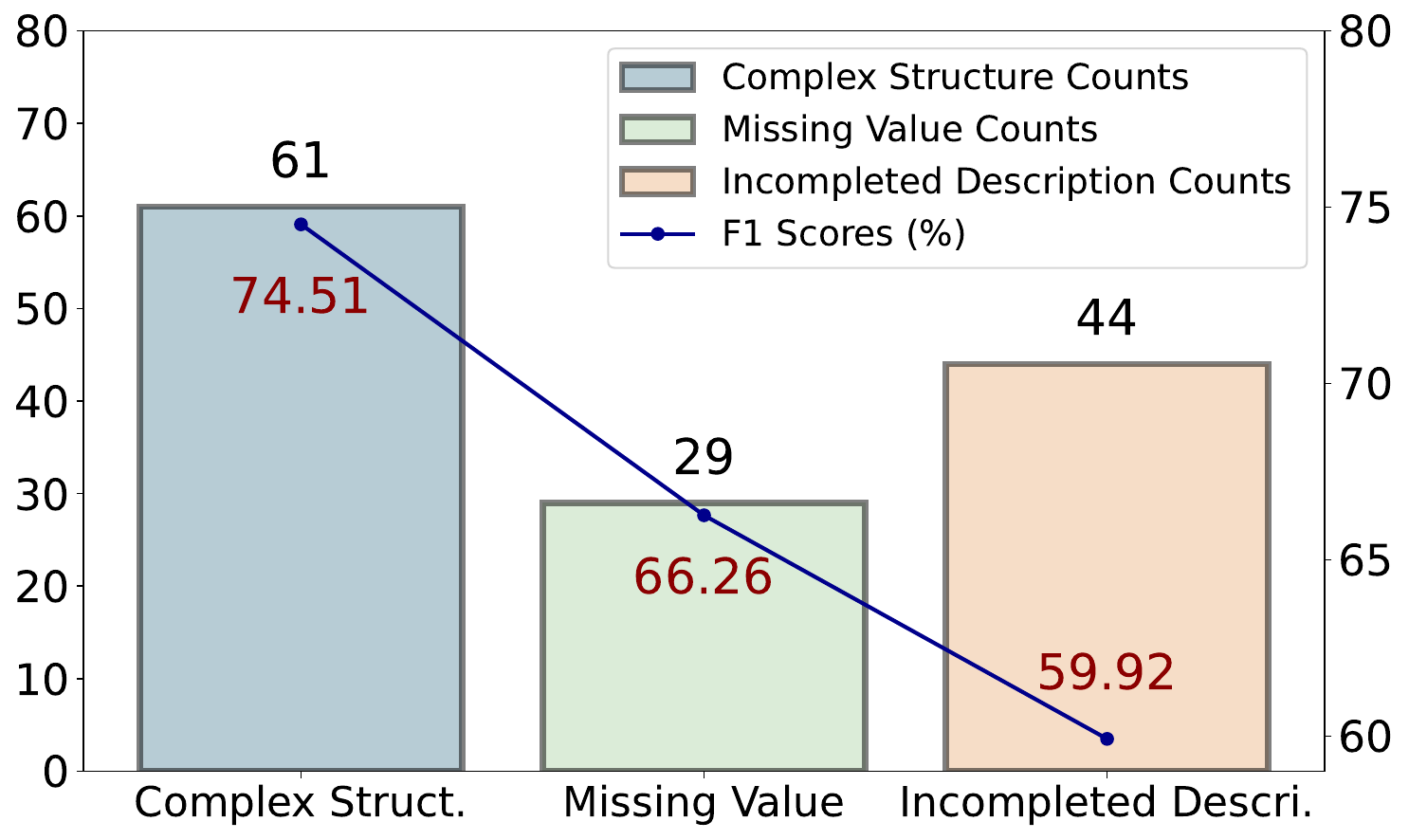}
    }  
  \end{minipage}
  \hfill %
  \begin{minipage}[t]{0.48\linewidth} %
    \subfigure[Structured and Unstructured]{
        \label{fig:combine}
        \includegraphics[width=\linewidth]{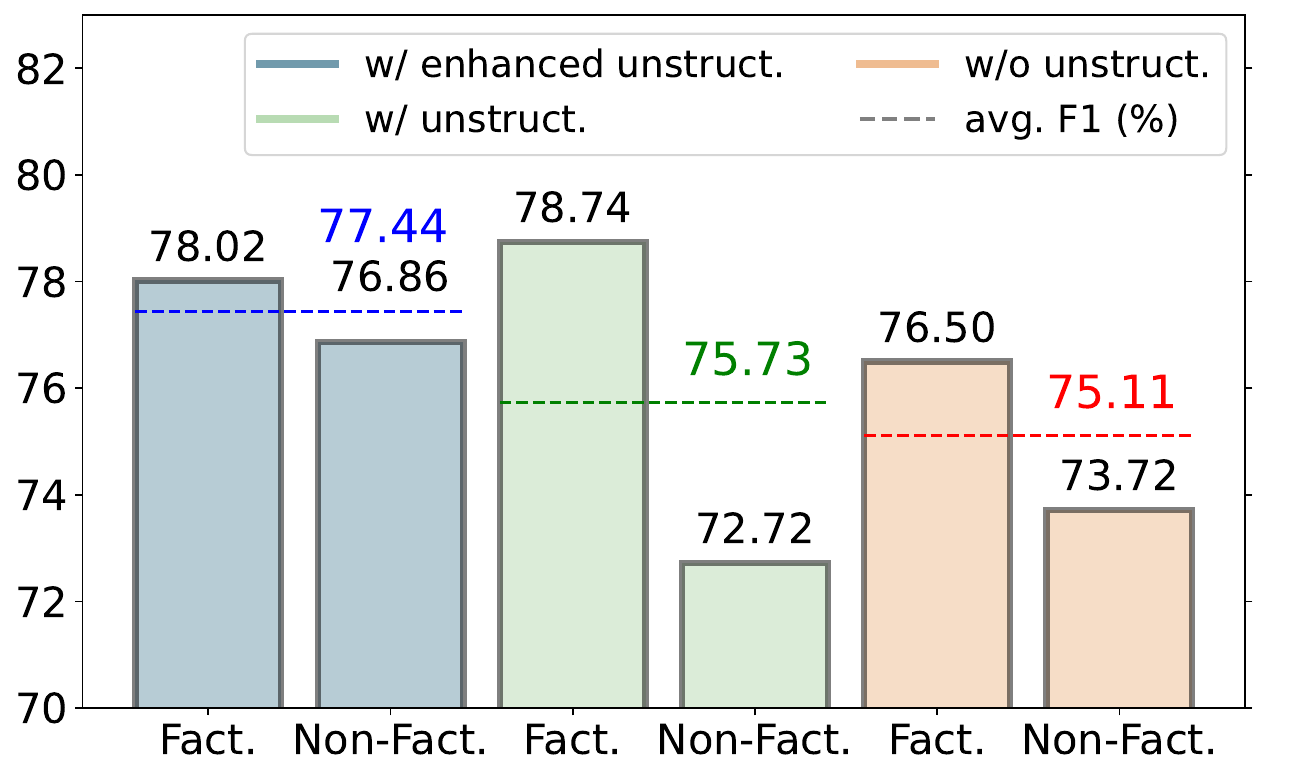}
    }  
  \end{minipage}
  \vspace{-1mm} %
  \\
  \begin{minipage}[t]{\linewidth} %
    \centering
    \subfigure[Spatiotemporal Cognition]{
        \label{fig:spatiotemporal}
        \includegraphics[width=\linewidth]{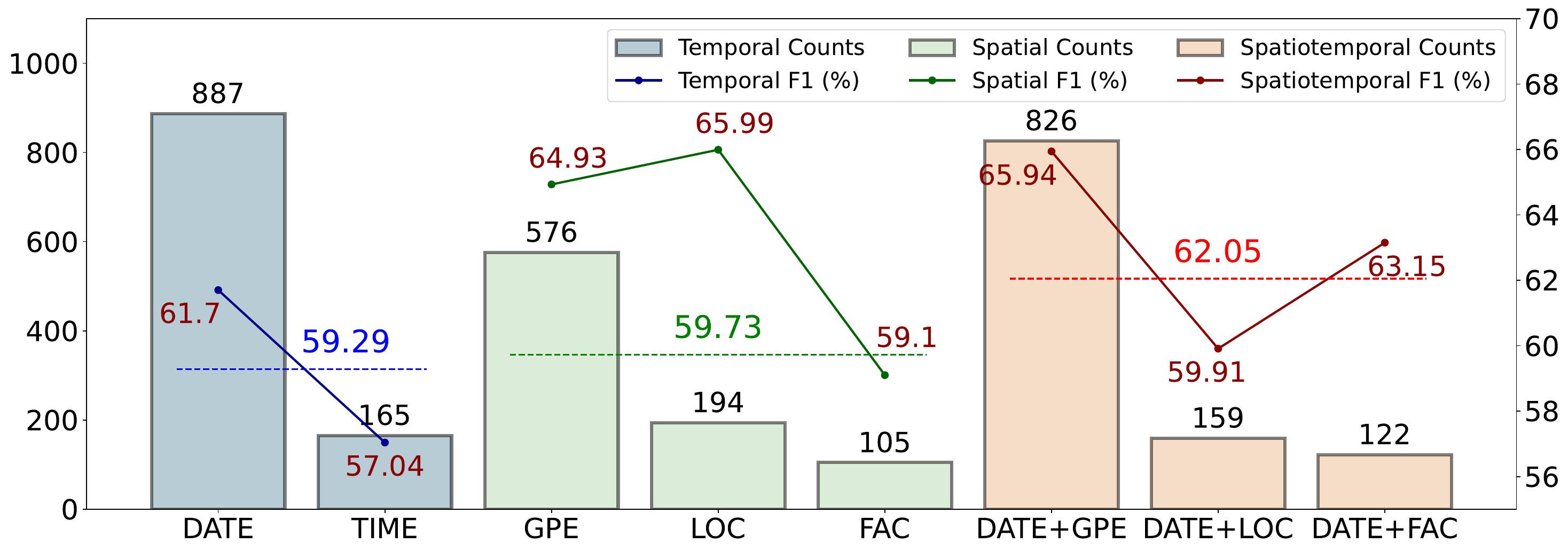}
    }  
  \end{minipage}
  \vspace{-3mm} %
  \caption{Fine-grained analysis of GPT-4o-mini's performance over the five tasks under zero-shot w/o CoT setting.}
  \vspace{-5mm}
\end{figure}

\subsubsection{Arithmetic Calculation} 
To assess whether large language models (LLMs) are capable of capturing and memorizing arithmetic rules, we categorize the questions in the arithmetic calculation task into three levels of mathematical problems depending on varying degrees of arithmetic difficulty: numerical matching, numerical comparison, and computational analysis. For instance, the factual question ``Are the White Blood Cell (WBC) counts within the normal range?'' as illustrated in Figure \ref{fig:prompts_main_res}, falls under the numerical comparison category. Figure \ref{fig:arithmetic} presents the performance of GPT-4o-mini across these three categories of mathematical problems. The results suggest that while LLMs can effectively handle numerical matching tasks, they encounter challenges with more complex computational analysis, such as statistical problems.

\subsubsection{Spatiotemporal Cognition} As shown in Table \ref{tab:tasks}, LLMs exhibit inadqequate performance in the Spatiotemporal Cognition task. We conducted a detailed analysis of GPT-4o-mini's performance across different named entity categories. In Figure \ref{fig:spatiotemporal}, we classified the Spatiotemporal Cognition questions in \name\ into three categories: (i) temporal, which includes questions about dates (DATE), and times (TIME); (ii) spatial, encompassing questions related to political regions such as countries and cities (GPE), as well as locations such as mountains and rivers (LOC), and artificial landmarks (FAC); and (iii) spatiotemporal, which involves questions containing both temporal and spatial entities (DATE+GPE, DATE+LOC, DATE+FAC). Overall, the LLM is more effective at cognizing and reasoning over spatiotemporal knowledge compared to reasoning over data containing only temporal or spatial entities. The performance variance across different entity types suggests that the model performs more effectively when dealing with entities that represent finer granularity in both temporal and spatial dimensions. Specifically, the model performs better on TIME entities compared to DATE, and within spatial entities, it shows higher accuracy in recognizing smaller units, with GPE outperforming LOC and FAC.%

\subsubsection{Multi-hop Reasoning} To investigate the capability of LLMs in recognizing and combining knowledge from various discontinuous sources of structured data, we categorized factual questions in the Multi-hop Reasoning task at a more fine-grained level based on the number of hops required to arrive at an answer.  A ``hop'' refers to the step in which the LLM needs to infer knowledge by combining knowledge from two data sources. In particular, in our analysis, each source is defined as a Wikipedia element (e.g., cells, headers, captions in tables, or items in lists) that serves as evidence supporting the gold answer. Figure \ref{fig:multihop} reveals a clear trend: as reasoning tasks become more complex, requiring an increasing number of hops, the LLMs' effectiveness in reasoning over factual knowledge from structured data diminishes. Notably, there is a significant performance decline after 3-hop questions, with a 7.58\% decrease in F1 score observed in 4-hop questions.

\subsubsection{Composition Understanding}
To answer whether LLMs can accurately reason factual knowledge from challenging compositions in structured data, we categorize these compositions into three types of irregularities: (i) complex structure, where compositions involve intricate dependencies such as a single table cell spanning multiple columns; (ii) missing values, where cells contain unknown values; and (iii) incomplete descriptions, where cells have ambiguous or insufficient descriptions. Figure \ref{fig:composition} illustrates that the primary bottleneck in enhancing LLM performance in the Composition Understanding task lies in addressing the challenge of incomplete descriptions. This challenge is associated with the characteristics of \textit{lack of prior}, indicating that accurately aligning with the domain-specific knowledge in structured data remains a significant obstacle for LLMs.

\subsubsection{Combining Structured and Unstructured}
A prominent strength of LLMs in factual reasoning is their ability to comprehend and reason with knowledge in textual data. When extending this capability to tasks that involve structured data, it becomes imperative to assess whether LLMs can effectively combine factual knowledge extracted from unstructured contexts with reasoning applied to structured data. Therefore, beyond the original unstructured context provided for the question in the Combining Structured and Unstructured task, we assess the capability of LLMs in scenarios with enhanced unstructured context, as well as in situations where unstructured context is absent. The results shown in Figure \ref{fig:combine} illustrate that the performance of LLMs can be enhanced by the availability of unstructured contexts when handling factual reasoning over structured data. It is noteworthy that in non-factual tasks, LLMs performed slightly worse when provided with the original unstructured context, compared to when no unstructured context was available. Upon reviewing the cases, we found that LLMs are more sensitive to the quality of unstructured context in non-factual tasks, as evidenced by the significant improvement when unstructured data is enhanced.

\begin{figure}[!t]
  \centering
  \vspace{-2mm}
  \includegraphics[scale=0.47]{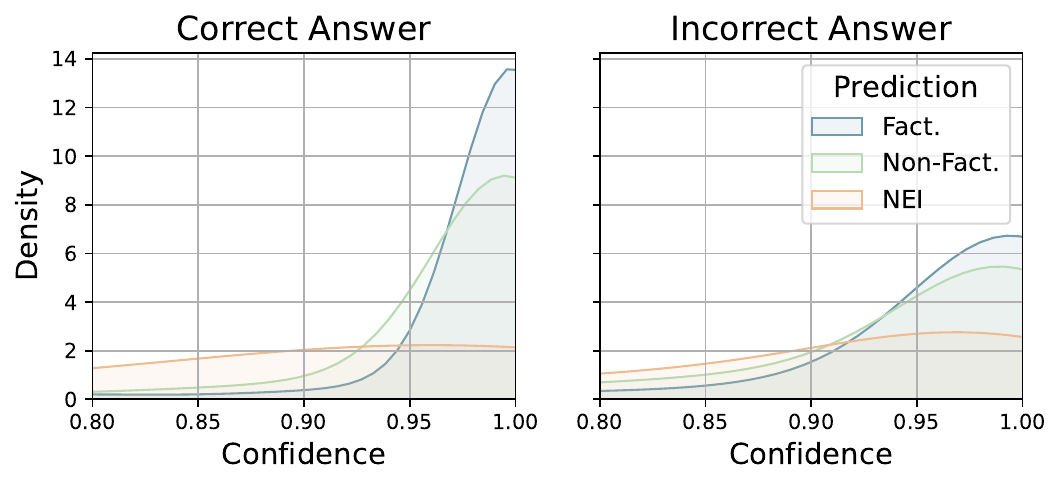}
  \vspace{-7mm}
  \caption{Confidence distribution in GPT-4o-mini's answers.}
  \vspace{-6mm}
  \label{fig:confidence}
\end{figure}

\subsection{Model Confidence}
Towards reliable reasoning outcomes, we also concern about the confidence of LLMs' outputs. Specifically, 
We used the probability of the model's first token output corresponding to the predicted label as the confidence score.
Figure \ref{fig:confidence} illustrates that the model shows higher confidence with its predicted facts, compared to those are predicted as non-factual. For NEI responses, whether predicted correctly or not, the confidence remains low with only minor variations. Notably, the models exhibit uncertainty in their incorrect answers, as evidenced by a significant drop in confidence, especially for factual and non-factual predictions. This indicates that the model's confidence levels are somehow aligned with the accuracy of their predictions, which could be useful for gauging the reliability of the model's answer.

\section{Related Work}\label{sec:related_work}
Structural data includes forms like tables, trees, and graphs are widely used for storing factual knowledge in a wide range of fields. For example, the Electronic Health Record (EHR). Previous research has explored the use of Large Language Models (LLMs) to understand information in structured data, such as tables~\shortcite{liu2023rethinking}. The recent surge in LLM development has enabled their application to comprehend and reason with structured data. Given the fixed architecture of LLMs and the common availability of APIs, researchers have focused on in-context learning and prompt engineering ~\shortcite{wangchain,zhao2023large}. For example, ~\shortcite{zhang2024trustuqa} uses a translator LLM to convert queries in Table Question Answering (TQA) tasks into a format more comprehensible for a LLM answerer. Additionally, LLMs are also trained to fit the objectives of structural tasks; for example, ~\shortcite{zhuang2024structlm} finetuned a generalist LLM model with large-scale instruction datasets on structural knowledge. However, LLMs are still struggling to extract factual knowledge from structural data~\shortcite{wang2023survey}. This poses challenges for applying LLMs to real-world tasks involving structural data. To address this, retrieval-augmented generation (RAG) methods are employed to query factual knowledge in LLMs ~\shortcite{guan2024mfort,ding2024survey}. For example, ~\shortcite{guan2024mfort} uses retrieved contents to enhance thoughts derived from chain-of-though reasoning. Additionally, multi-modal data, such as charts, are utilized to prevent factual errors ~\shortcite{krichene2024faithful,deng2024tables}.

\section{Future Directions}
Based on these findings, we propose several future research directions to promote the use of LLMs in downstream knowledge-sensitive tasks that rely on structured data. 

The performance of LLMs diminishes as the availability of structured evidence shifts from abundant to absent. Considering the limited enhancements achieved through prompt engineering on instruction-tuned models, incorporating an additional structure-aware module may be a more effective approach to further learning from structured data. Such specialized modules facilitate task-adaptive learning and knowledge transfer while maintaining computational and time costs at an effective level. Furthermore, LLMs demonstrate significant potential in leveraging unstructured knowledge to supplement structured data. %
A key challenge in this supplementation is preventing the distortion of specific knowledge within the structured data. Future research could focus on employing reinforcement learning to iteratively correct distortions in structured data reasoning.

\section{Conclusions}
In this work, we present \name, a benchmark specifically developed to assess the factual reasoning abilities of LLMs when dealing with structured data. The benchmark comprises 8,340 questions spanning five distinct tasks. Upon applying the \name\ we observed that LLMs encounter significant challenges in reasoning over heterogeneous data embedded in structures, particularly in executing complex arithmetic operations. Additionally, the diminished resilience of LLMs towards evidence indicates that effectively utilizing their knowledge base for reasoning on structured facts remains a formidable challenge. Our work underscores the pressing need to develop advanced techniques that help LLMs to better comprehend and utilize structured data. With the vast wealth of knowledge stored in structured data, we believe this will enhance reliable reasoning and pave the way for broader applications of LLMs in the future.

\bibliography{main}

\begin{thebibliography}{66}
\providecommand{\natexlab}[1]{#1}

\bibitem[{Akhtar, Cocarascu, and Simperl(2022)}]{pubhealthtab}
Akhtar, M.; Cocarascu, O.; and Simperl, E. 2022.
\newblock {P}ub{H}ealth{T}ab: {A} Public Health Table-based Dataset for Evidence-based Fact Checking.
\newblock In Carpuat, M.; de~Marneffe, M.-C.; and Meza~Ruiz, I.~V., eds., \emph{Findings of the Association for Computational Linguistics: NAACL 2022}, 1--16. Seattle, United States: Association for Computational Linguistics.

\bibitem[{Aly et~al.(2021)Aly, Guo, Schlichtkrull, Thorne, Vlachos, Christodoulopoulos, Cocarascu, and Mittal}]{FEVEROUS}
Aly, R.; Guo, Z.; Schlichtkrull, M.~S.; Thorne, J.; Vlachos, A.; Christodoulopoulos, C.; Cocarascu, O.; and Mittal, A. 2021.
\newblock FEVEROUS: Fact Extraction and VERification Over Unstructured and Structured information.
\newblock In \emph{Thirty-fifth Conference on Neural Information Processing Systems Datasets and Benchmarks Track (Round 1)}.

\bibitem[{Amazon Web~Services(2024)}]{amzstructured}
Amazon Web~Services, I. 2024.
\newblock What is structured data?

\bibitem[{Chen et~al.(2024)Chen, Sarkar, Lausen, Srinivasan, Zha, Huang, and Karypis}]{chen2024hytrel}
Chen, P.; Sarkar, S.; Lausen, L.; Srinivasan, B.; Zha, S.; Huang, R.; and Karypis, G. 2024.
\newblock HYTREL: Hypergraph-enhanced tabular data representation learning.
\newblock \emph{Advances in Neural Information Processing Systems}, 36.

\bibitem[{Chen et~al.(2020{\natexlab{a}})Chen, Chang, Schlinger, Wang, and Cohen}]{chen2020open}
Chen, W.; Chang, M.-W.; Schlinger, E.; Wang, W.~Y.; and Cohen, W.~W. 2020{\natexlab{a}}.
\newblock Open Question Answering over Tables and Text.
\newblock In \emph{International Conference on Learning Representations}.

\bibitem[{Chen et~al.(2020{\natexlab{b}})Chen, Wang, Chen, Zhang, Wang, Li, Zhou, and Wang}]{TabFact}
Chen, W.; Wang, H.; Chen, J.; Zhang, Y.; Wang, H.; Li, S.; Zhou, X.; and Wang, W.~Y. 2020{\natexlab{b}}.
\newblock TabFact : A Large-scale Dataset for Table-based Fact Verification.
\newblock In \emph{International Conference on Learning Representations (ICLR)}. Addis Ababa, Ethiopia.

\bibitem[{Chen et~al.(2020{\natexlab{c}})Chen, Zha, Chen, Xiong, Wang, and Wang}]{hybridqa}
Chen, W.; Zha, H.; Chen, Z.; Xiong, W.; Wang, H.; and Wang, W.~Y. 2020{\natexlab{c}}.
\newblock {H}ybrid{QA}: A Dataset of Multi-Hop Question Answering over Tabular and Textual Data.
\newblock In Cohn, T.; He, Y.; and Liu, Y., eds., \emph{Findings of the Association for Computational Linguistics: EMNLP 2020}, 1026--1036. Online: Association for Computational Linguistics.

\bibitem[{Chen et~al.(2021)Chen, Zhao, Chen, Zhang, Ji, Luo, Xiong, and Yu}]{websrc}
Chen, X.; Zhao, Z.; Chen, L.; Zhang, D.; Ji, J.; Luo, A.; Xiong, Y.; and Yu, K. 2021.
\newblock Websrc: A dataset for web-based structural reading comprehension.
\newblock \emph{arXiv preprint arXiv:2101.09465}.

\bibitem[{Cobbe et~al.(2021)Cobbe, Kosaraju, Bavarian, Chen, Jun, Kaiser, Plappert, Tworek, Hilton, Nakano et~al.}]{gsm8k}
Cobbe, K.; Kosaraju, V.; Bavarian, M.; Chen, M.; Jun, H.; Kaiser, L.; Plappert, M.; Tworek, J.; Hilton, J.; Nakano, R.; et~al. 2021.
\newblock Training verifiers to solve math word problems.
\newblock \emph{arXiv preprint arXiv:2110.14168}.

\bibitem[{Colon-Hernandez et~al.(2021)Colon-Hernandez, Havasi, Alonso, Huggins, and Breazeal}]{colon2021combining}
Colon-Hernandez, P.; Havasi, C.; Alonso, J.; Huggins, M.; and Breazeal, C. 2021.
\newblock Combining pre-trained language models and structured knowledge.
\newblock \emph{arXiv preprint arXiv:2101.12294}.

\bibitem[{Cui et~al.(2024)Cui, Li, Chen, Shou, and Chen}]{cui2024tabular}
Cui, L.; Li, H.; Chen, K.; Shou, L.; and Chen, G. 2024.
\newblock Tabular Data Augmentation for Machine Learning: Progress and Prospects of Embracing Generative AI.
\newblock \emph{arXiv preprint arXiv:2407.21523}.

\bibitem[{Deng et~al.(2024)Deng, Sun, He, Sikka, Chen, Ma, Zhang, and Mihalcea}]{deng2024tables}
Deng, N.; Sun, Z.; He, R.; Sikka, A.; Chen, Y.; Ma, L.; Zhang, Y.; and Mihalcea, R. 2024.
\newblock Tables as Images? Exploring the Strengths and Limitations of LLMs on Multimodal Representations of Tabular Data.
\newblock \emph{arXiv preprint arXiv:2402.12424}.

\bibitem[{Ding et~al.(2024)Ding, Fan, Ning, Wang, Li, Yin, Chua, and Li}]{ding2024survey}
Ding, Y.; Fan, W.; Ning, L.; Wang, S.; Li, H.; Yin, D.; Chua, T.-S.; and Li, Q. 2024.
\newblock A survey on rag meets llms: Towards retrieval-augmented large language models.
\newblock \emph{arXiv preprint arXiv:2405.06211}.

\bibitem[{Fang et~al.(2024)Fang, Xu, Tan, Zhang, Hu, Qi, Nickleach, Socolinsky, Sengamedu, Faloutsos et~al.}]{fang2024large}
Fang, X.; Xu, W.; Tan, F.~A.; Zhang, J.; Hu, Z.; Qi, Y.~J.; Nickleach, S.; Socolinsky, D.; Sengamedu, S.; Faloutsos, C.; et~al. 2024.
\newblock Large language models (LLMs) on tabular data: Prediction, generation, and understanding-a survey.

\bibitem[{Gardent et~al.(2017)Gardent, Shimorina, Narayan, and Perez-Beltrachini}]{gardent-etal-2017-webnlg}
Gardent, C.; Shimorina, A.; Narayan, S.; and Perez-Beltrachini, L. 2017.
\newblock The {W}eb{NLG} Challenge: Generating Text from {RDF} Data.
\newblock In Alonso, J.~M.; Bugar{\'\i}n, A.; and Reiter, E., eds., \emph{Proceedings of the 10th International Conference on Natural Language Generation}, 124--133. Santiago de Compostela, Spain: Association for Computational Linguistics.

\bibitem[{GLM et~al.(2024)GLM, Zeng, Xu, Wang, Zhang, Yin, Rojas, Feng, Zhao, Lai, Yu, Wang, Sun, Zhang, Cheng, Gui, Tang, Zhang, Li, Zhao, Wu, Zhong, Liu, Huang, Zhang, Zheng, Lu, Duan, Zhang, Cao, Yang, Tam, Zhao, Liu, Xia, Zhang, Gu, Lv, Liu, Liu, Yang, Song, Zhang, An, Xu, Niu, Yang, Li, Bai, Dong, Qi, Wang, Yang, Du, Hou, and Wang}]{glm2024chatglm}
GLM, T.; Zeng, A.; Xu, B.; Wang, B.; Zhang, C.; Yin, D.; Rojas, D.; Feng, G.; Zhao, H.; Lai, H.; Yu, H.; Wang, H.; Sun, J.; Zhang, J.; Cheng, J.; Gui, J.; Tang, J.; Zhang, J.; Li, J.; Zhao, L.; Wu, L.; Zhong, L.; Liu, M.; Huang, M.; Zhang, P.; Zheng, Q.; Lu, R.; Duan, S.; Zhang, S.; Cao, S.; Yang, S.; Tam, W.~L.; Zhao, W.; Liu, X.; Xia, X.; Zhang, X.; Gu, X.; Lv, X.; Liu, X.; Liu, X.; Yang, X.; Song, X.; Zhang, X.; An, Y.; Xu, Y.; Niu, Y.; Yang, Y.; Li, Y.; Bai, Y.; Dong, Y.; Qi, Z.; Wang, Z.; Yang, Z.; Du, Z.; Hou, Z.; and Wang, Z. 2024.
\newblock ChatGLM: A Family of Large Language Models from GLM-130B to GLM-4 All Tools.
\newblock arXiv:2406.12793.

\bibitem[{Guan, Huang, and Zhang(2024)}]{guan2024mfort}
Guan, C.; Huang, M.; and Zhang, P. 2024.
\newblock MFORT-QA: Multi-hop Few-shot Open Rich Table Question Answering.
\newblock \emph{arXiv preprint arXiv:2403.19116}.

\bibitem[{Gupta et~al.(2020)Gupta, Mehta, Nokhiz, and Srikumar}]{gupta-etal-2020-infotabs}
Gupta, V.; Mehta, M.; Nokhiz, P.; and Srikumar, V. 2020.
\newblock {INFOTABS}: Inference on Tables as Semi-structured Data.
\newblock In Jurafsky, D.; Chai, J.; Schluter, N.; and Tetreault, J., eds., \emph{Proceedings of the 58th Annual Meeting of the Association for Computational Linguistics}, 2309--2324. Online: Association for Computational Linguistics.

\bibitem[{Hendrycks et~al.(2021{\natexlab{a}})Hendrycks, Burns, Kadavath, Arora, Basart, Tang, Song, and Steinhardt}]{mmlu}
Hendrycks, D.; Burns, C.; Kadavath, S.; Arora, A.; Basart, S.; Tang, E.; Song, D.; and Steinhardt, J. 2021{\natexlab{a}}.
\newblock Measuring Mathematical Problem Solving With the MATH Dataset.
\newblock In \emph{Thirty-fifth Conference on Neural Information Processing Systems Datasets and Benchmarks Track (Round 2)}.

\bibitem[{Hendrycks et~al.(2021{\natexlab{b}})Hendrycks, Burns, Kadavath, Arora, Basart, Tang, Song, and Steinhardt}]{math}
Hendrycks, D.; Burns, C.; Kadavath, S.; Arora, A.; Basart, S.; Tang, E.; Song, D.; and Steinhardt, J. 2021{\natexlab{b}}.
\newblock Measuring Mathematical Problem Solving With the MATH Dataset.
\newblock In \emph{Thirty-fifth Conference on Neural Information Processing Systems Datasets and Benchmarks Track (Round 2)}.

\bibitem[{Herzig et~al.(2021)Herzig, M{\"u}ller, Krichene, and Eisenschlos}]{herzig2021open}
Herzig, J.; M{\"u}ller, T.; Krichene, S.; and Eisenschlos, J.~M. 2021.
\newblock Open domain question answering over tables via dense retrieval.
\newblock \emph{arXiv preprint arXiv:2103.12011}.

\bibitem[{Hu et~al.(2023)Hu, Chen, Li, Guo, Wen, Philip, and Guo}]{hu2023large}
Hu, X.; Chen, J.; Li, X.; Guo, Y.; Wen, L.; Philip, S.~Y.; and Guo, Z. 2023.
\newblock Do Large Language Models Know about Facts?
\newblock In \emph{The Twelfth International Conference on Learning Representations}.

\bibitem[{IBM(2024)}]{ibmstructured}
IBM. 2024.
\newblock Structured versus unstructured data: What's the difference?

\bibitem[{Iyyer, Yih, and Chang(2017)}]{sqa}
Iyyer, M.; Yih, W.-t.; and Chang, M.-W. 2017.
\newblock Search-based neural structured learning for sequential question answering.
\newblock In \emph{Proceedings of the 55th Annual Meeting of the Association for Computational Linguistics (Volume 1: Long Papers)}, 1821--1831.

\bibitem[{Kadavath et~al.(2022)Kadavath, Conerly, Askell, Henighan, Drain, Perez, Schiefer, Hatfield-Dodds, DasSarma, Tran-Johnson et~al.}]{kadavath2022language}
Kadavath, S.; Conerly, T.; Askell, A.; Henighan, T.; Drain, D.; Perez, E.; Schiefer, N.; Hatfield-Dodds, Z.; DasSarma, N.; Tran-Johnson, E.; et~al. 2022.
\newblock Language models (mostly) know what they know.
\newblock \emph{arXiv preprint arXiv:2207.05221}.

\bibitem[{Kim et~al.(2023)Kim, Park, Kwon, Jo, Thorne, and Choi}]{kim2023factkg}
Kim, J.; Park, S.; Kwon, Y.; Jo, Y.; Thorne, J.; and Choi, E. 2023.
\newblock FactKG: Fact Verification via Reasoning on Knowledge Graphs.
\newblock In \emph{Proceedings of the 61st Annual Meeting of the Association for Computational Linguistics (Volume 1: Long Papers)}, 16190--16206.

\bibitem[{Kojima et~al.(2022)Kojima, Gu, Reid, Matsuo, and Iwasawa}]{kojima2022large}
Kojima, T.; Gu, S.~S.; Reid, M.; Matsuo, Y.; and Iwasawa, Y. 2022.
\newblock Large language models are zero-shot reasoners.
\newblock \emph{Advances in neural information processing systems}, 35: 22199--22213.

\bibitem[{Kotonya and Toni(2020)}]{kotonya-toni-2020-explainable-automated}
Kotonya, N.; and Toni, F. 2020.
\newblock Explainable Automated Fact-Checking for Public Health Claims.
\newblock In Webber, B.; Cohn, T.; He, Y.; and Liu, Y., eds., \emph{Proceedings of the 2020 Conference on Empirical Methods in Natural Language Processing (EMNLP)}, 7740--7754. Online: Association for Computational Linguistics.

\bibitem[{Krichene et~al.(2024)Krichene, Piccinno, Liu, and Eisenschlos}]{krichene2024faithful}
Krichene, S.; Piccinno, F.; Liu, F.; and Eisenschlos, J.~M. 2024.
\newblock Faithful Chart Summarization with ChaTS-Pi.
\newblock \emph{arXiv e-prints}, arXiv--2405.

\bibitem[{Kweon et~al.(2023)Kweon, Kwon, Cho, Jo, and Choi}]{openwikitable}
Kweon, S.; Kwon, Y.; Cho, S.; Jo, Y.; and Choi, E. 2023.
\newblock Open-{W}iki{T}able : Dataset for Open Domain Question Answering with Complex Reasoning over Table.
\newblock In Rogers, A.; Boyd-Graber, J.; and Okazaki, N., eds., \emph{Findings of the Association for Computational Linguistics: ACL 2023}, 8285--8297. Toronto, Canada: Association for Computational Linguistics.

\bibitem[{Kwon et~al.(2023)Kwon, Li, Zhuang, Sheng, Zheng, Yu, Gonzalez, Zhang, and Stoica}]{kwon2023efficient}
Kwon, W.; Li, Z.; Zhuang, S.; Sheng, Y.; Zheng, L.; Yu, C.~H.; Gonzalez, J.~E.; Zhang, H.; and Stoica, I. 2023.
\newblock Efficient Memory Management for Large Language Model Serving with PagedAttention.
\newblock arXiv:2309.06180.

\bibitem[{Li et~al.(2024)Li, Su, Chen, Li, and ZHANG}]{li2024sheetcopilot}
Li, H.; Su, J.; Chen, Y.; Li, Q.; and ZHANG, Z.-X. 2024.
\newblock SheetCopilot: Bringing Software Productivity to the Next Level through Large Language Models.
\newblock \emph{Advances in Neural Information Processing Systems}, 36.

\bibitem[{Li, Sun, and Cheng(2021)}]{li2021tsqa}
Li, X.; Sun, Y.; and Cheng, G. 2021.
\newblock TSQA: tabular scenario based question answering.
\newblock In \emph{Proceedings of the AAAI Conference on Artificial Intelligence}, volume~35, 13297--13305.

\bibitem[{Li et~al.(2023)Li, Zhao, Chia, Ding, Bing, Joty, and Poria}]{li2023chain}
Li, X.; Zhao, R.; Chia, Y.~K.; Ding, B.; Bing, L.; Joty, S.; and Poria, S. 2023.
\newblock Chain of knowledge: A framework for grounding large language models with structured knowledge bases.
\newblock \emph{arXiv preprint arXiv:2305.13269}.

\bibitem[{Lin, Hilton, and Evans(2022)}]{lin2022teaching}
Lin, S.; Hilton, J.; and Evans, O. 2022.
\newblock Teaching models to express their uncertainty in words.
\newblock \emph{arXiv preprint arXiv:2205.14334}.

\bibitem[{Liu, Wang, and Chen(2023)}]{liu2023rethinking}
Liu, T.; Wang, F.; and Chen, M. 2023.
\newblock Rethinking Tabular Data Understanding with Large Language Models.
\newblock \emph{arXiv preprint arXiv:2312.16702}.

\bibitem[{Madaan et~al.(2024)Madaan, Tandon, Gupta, Hallinan, Gao, Wiegreffe, Alon, Dziri, Prabhumoye, Yang et~al.}]{selfrefine}
Madaan, A.; Tandon, N.; Gupta, P.; Hallinan, S.; Gao, L.; Wiegreffe, S.; Alon, U.; Dziri, N.; Prabhumoye, S.; Yang, Y.; et~al. 2024.
\newblock Self-refine: Iterative refinement with self-feedback.
\newblock \emph{Advances in Neural Information Processing Systems}, 36.

\bibitem[{Nan et~al.(2022)Nan, Hsieh, Mao, Lin, Verma, Zhang, Kry{\'s}ci{\'n}ski, Schoelkopf, Kong, Tang, Mutuma, Rosand, Trindade, Bandaru, Cunningham, Xiong, Radev, and Radev}]{fetaqa}
Nan, L.; Hsieh, C.; Mao, Z.; Lin, X.~V.; Verma, N.; Zhang, R.; Kry{\'s}ci{\'n}ski, W.; Schoelkopf, H.; Kong, R.; Tang, X.; Mutuma, M.; Rosand, B.; Trindade, I.; Bandaru, R.; Cunningham, J.; Xiong, C.; Radev, D.; and Radev, D. 2022.
\newblock {F}e{T}a{QA}: Free-form Table Question Answering.
\newblock \emph{Transactions of the Association for Computational Linguistics}, 10: 35--49.

\bibitem[{Nan et~al.(2021)Nan, Radev, Zhang, Rau, Sivaprasad, Hsieh, Tang, Vyas, Verma, Krishna, Liu, Irwanto, Pan, Rahman, Zaidi, Mutuma, Tarabar, Gupta, Yu, Tan, Lin, Xiong, Socher, and Rajani}]{nan-etal-2021-dart}
Nan, L.; Radev, D.; Zhang, R.; Rau, A.; Sivaprasad, A.; Hsieh, C.; Tang, X.; Vyas, A.; Verma, N.; Krishna, P.; Liu, Y.; Irwanto, N.; Pan, J.; Rahman, F.; Zaidi, A.; Mutuma, M.; Tarabar, Y.; Gupta, A.; Yu, T.; Tan, Y.~C.; Lin, X.~V.; Xiong, C.; Socher, R.; and Rajani, N.~F. 2021.
\newblock {DART}: Open-Domain Structured Data Record to Text Generation.
\newblock In Toutanova, K.; Rumshisky, A.; Zettlemoyer, L.; Hakkani-Tur, D.; Beltagy, I.; Bethard, S.; Cotterell, R.; Chakraborty, T.; and Zhou, Y., eds., \emph{Proceedings of the 2021 Conference of the North American Chapter of the Association for Computational Linguistics: Human Language Technologies}, 432--447. Online: Association for Computational Linguistics.

\bibitem[{OpenAI(2024)}]{gpt4ominiurl}
OpenAI. 2024.
\newblock GPT-4o mini: advancing cost-efficient intelligence.

\bibitem[{Parikh et~al.(2020)Parikh, Wang, Gehrmann, Faruqui, Dhingra, Yang, and Das}]{totto}
Parikh, A.; Wang, X.; Gehrmann, S.; Faruqui, M.; Dhingra, B.; Yang, D.; and Das, D. 2020.
\newblock {ToTTo}: A Controlled Table-To-Text Generation Dataset.
\newblock In Webber, B.; Cohn, T.; He, Y.; and Liu, Y., eds., \emph{Proceedings of the 2020 Conference on Empirical Methods in Natural Language Processing (EMNLP)}, 1173--1186. Online: Association for Computational Linguistics.

\bibitem[{Pasupat and Liang(2015)}]{wikitablequestion}
Pasupat, P.; and Liang, P. 2015.
\newblock Compositional Semantic Parsing on Semi-Structured Tables.
\newblock In \emph{Proceedings of the 53rd Annual Meeting of the Association for Computational Linguistics and the 7th International Joint Conference on Natural Language Processing (Volume 1: Long Papers)}, 1470--1480.

\bibitem[{Pezeshkpour and Hruschka(2024)}]{pezeshkpour-hruschka-2024-large}
Pezeshkpour, P.; and Hruschka, E. 2024.
\newblock Large Language Models Sensitivity to The Order of Options in Multiple-Choice Questions.
\newblock In Duh, K.; Gomez, H.; and Bethard, S., eds., \emph{Findings of the Association for Computational Linguistics: NAACL 2024}, 2006--2017. Mexico City, Mexico: Association for Computational Linguistics.

\bibitem[{Qi et~al.(2022)Qi, Deng, Zhu, Lee, Witbrock, and Liu}]{qi-etal-2022-takg}
Qi, Q.; Deng, Z.; Zhu, Y.; Lee, L.~J.; Witbrock, M.; and Liu, J. 2022.
\newblock {T}a{KG}: A New Dataset for Paragraph-level Table-to-Text Generation Enhanced with Knowledge Graphs.
\newblock In He, Y.; Ji, H.; Li, S.; Liu, Y.; and Chang, C.-H., eds., \emph{Findings of the Association for Computational Linguistics: AACL-IJCNLP 2022}, 176--187. Online only: Association for Computational Linguistics.

\bibitem[{Singha et~al.(2023)Singha, Cambronero, Gulwani, Le, and Parnin}]{singha2023tabular}
Singha, A.; Cambronero, J.; Gulwani, S.; Le, V.; and Parnin, C. 2023.
\newblock Tabular representation, noisy operators, and impacts on table structure understanding tasks in LLMs.
\newblock \emph{arXiv preprint arXiv:2310.10358}.

\bibitem[{Slack and Singh(2023)}]{slack2023tablet}
Slack, D.; and Singh, S. 2023.
\newblock Tablet: Learning from instructions for tabular data.
\newblock \emph{arXiv preprint arXiv:2304.13188}.

\bibitem[{Sui et~al.(2024)Sui, Zhou, Zhou, Han, and Zhang}]{tablemeetsllm}
Sui, Y.; Zhou, M.; Zhou, M.; Han, S.; and Zhang, D. 2024.
\newblock Table Meets LLM: Can Large Language Models Understand Structured Table Data? A Benchmark and Empirical Study.
\newblock In \emph{Proceedings of the 17th ACM International Conference on Web Search and Data Mining}, WSDM '24, 645–654. New York, NY, USA: Association for Computing Machinery.
\newblock ISBN 9798400703713.

\bibitem[{Team et~al.(2024)Team, Mesnard, Hardin, Dadashi, Bhupatiraju, Pathak, Sifre, Rivière, Kale, Love, Tafti, Hussenot, Sessa, Chowdhery, Roberts, Barua, Botev, Castro-Ros, Slone, Héliou, Tacchetti, Bulanova, Paterson, Tsai, Shahriari, Lan, Choquette-Choo, Crepy, Cer, Ippolito, Reid, Buchatskaya, Ni, Noland, Yan, Tucker, Muraru, Rozhdestvenskiy, Michalewski, Tenney, Grishchenko, Austin, Keeling, Labanowski, Lespiau, Stanway, Brennan, Chen, Ferret, Chiu, Mao-Jones, Lee, Yu, Millican, Sjoesund, Lee, Dixon, Reid, Mikuła, Wirth, Sharman, Chinaev, Thain, Bachem, Chang, Wahltinez, Bailey, Michel, Yotov, Chaabouni, Comanescu, Jana, Anil, McIlroy, Liu, Mullins, Smith, Borgeaud, Girgin, Douglas, Pandya, Shakeri, De, Klimenko, Hennigan, Feinberg, Stokowiec, hui Chen, Ahmed, Gong, Warkentin, Peran, Giang, Farabet, Vinyals, Dean, Kavukcuoglu, Hassabis, Ghahramani, Eck, Barral, Pereira, Collins, Joulin, Fiedel, Senter, Andreev, and Kenealy}]{gemmateam2024gemmaopenmodelsbased}
Team, G.; Mesnard, T.; Hardin, C.; Dadashi, R.; Bhupatiraju, S.; Pathak, S.; Sifre, L.; Rivière, M.; Kale, M.~S.; Love, J.; Tafti, P.; Hussenot, L.; Sessa, P.~G.; Chowdhery, A.; Roberts, A.; Barua, A.; Botev, A.; Castro-Ros, A.; Slone, A.; Héliou, A.; Tacchetti, A.; Bulanova, A.; Paterson, A.; Tsai, B.; Shahriari, B.; Lan, C.~L.; Choquette-Choo, C.~A.; Crepy, C.; Cer, D.; Ippolito, D.; Reid, D.; Buchatskaya, E.; Ni, E.; Noland, E.; Yan, G.; Tucker, G.; Muraru, G.-C.; Rozhdestvenskiy, G.; Michalewski, H.; Tenney, I.; Grishchenko, I.; Austin, J.; Keeling, J.; Labanowski, J.; Lespiau, J.-B.; Stanway, J.; Brennan, J.; Chen, J.; Ferret, J.; Chiu, J.; Mao-Jones, J.; Lee, K.; Yu, K.; Millican, K.; Sjoesund, L.~L.; Lee, L.; Dixon, L.; Reid, M.; Mikuła, M.; Wirth, M.; Sharman, M.; Chinaev, N.; Thain, N.; Bachem, O.; Chang, O.; Wahltinez, O.; Bailey, P.; Michel, P.; Yotov, P.; Chaabouni, R.; Comanescu, R.; Jana, R.; Anil, R.; McIlroy, R.; Liu, R.; Mullins, R.; Smith, S.~L.; Borgeaud, S.; Girgin, S.; Douglas, S.;
  Pandya, S.; Shakeri, S.; De, S.; Klimenko, T.; Hennigan, T.; Feinberg, V.; Stokowiec, W.; hui Chen, Y.; Ahmed, Z.; Gong, Z.; Warkentin, T.; Peran, L.; Giang, M.; Farabet, C.; Vinyals, O.; Dean, J.; Kavukcuoglu, K.; Hassabis, D.; Ghahramani, Z.; Eck, D.; Barral, J.; Pereira, F.; Collins, E.; Joulin, A.; Fiedel, N.; Senter, E.; Andreev, A.; and Kenealy, K. 2024.
\newblock Gemma: Open Models Based on Gemini Research and Technology.
\newblock arXiv:2403.08295.

\bibitem[{Tirumala et~al.(2022)Tirumala, Markosyan, Zettlemoyer, and Aghajanyan}]{tirumala2022memorization}
Tirumala, K.; Markosyan, A.; Zettlemoyer, L.; and Aghajanyan, A. 2022.
\newblock Memorization without overfitting: Analyzing the training dynamics of large language models.
\newblock \emph{Advances in Neural Information Processing Systems}, 35: 38274--38290.

\bibitem[{Touvron et~al.(2023)Touvron, Lavril, Izacard, Martinet, Lachaux, Lacroix, Rozière, Goyal, Hambro, Azhar, Rodriguez, Joulin, Grave, and Lample}]{touvron2023llamaopenefficientfoundation}
Touvron, H.; Lavril, T.; Izacard, G.; Martinet, X.; Lachaux, M.-A.; Lacroix, T.; Rozière, B.; Goyal, N.; Hambro, E.; Azhar, F.; Rodriguez, A.; Joulin, A.; Grave, E.; and Lample, G. 2023.
\newblock LLaMA: Open and Efficient Foundation Language Models.
\newblock arXiv:2302.13971.

\bibitem[{Vaswani et~al.(2017)Vaswani, Shazeer, Parmar, Uszkoreit, Jones, Gomez, Kaiser, and Polosukhin}]{vaswani2017attention}
Vaswani, A.; Shazeer, N.; Parmar, N.; Uszkoreit, J.; Jones, L.; Gomez, A.~N.; Kaiser, {\L}.; and Polosukhin, I. 2017.
\newblock Attention is all you need.
\newblock \emph{Advances in neural information processing systems}, 30.

\bibitem[{Wang et~al.(2023{\natexlab{a}})Wang, Liu, Yue, Tang, Zhang, Jiayang, Yao, Gao, Hu, Qi et~al.}]{wang2023survey}
Wang, C.; Liu, X.; Yue, Y.; Tang, X.; Zhang, T.; Jiayang, C.; Yao, Y.; Gao, W.; Hu, X.; Qi, Z.; et~al. 2023{\natexlab{a}}.
\newblock Survey on factuality in large language models: Knowledge, retrieval and domain-specificity.
\newblock \emph{arXiv preprint arXiv:2310.07521}.

\bibitem[{Wang et~al.(2021)Wang, Mahajan, Danilevsky, and Rosenthal}]{wang-etal-2021-semeval}
Wang, N. X.~R.; Mahajan, D.; Danilevsky, M.; and Rosenthal, S. 2021.
\newblock {S}em{E}val-2021 Task 9: Fact Verification and Evidence Finding for Tabular Data in Scientific Documents ({SEM}-{TAB}-{FACTS}).
\newblock In Palmer, A.; Schneider, N.; Schluter, N.; Emerson, G.; Herbelot, A.; and Zhu, X., eds., \emph{Proceedings of the 15th International Workshop on Semantic Evaluation (SemEval-2021)}, 317--326. Online: Association for Computational Linguistics.

\bibitem[{Wang et~al.(2023{\natexlab{b}})Wang, Wei, Schuurmans, Le, Chi, Narang, Chowdhery, and Zhou}]{selfconsis}
Wang, X.; Wei, J.; Schuurmans, D.; Le, Q.~V.; Chi, E.~H.; Narang, S.; Chowdhery, A.; and Zhou, D. 2023{\natexlab{b}}.
\newblock Self-Consistency Improves Chain of Thought Reasoning in Language Models.
\newblock In \emph{The Eleventh International Conference on Learning Representations}.

\bibitem[{Wang et~al.(2024)Wang, Zhang, Li, Eisenschlos, Perot, Wang, Miculicich, Fujii, Shang, Lee et~al.}]{wangchain}
Wang, Z.; Zhang, H.; Li, C.-L.; Eisenschlos, J.~M.; Perot, V.; Wang, Z.; Miculicich, L.; Fujii, Y.; Shang, J.; Lee, C.-Y.; et~al. 2024.
\newblock Chain-of-Table: Evolving Tables in the Reasoning Chain for Table Understanding.
\newblock In \emph{The Twelfth International Conference on Learning Representations}.

\bibitem[{Wei et~al.(2022)Wei, Wang, Schuurmans, Bosma, Xia, Chi, Le, Zhou et~al.}]{wei2022chain}
Wei, J.; Wang, X.; Schuurmans, D.; Bosma, M.; Xia, F.; Chi, E.; Le, Q.~V.; Zhou, D.; et~al. 2022.
\newblock Chain-of-thought prompting elicits reasoning in large language models.
\newblock \emph{Advances in neural information processing systems}, 35: 24824--24837.

\bibitem[{Yang et~al.(2024{\natexlab{a}})Yang, Yang, Hui, Zheng, Yu, Zhou, Li, Li, Liu, Huang, Dong, Wei, Lin, Tang, Wang, Yang, Tu, Zhang, Ma, Xu, Zhou, Bai, He, Lin, Dang, Lu, Chen, Yang, Li, Xue, Ni, Zhang, Wang, Peng, Men, Gao, Lin, Wang, Bai, Tan, Zhu, Li, Liu, Ge, Deng, Zhou, Ren, Zhang, Wei, Ren, Fan, Yao, Zhang, Wan, Chu, Liu, Cui, Zhang, and Fan}]{qwen2}
Yang, A.; Yang, B.; Hui, B.; Zheng, B.; Yu, B.; Zhou, C.; Li, C.; Li, C.; Liu, D.; Huang, F.; Dong, G.; Wei, H.; Lin, H.; Tang, J.; Wang, J.; Yang, J.; Tu, J.; Zhang, J.; Ma, J.; Xu, J.; Zhou, J.; Bai, J.; He, J.; Lin, J.; Dang, K.; Lu, K.; Chen, K.; Yang, K.; Li, M.; Xue, M.; Ni, N.; Zhang, P.; Wang, P.; Peng, R.; Men, R.; Gao, R.; Lin, R.; Wang, S.; Bai, S.; Tan, S.; Zhu, T.; Li, T.; Liu, T.; Ge, W.; Deng, X.; Zhou, X.; Ren, X.; Zhang, X.; Wei, X.; Ren, X.; Fan, Y.; Yao, Y.; Zhang, Y.; Wan, Y.; Chu, Y.; Liu, Y.; Cui, Z.; Zhang, Z.; and Fan, Z. 2024{\natexlab{a}}.
\newblock Qwen2 Technical Report.
\newblock \emph{arXiv preprint arXiv:2407.10671}.

\bibitem[{Yang et~al.(2024{\natexlab{b}})Yang, Yang, Hui, Zheng, Yu, Zhou, Li, Li, Liu, Huang, Dong, Wei, Lin, Tang, Wang, Yang, Tu, Zhang, Ma, Yang, Xu, Zhou, Bai, He, Lin, Dang, Lu, Chen, Yang, Li, Xue, Ni, Zhang, Wang, Peng, Men, Gao, Lin, Wang, Bai, Tan, Zhu, Li, Liu, Ge, Deng, Zhou, Ren, Zhang, Wei, Ren, Liu, Fan, Yao, Zhang, Wan, Chu, Liu, Cui, Zhang, Guo, and Fan}]{yang2024qwen2technicalreport}
Yang, A.; Yang, B.; Hui, B.; Zheng, B.; Yu, B.; Zhou, C.; Li, C.; Li, C.; Liu, D.; Huang, F.; Dong, G.; Wei, H.; Lin, H.; Tang, J.; Wang, J.; Yang, J.; Tu, J.; Zhang, J.; Ma, J.; Yang, J.; Xu, J.; Zhou, J.; Bai, J.; He, J.; Lin, J.; Dang, K.; Lu, K.; Chen, K.; Yang, K.; Li, M.; Xue, M.; Ni, N.; Zhang, P.; Wang, P.; Peng, R.; Men, R.; Gao, R.; Lin, R.; Wang, S.; Bai, S.; Tan, S.; Zhu, T.; Li, T.; Liu, T.; Ge, W.; Deng, X.; Zhou, X.; Ren, X.; Zhang, X.; Wei, X.; Ren, X.; Liu, X.; Fan, Y.; Yao, Y.; Zhang, Y.; Wan, Y.; Chu, Y.; Liu, Y.; Cui, Z.; Zhang, Z.; Guo, Z.; and Fan, Z. 2024{\natexlab{b}}.
\newblock Qwen2 Technical Report.
\newblock arXiv:2407.10671.

\bibitem[{Yang et~al.(2023)Yang, Chen, Wang, Hu, and Zhang}]{10.1145/3581783.3611964}
Yang, Q.; Chen, Q.; Wang, W.; Hu, B.; and Zhang, M. 2023.
\newblock Enhancing Multi-modal Multi-hop Question Answering via Structured Knowledge and Unified Retrieval-Generation.
\newblock In \emph{Proceedings of the 31st ACM International Conference on Multimedia}, MM '23, 5223–5234. New York, NY, USA: Association for Computing Machinery.
\newblock ISBN 9798400701085.

\bibitem[{Zhang et~al.(2024)Zhang, Jin, Zhu, Chen, Huang, Wang, Hua, Liang, and Chen}]{zhang2024trustuqa}
Zhang, W.; Jin, L.; Zhu, Y.; Chen, J.; Huang, Z.; Wang, J.; Hua, Y.; Liang, L.; and Chen, H. 2024.
\newblock TrustUQA: A Trustful Framework for Unified Structured Data Question Answering.
\newblock \emph{arXiv preprint arXiv:2406.18916}.

\bibitem[{Zhao et~al.(2023{\natexlab{a}})Zhao, Ji, Zhang, He, Wang, Wang, Feng, and Zhang}]{zhao2023large}
Zhao, B.; Ji, C.; Zhang, Y.; He, W.; Wang, Y.; Wang, Q.; Feng, R.; and Zhang, X. 2023{\natexlab{a}}.
\newblock Large Language Models are Complex Table Parsers.
\newblock In \emph{Proceedings of the 2023 Conference on Empirical Methods in Natural Language Processing}, 14786--14802.

\bibitem[{Zhao et~al.(2023{\natexlab{b}})Zhao, Zhang, Si, Nan, Tang, and Cohan}]{zhao2023investigating}
Zhao, Y.; Zhang, H.; Si, S.; Nan, L.; Tang, X.; and Cohan, A. 2023{\natexlab{b}}.
\newblock Investigating Table-to-Text Generation Capabilities of LLMs in Real-World Information Seeking Scenarios.
\newblock \emph{arXiv preprint arXiv:2305.14987}.

\bibitem[{Zheng et~al.(2024{\natexlab{a}})Zheng, Zhou, Meng, Zhou, and Huang}]{zheng2024large}
Zheng, C.; Zhou, H.; Meng, F.; Zhou, J.; and Huang, M. 2024{\natexlab{a}}.
\newblock Large Language Models Are Not Robust Multiple Choice Selectors.
\newblock In \emph{The Twelfth International Conference on Learning Representations}.

\bibitem[{Zheng et~al.(2024{\natexlab{b}})Zheng, Chiang, Sheng, Zhuang, Wu, Zhuang, Lin, Li, Li, Xing, Zhang, Gonzalez, and Stoica}]{judging2024zheng}
Zheng, L.; Chiang, W.-L.; Sheng, Y.; Zhuang, S.; Wu, Z.; Zhuang, Y.; Lin, Z.; Li, Z.; Li, D.; Xing, E.~P.; Zhang, H.; Gonzalez, J.~E.; and Stoica, I. 2024{\natexlab{b}}.
\newblock Judging LLM-as-a-judge with MT-bench and Chatbot Arena.
\newblock In \emph{Proceedings of the 37th International Conference on Neural Information Processing Systems}, NIPS '23. Red Hook, NY, USA: Curran Associates Inc.

\bibitem[{Zhu et~al.(2021)Zhu, Lei, Huang, Wang, Zhang, Lv, Feng, and Chua}]{zhu-etal-2021-tat}
Zhu, F.; Lei, W.; Huang, Y.; Wang, C.; Zhang, S.; Lv, J.; Feng, F.; and Chua, T.-S. 2021.
\newblock {TAT}-{QA}: A Question Answering Benchmark on a Hybrid of Tabular and Textual Content in Finance.
\newblock In Zong, C.; Xia, F.; Li, W.; and Navigli, R., eds., \emph{Proceedings of the 59th Annual Meeting of the Association for Computational Linguistics and the 11th International Joint Conference on Natural Language Processing (Volume 1: Long Papers)}, 3277--3287. Online: Association for Computational Linguistics.

\bibitem[{Zhuang et~al.(2024)Zhuang, Zhang, Zheng, Du, Wang, Ren, Huang, Fu, Yue, and Chen}]{zhuang2024structlm}
Zhuang, A.; Zhang, G.; Zheng, T.; Du, X.; Wang, J.; Ren, W.; Huang, S.~W.; Fu, J.; Yue, X.; and Chen, W. 2024.
\newblock StructLM: Towards Building Generalist Models for Structured Knowledge Grounding.
\newblock \emph{arXiv preprint arXiv:2402.16671}.

\end{thebibliography}

\appendix

\clearpage
\section*{Technical Appendix}
\section{Implementation Details}
\label{sec:implementation_details}

\label{evaluate-env-hyperparam}
We use 32GB memory with Ubuntu 20.04 LTS (a open-source Operating System using the Linux kernel and based on Debian) and 4 Nvidia A800 with 80GB memory for inference. we adopt vllm~\citep{kwon2023efficient} 0.5.4 to speed up inference. All models share a set of hyperparameters, as detailed in Table \ref{tab:model_hyperparams}.

\begin{table}[h]
    \centering
    \resizebox{0.48 \textwidth}{!}
    {    
        \begin{tabular}{|p{4cm}|p{4cm}|}
        \hline
        \textbf{Hyperparameter} & \textbf{Value} \\ \hline
        \texttt{top\_p} & 0.95 \\ \hline
        \texttt{temperature} & 0.6 \\ \hline
        \texttt{max\_generation\_token (w/o CoT)} & 10 \\ \hline
        \texttt{max\_generation\_token (w/ CoT)} & 512 \\ \hline
        \texttt{max\_evidence\_token} & 2500 \\ \hline
        \end{tabular}
    }
    \vspace{-3mm}
    \caption{Hyperparameters of LLMs}
    \vspace{-5mm}\label{tab:model_hyperparams}
\end{table}

\section{Evaluation Protocol}
\label{sec:apdx_metrics}
In this paper, we use six different metrics for evaluating the reasoning performance of LLMs on structured knowledge. We formulate all the evaluation metrics used in this section.

\begin{itemize}
    \item Accuracy.
    \begin{equation}
    \setlength{\abovedisplayskip}{2pt}
        Acc. = \frac{TP+TN}{TP+TN+FP+FN}
    \setlength{\belowdisplayskip}{2pt}
    \end{equation}
    where $TP$, $TN$, $FP$, $FN$ represent the number of true positive, true negative, false positive, and false negative, respectively.
    \item Weighted F1 score.
    \begin{equation}
    \setlength{\abovedisplayskip}{2pt}
        F1 = \sum_{i=1}^N \frac{n_i}{N} F1_i
    \setlength{\belowdisplayskip}{2pt}
    \end{equation}
    where $n_i$ is the number of samples in label $i$, $N$ is the number of all samples, $F1_i$ is the F1 score for label $i$.
    \item Balanced accuracy.
    \begin{equation}
    \setlength{\abovedisplayskip}{2pt}
        BA = \frac{1}{N}\sum_{i=1}^N (TPR_i), TPR = \frac{TP}{TP+FN}
    \setlength{\belowdisplayskip}{2pt}
    \end{equation}
    where $TPR_i$ is the true positive rate of label $i$.
    \item Macro F1 score.
    \begin{equation}
    \setlength{\abovedisplayskip}{2pt}
        Macro F1 = \frac{1}{N}\sum_{i=1}^N F1_i
    \setlength{\belowdisplayskip}{2pt}
    \end{equation}
    \item Precision.
    \begin{equation}
    \setlength{\abovedisplayskip}{2pt}
        Prec. = \frac{TP}{TP+FP}
    \setlength{\belowdisplayskip}{2pt}
    \end{equation}
    \item Recall.
    \begin{equation}
    \setlength{\abovedisplayskip}{2pt}
        Recall = \frac{TP}{TP+FN}
    \setlength{\belowdisplayskip}{2pt}
    \end{equation}
\end{itemize}

\section{Detailed Introduction to selected LLMs}
\label{sec:apdx_llms_intro}
Meta’s Llama series, including Llama 2 and Llama 3~\shortcite{touvron2023llamaopenefficientfoundation}, released in 2023 and 2024, are designed for various tasks like text generation and programming. Llama3 is designed to be more intelligent, faster, and more versatile, making it suitable for a wide range of applications.
Qwen2~\shortcite{qwen2}~\shortcite{yang2024qwen2technicalreport} is a strong language models developed by Alibaba Cloud, showing state-of-the-art performance in several benchmarks, especially in coding and mathematics.
ChatGLM3~\shortcite{glm2024chatglm} is the latest generation of pre-trained dialogue models developed by Zhipu AI in collaboration with Tsinghua University’s Knowledge Engineering Group (KEG).
Developed by OpenAI, GPT-4o-mini~\shortcite{gpt4ominiurl} is its most cost-efficient small model in the GPT series, featuring enhanced context understanding and text generation capabilities, scoring 82\% on MMLU~\cite{mmlu}. Gemma2~\shortcite{gemmateam2024gemmaopenmodelsbased} is Google’s latest iteration of open large language models (LLMs), building on the success of the original Gemma series. Coming with two sizes, 9 billion and 27 billion parameters, each size has a base model (pre-trained) and an instruction-tuned version.

\section{Ethical Statement}
\label{sec:apdx_ethic}
We affirm that our \name\ benchmark is constructed using open-source datasets and adheres to the CC-BY-4.0 license. To uphold privacy and confidentiality, we have ensured that our dataset contains no direct or indirect sensitive personal information. 
Users accessing our \name\ should ensure that no personally identifiable information or toxic content is included.

Our research postulate that our \name\ benchmark is under an environment devoid of possible attacks. However, given that the structured data in our proposed benchmark is sourced from publicly editable WikiPedia pages, it is inherently vulnerable to various threats, including adversarial attacks. Intended attacks, such as data poisoning, involve malicious actors deliberately inserting false or misleading information or altering existing structured data. These actions can compromise the integrity of the data, distorting the knowledge within LLMs and undermining accurate factual reasoning. Unintentional attacks, such as accidental data deletion or incorrect data entry, also pose significant risks. These errors can degrade both the quality and structure of the data, potentially leading LLMs to draw incorrect inferences, thus might compromising the overall factuality of the benchmark.

Moreover, while the questions in our \name\ benchmark reflect real-world facts, they do not originate from practical applications. Therefore, we offer \name\ as a resource to guide users in their inferences, without claiming to provide absolute assertions. We advise against using \name\ as a basis for developing models intended to verify facts in real-world applications. 

\section{Structural Datasets}
\label{sec:apdx_structural_datasets}
We conducted a thorough comparison of our proposed \name\ against a variety of public datasets that contain structural knowledge. Table \ref{tab:qa} provides a comprehensive comparison of these datasets, evaluating them across various dimensions, including tasks, sources, types of evidence, types of answer and knowledge domains. Additionally, we present the distribution of the five proposed factual tasks across these various public datasets in Table \ref{tab:qa_tasks}.

\section{Case Study}
\label{sec:apdx_case_study}
\begin{figure*}[!htbp]
  \centering
  \includegraphics[scale=0.6]{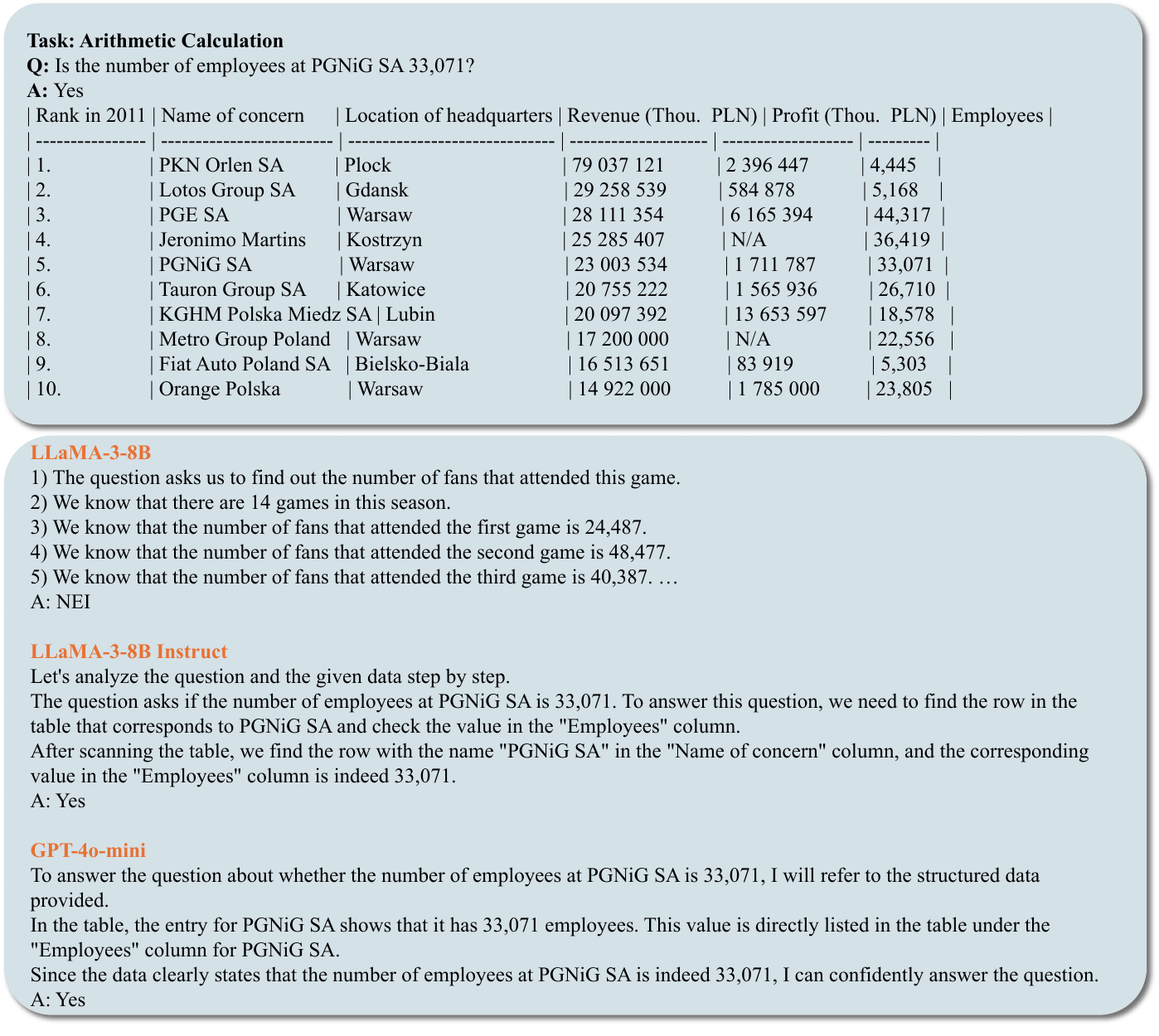}
  \caption{Case Study of Arithmetic Calculation.}
  \label{fig:case_arithmetic}
\end{figure*}

\begin{figure*}[!htbp]
  \centering
  \includegraphics[scale=0.6]{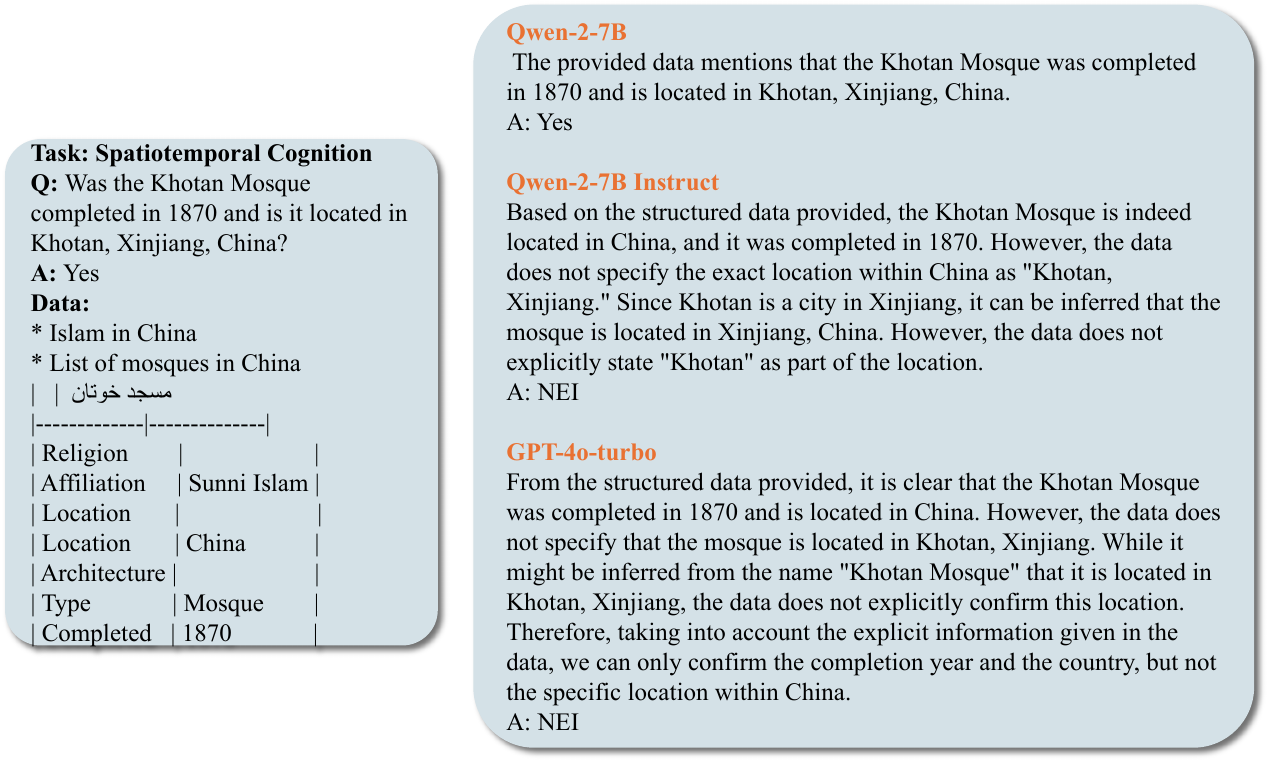}
  \caption{Case Study of Spatiotemporal Cognition.}
  \label{fig:case_spatiotemporal}
\end{figure*}

\begin{figure*}[!htbp]
  \centering
    \rotatebox{90}{\includegraphics[scale=0.6]{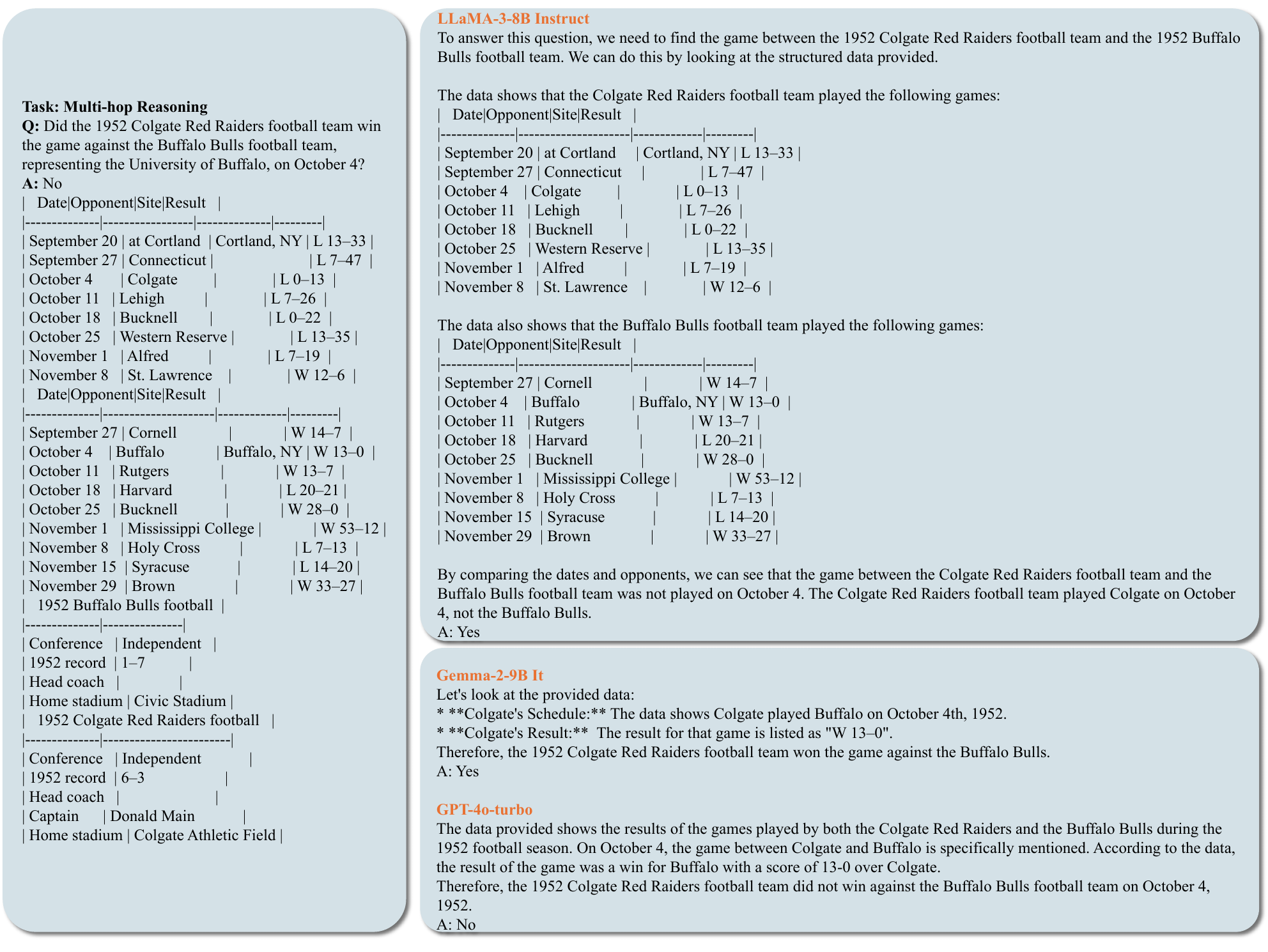}}
  \caption{Case Study of Multi-hop Reasoning.}
  \label{fig:case_multihop}
\end{figure*}

\begin{figure*}[!htbp]
  \centering
  \includegraphics[scale=0.6]{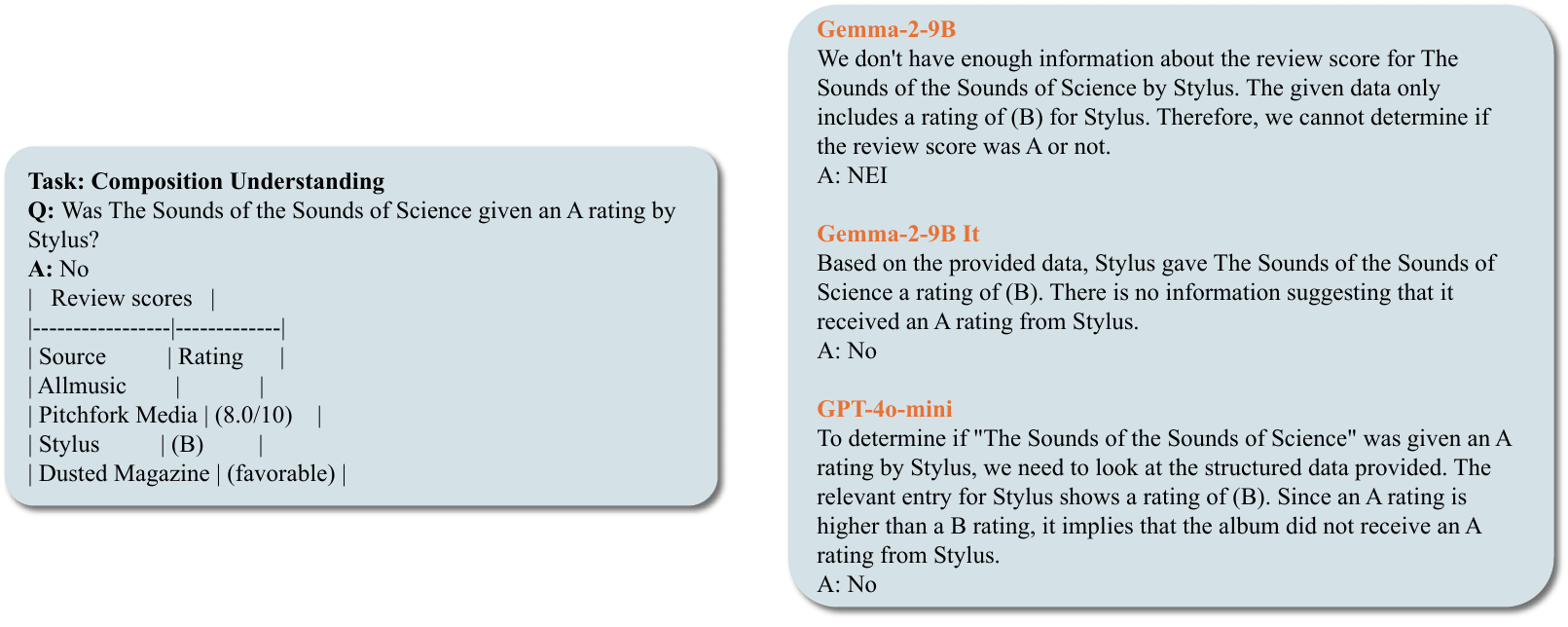}
  \caption{Case Study of Composition Understanding.}
  \label{fig:case_comp}
\end{figure*}

\begin{figure*}[!htbp]
  \centering
  \includegraphics[scale=0.6]{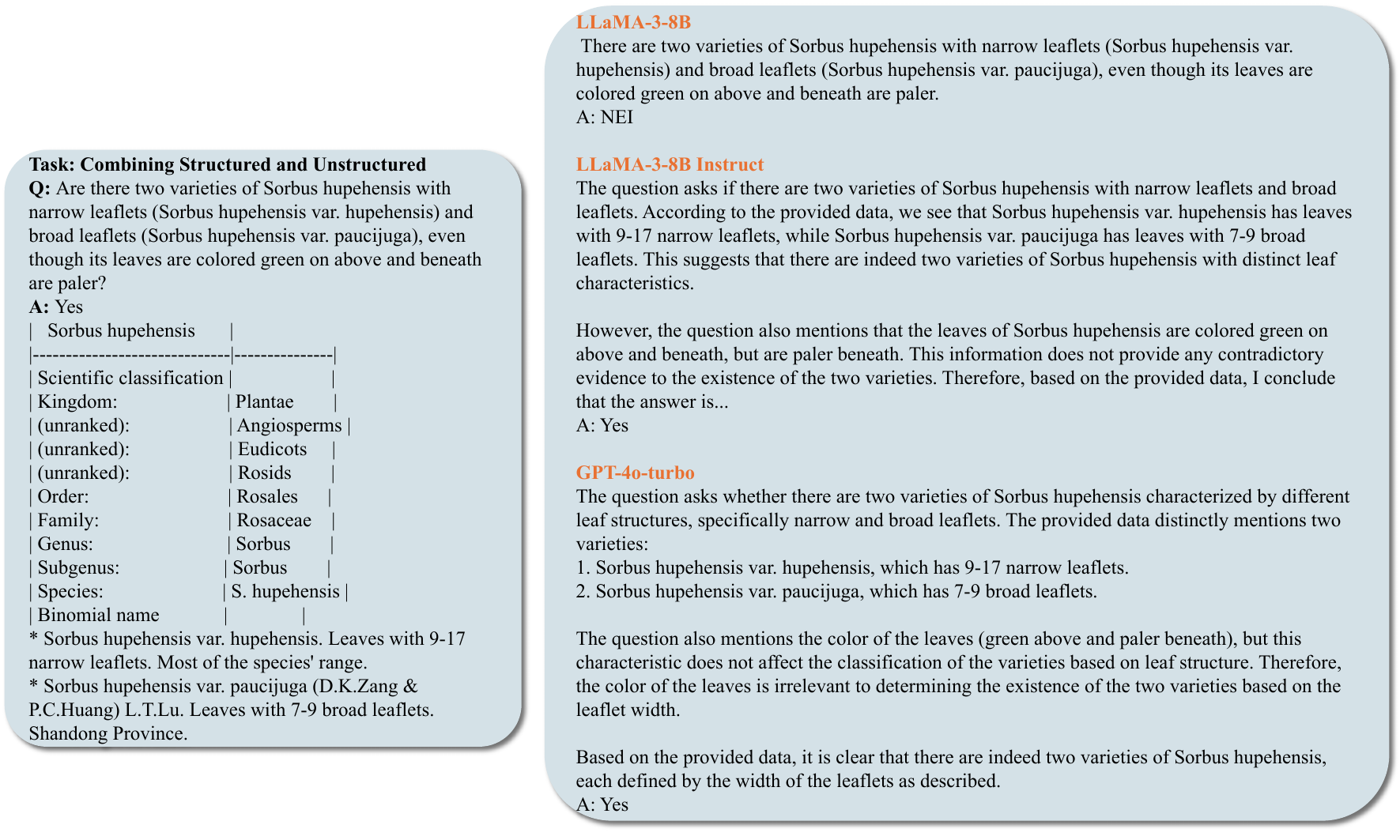}
  \caption{Case Study of Combining Structured and Unstructured.}
  \label{fig:case_combine}
\end{figure*}

Please see \cref{fig:case_arithmetic,fig:case_spatiotemporal,fig:case_multihop,fig:case_comp,fig:case_combine} for case studies for each task and the responses from different LLMs. 

\section{Prompt Strategies Analysis}
\subsection{Detailed Introduction to Employed Prompts}
\label{sec:apdx_prompts}
\begin{figure*}[!htbp]
  \centering
  \includegraphics[scale=0.5]{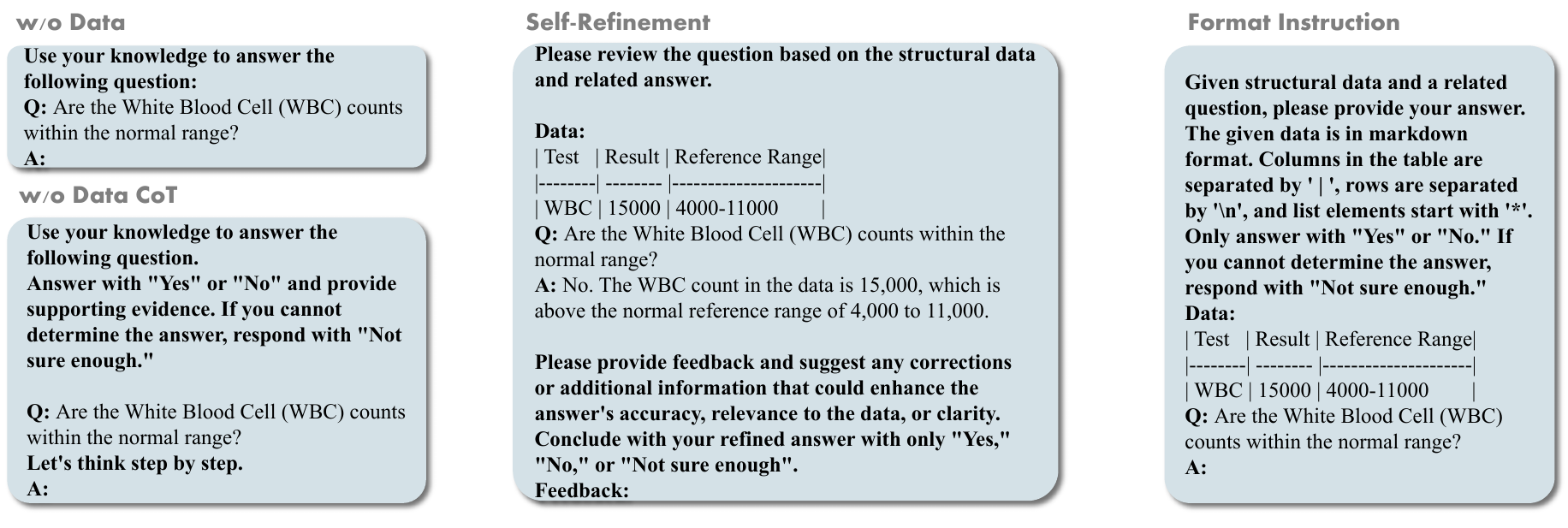}
  \caption{Examples of different prompting strategies.}
  \label{fig:prompts_other}
\end{figure*}

Each LLM in our main result decipted in Table \ref{tab:main_res} is experimented with different prompting strategies: Zero-shot without CoT ~\cite{kojima2022large}, Zero-shot with CoT, Few-shot with CoT, Few-shot with CoT. All the strategies used in this paper begin with an instruction denoted as $p=$ ``\textit{You will be given with a question. Please response with `Yes', `No', or `Not Sure Enough'.}'' For any input question $q_i\in\mathcal{Q}$, structural data $d_i\in\mathcal{D}$ the model $LLM(\cdot)$ is expected to generate an answer $y_i\in\mathcal{Y}=\{\texttt{`Yes',`No',`Not Sure Enough'}\}$. Each question is categorized into one task $t$ from the five aforementioned reasoning tasks in $\mathcal{T}$. Examples of the prompts used in our experiments are shown in Figure \ref{fig:prompts_other}.

\subsubsection{Prompts in Main Results}  
\textit{Prompt with Zero-shot.} In the prompting strategy with zero-shot setting, the LLM is expected to output the answer $y_i$ to the question $q_i$ directly, formally, $y_i=LLM(p, q_i, d_i)$. For example, the factual answer $y_i=\texttt{"No"}$ should be responded from the LLMs when being asked with the question $q_i=\texttt{"Is London the host city of the 2024 Oly-}\\\texttt{-mpic Games?"}$, together with the table of Olympic Games host cities denoted by $d_i$.

\textit{Prompt with Few-shot.} In the few-shot prompting strategy, to guide the LLM to correctly reason, we include an example question $q_x$ and structural data $d_x$ together with prompt $p$ for question $q_i$, where the example question $q_x$ and question $q_i$ fall in the same task, i.e., $q_x, q_i\in t$. This process is formulated as $y_i=LLM(p\Vert q_x\Vert d_x, q_i, d_i)$. The LLM is expected to answer with $y_i=\texttt{"Yes"}$ when given question $q_i=\texttt{"Has Paris hosted the Olympic Games three}\\\texttt{times?"}$ and the table of Olympic Games host cities $d_i$.

\textit{Prompt with Chain of Thought (CoT).} In the prompting strategy with CoT~\cite{kojima2022large}, a two-stage prompt is employed to derive the reasoning process along with with the answer. To guide the LLM in carefully considering the process of determining the answer $y_i$, the prompting sentence $s=\texttt{"Let's think step by step"}$ is added to the question $q_i$, formally, $y_i =LLM(p, q_i\Vert s, d_i)$.

\subsubsection{Prompts in Evidence Resilience Analysis} \textit{Prompt with Shuffled Structured Data.} To investigate the performances of LLMs towards different prompting context, we shuffle the structure of data. Specifically, we shuffle the rows/columns in tables, and the elements in lists. Formally, for question $q_i$, the output can be presented as $y_i=LLM(p, q_i, d_i')$, where $d_i'$ denotes the shuffled data.

\textit{Prompt without Structured Data.} Given that the structural data is sourced from Wikipedia, it is assumed that LLMs have been exposed to these data during their training phase. Therefore, we are also interested in the ability of LLMs to answer factual questions $q_i$ without being provided with the contextual structural data $d_i$. The process under this strategy can be formulated as $f_5:y_i=LLM(p, q_i)$.

\textit{Prompt with self-refinement.} The self-refinement strategy is designed to enhance the performance of LLMs by prompting them to iteratively providing feedback to its previous responses. Formally, the process at $n$-th round of refinement can be presented as $y_i^n=LLM^n(p, q_i, d_i, r_i^{n-1})$, where $r_i$ represents the LLM's response in the last round. In our experiments, due to constraints on computing resources and time, we set $n$=1.  

\textit{Prompt with self-consistency.} The self-consistency strategy is designed to enhance the performance of LLMs by employing majority voting on multiple rounds of queries. Assume the response from the model at the $n$-th round as $y_n$, the final prediction of LLM can be formualted as $y_{final} = argmax_{c_j}\sum_{j=1}^k counts(y_n=c_j)$, where $c$ denotes the available choices of the prediction label, i.e., `Fact.', `Non-Fact.', and `NEI' in this paper.

\textit{Prompt with format instructions.} We also provide instructions of the formats of the structured data to the zero-shot prompts. Given format instructions as $f$, which illustrates how the structured data looks like, the process can be formulated as $y_i=LLM(p\Vert f, p_i, d_i)$.

\subsection{Analysis towards Other Prompting Strategies}
Given the successes of other CoT strategies and input data format instructions~\shortcite{slack2023tablet}, we are interested in exploring their impact on reasoning about factual knowledge within structured data. We include three prompting strategies: (i) self-refinement~\shortcite{selfrefine}, which guides the LLM to iteratively evaluate and refine its previous responses to reach the correct answer, (ii) self-consistency~\shortcite{selfconsis}, which mitigates hallucination through majority voting on multiple responses from the LLM, and (iii) format instructions, which prompts with descriptions of the format of the inputted structured data. There are the following notable observations from the results in Table ~\ref{aptab:other_prompts}. i) Self-consistency marginally improves performance across five tasks, with an overall enhancement of 0.23\%, compared to the zero-shot results without CoT in Table \ref{tab:main_res}. ii) Format descriptions help the LLM better interpret numerical compositions, leading to a 1.02\% improvement in accuracy on Arithmetic Calculation tasks. Detailed results under these strategies please refer to Table \ref{aptab:other_prompts}.

\section{Supplementary Results}
\label{sec:apdx_sup_res}

\subsection{Comprehensive Results}
\noindent
Results for Different Prompts under Other Metrics:Please refer to Tables \ref{aptab:main_res_balanced} and \ref{aptab:main_res_pr}.

\noindent
Results for Different Tasks under Other Metrics: Please refer to Tables \ref{aptab:task_pr_zero_shot_wo_cot} to \ref{aptab:task_balanced_few_shot_w_cot}.

\subsection{Analysis of Other LLMs}

\begin{figure*}[ht]
  \centering
  \begin{minipage}[t]{0.45\linewidth} %
    \subfigure[GPT-4O Mini]{
        \includegraphics[width=\linewidth]{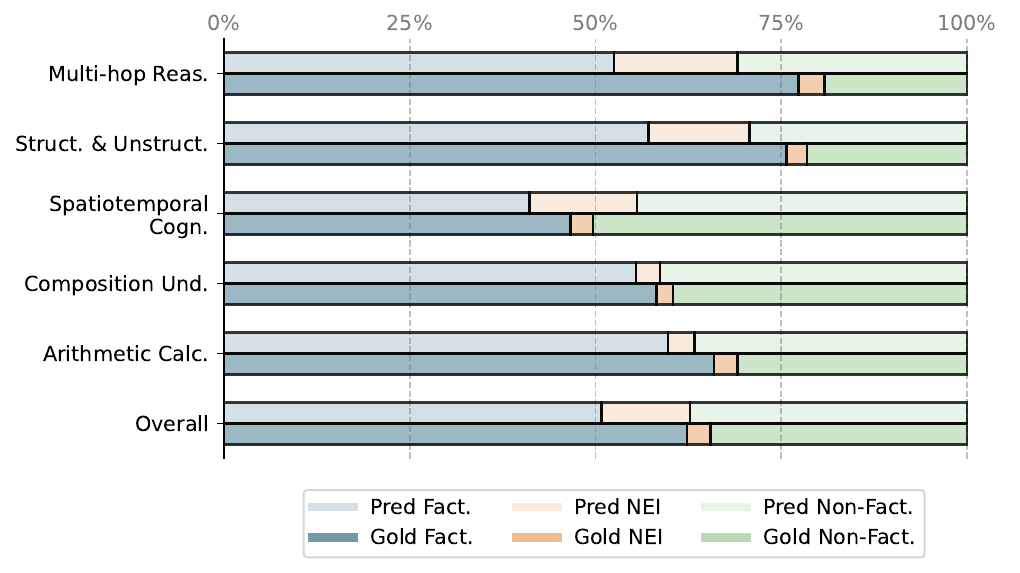}
    }
  \end{minipage}
  \begin{minipage}[t]{0.45\linewidth} %
    \subfigure[Llama3-8B Instruct]{
        \includegraphics[width=\linewidth]{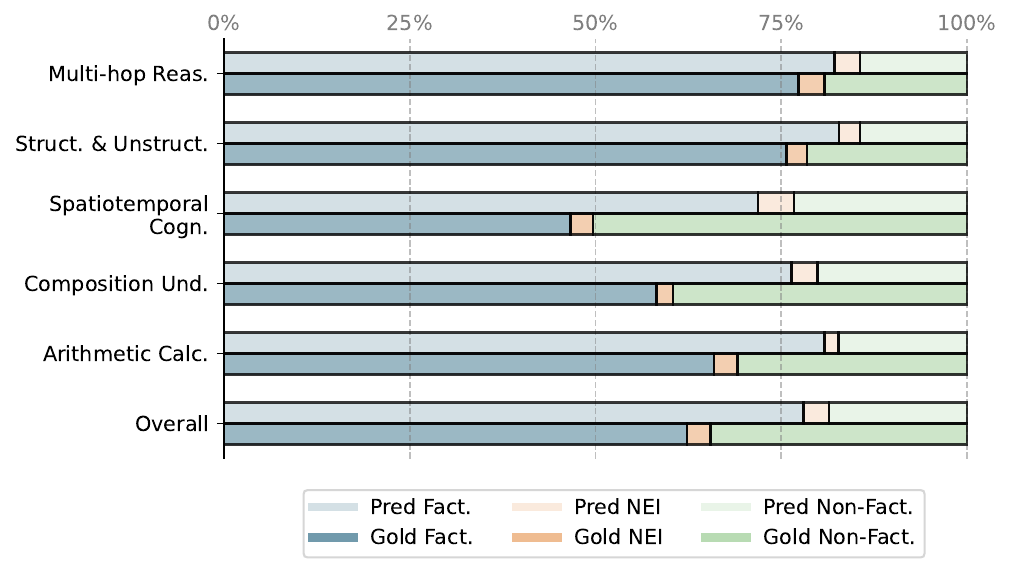}
    }
  \end{minipage}
  \begin{minipage}[t]{0.45\linewidth} %
    \subfigure[Gemma2-9B It]{
        \includegraphics[width=\linewidth]{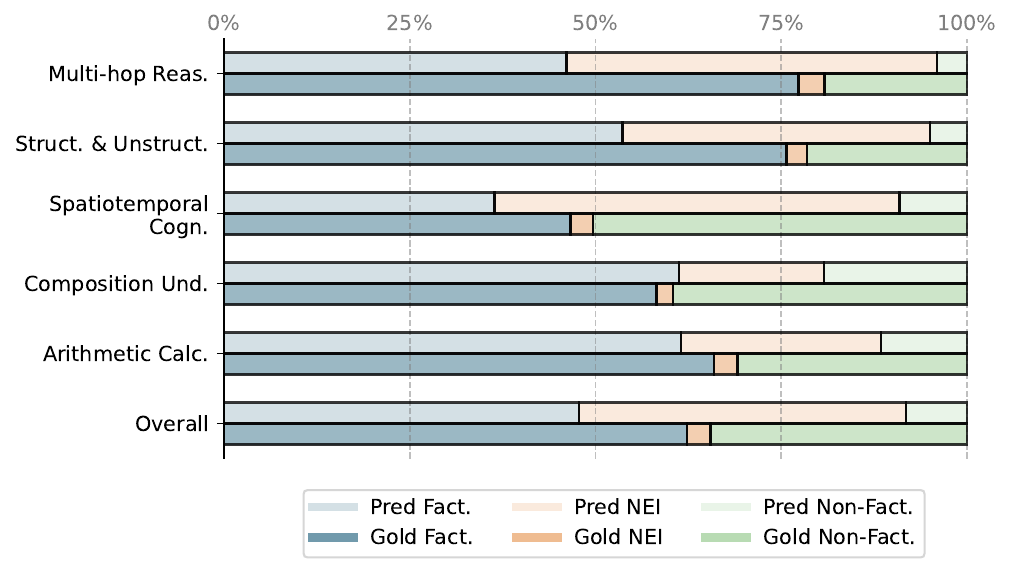}
    }
  \end{minipage}
  \begin{minipage}[t]{0.45\linewidth} %
    \subfigure[Qwen2-7B-Instruct]{
        \includegraphics[width=\linewidth]{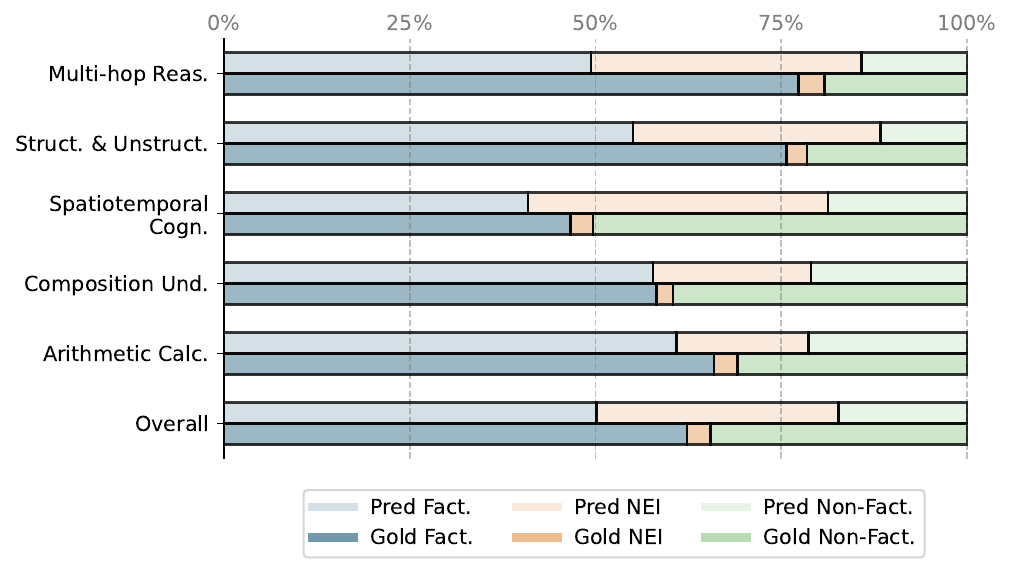}
    }  
  \end{minipage}
  \caption{Responses Distributions of Different Models.}
  \label{apfig:distribution}
\end{figure*}

\noindent
Model Responses Distributions: Please refer to Figure \ref{apfig:distribution}.

\begin{figure*}[!htp]
  \centering
  \vspace{-12mm}
  \includegraphics[scale=0.5]{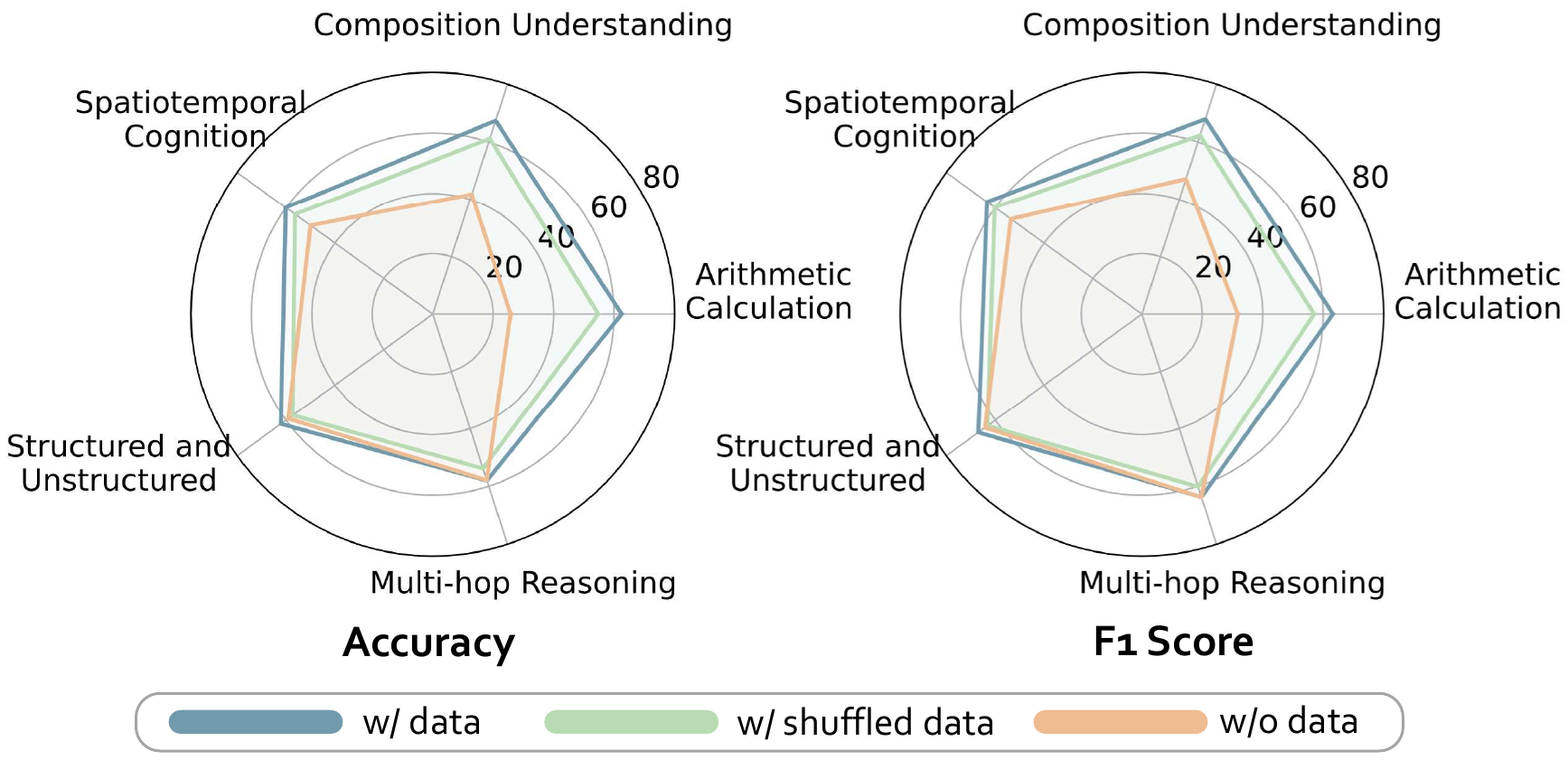}
  \vspace{-20mm}
  \caption{Accuracy and F1 score of GPT-4o-mini under different settings of structured evidence.}
  \label{apfig:radar}
\end{figure*}

\noindent
Model Resilience to Evidence: Please refer to Table \ref{aptab:radar} and Figure \ref{apfig:radar}.

\begin{figure*}[!t]
  \centering
  \begin{minipage}[t]{0.3\linewidth} %
    \subfigure[Llama3-8B Instruct]{
        \includegraphics[width=\linewidth]{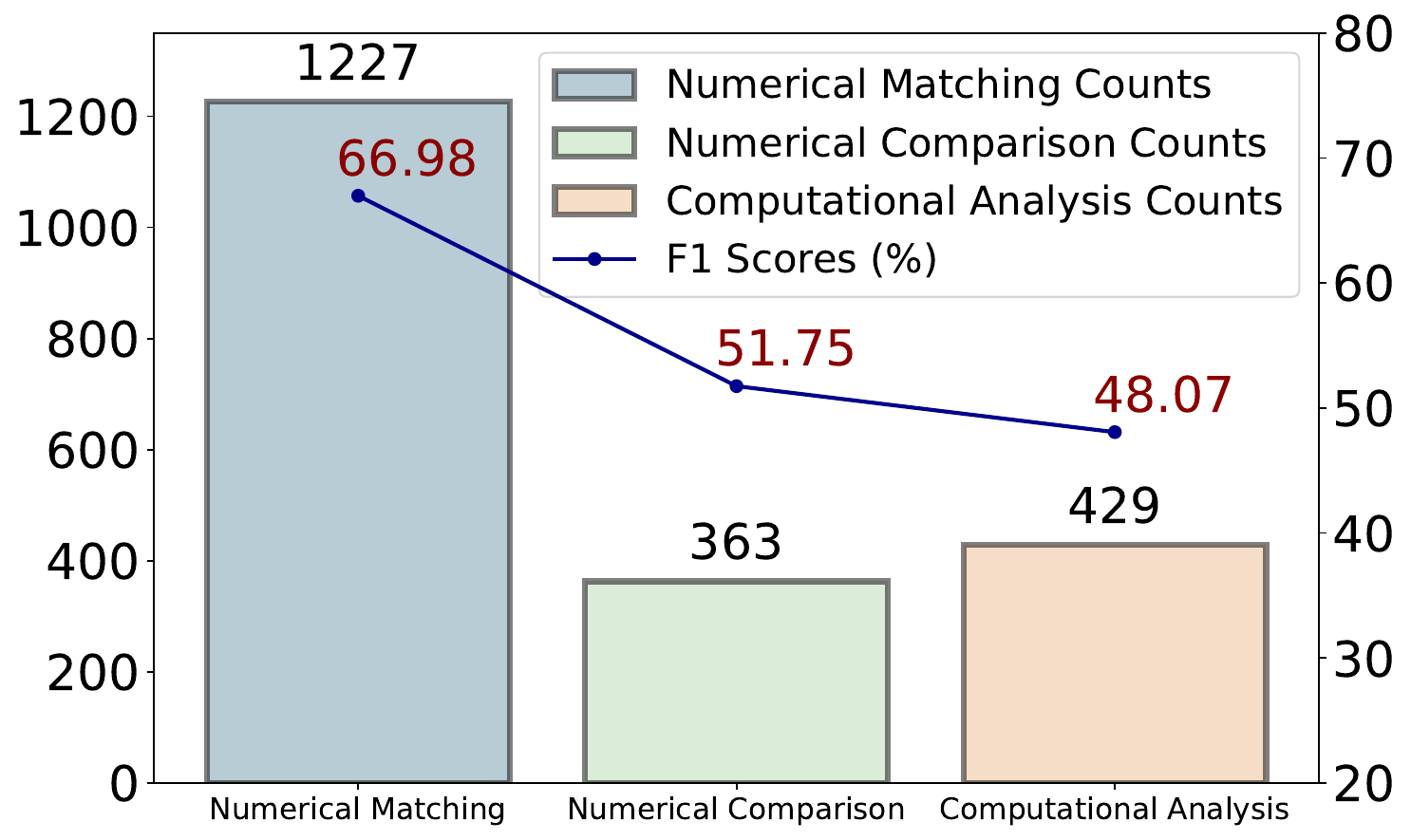}
    }
  \end{minipage}
  \begin{minipage}[t]{0.3\linewidth} %
    \subfigure[Gemma2-9B It]{
        \includegraphics[width=\linewidth]{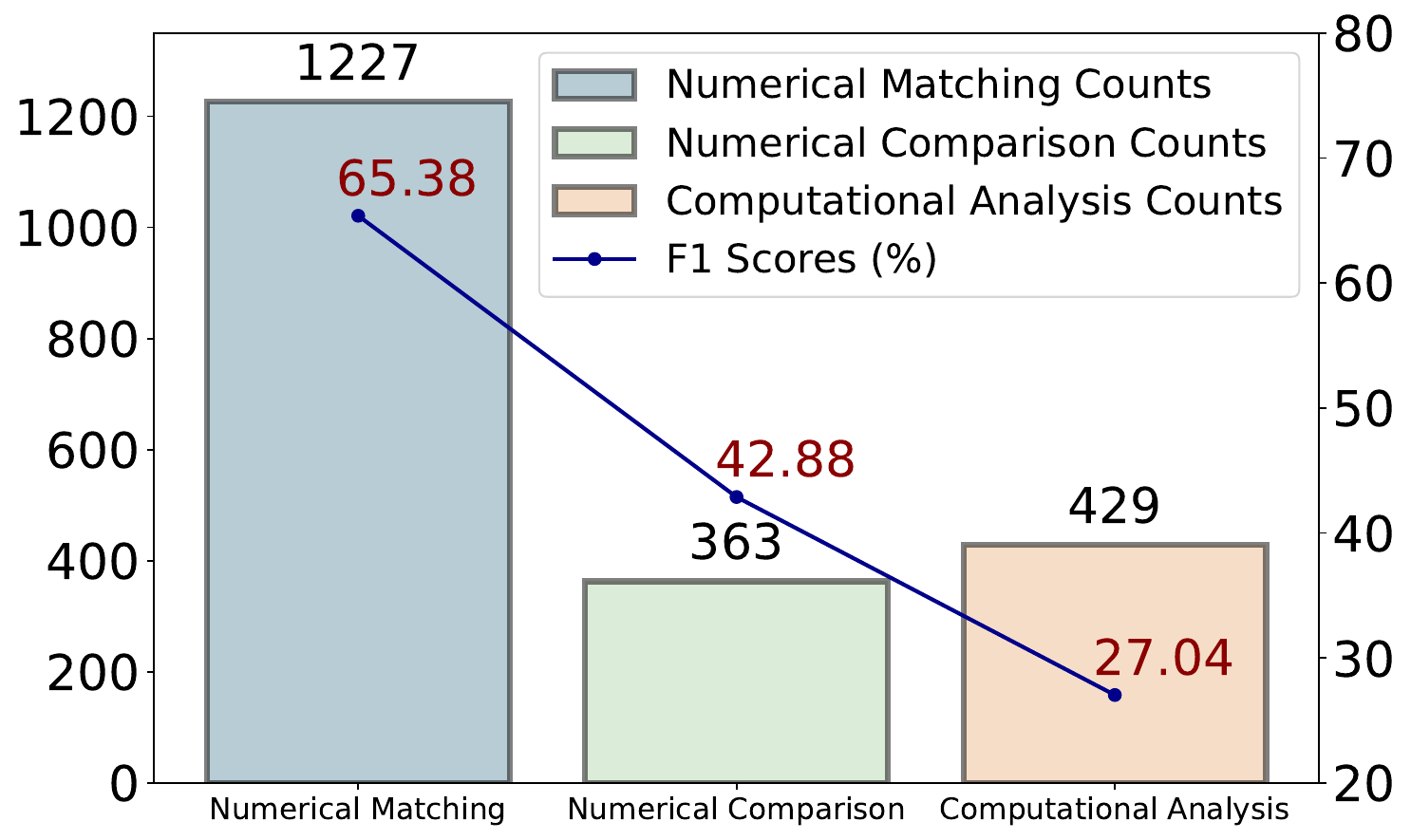}
    }  
  \end{minipage}
  \begin{minipage}[t]{0.3\linewidth} %
    \subfigure[Qwen2-7B-Instruct]{
        \includegraphics[width=\linewidth]{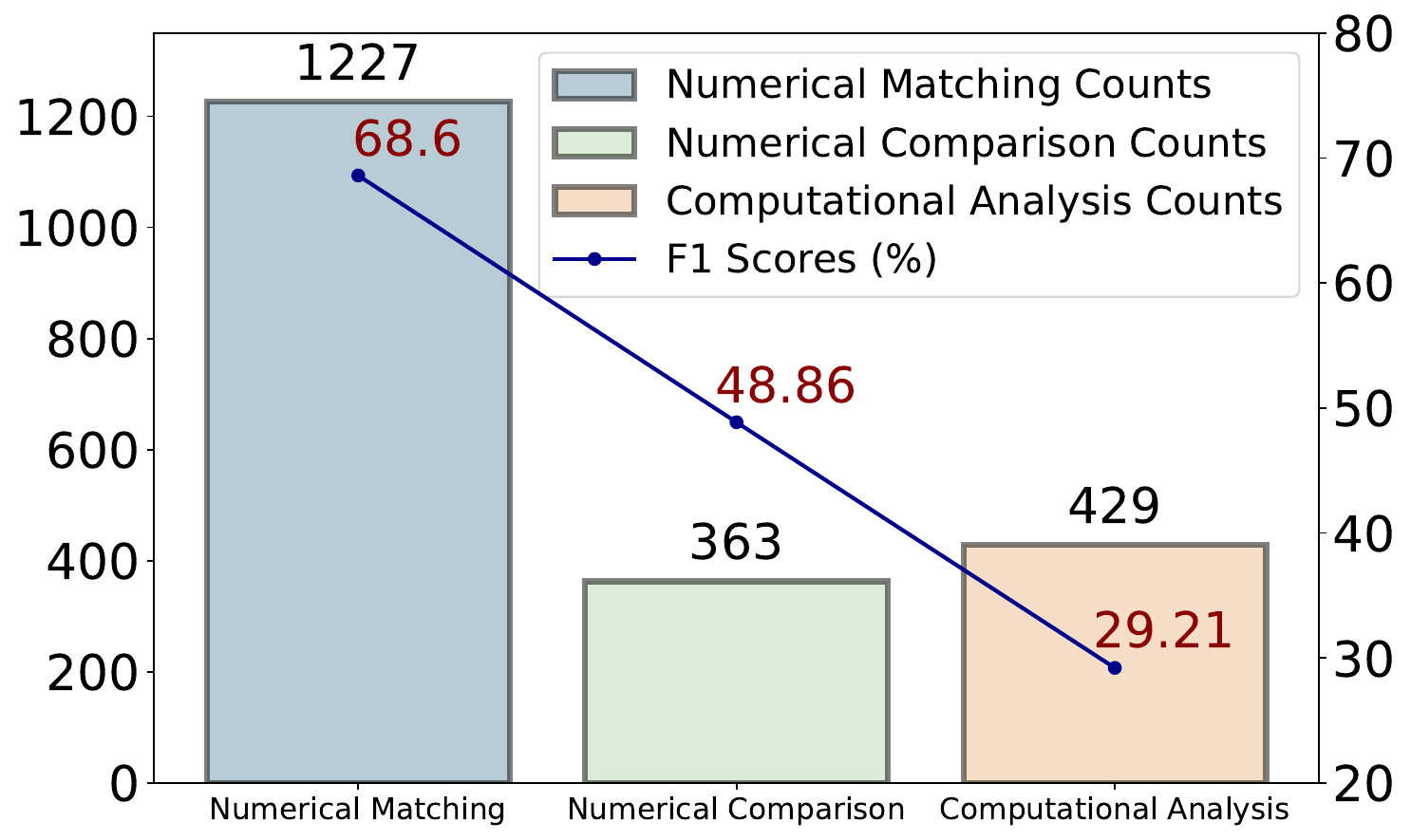}
    }  
  \end{minipage}
  \caption{Performance of Different Models on Arithmetic Calculation.}
  \label{apfig:arithmetic}
\end{figure*}

\begin{figure*}[!t]
  \centering
  \begin{minipage}[t]{0.75\linewidth} %
    \subfigure[Llama3-8B Instruct]{
        \includegraphics[width=\linewidth]{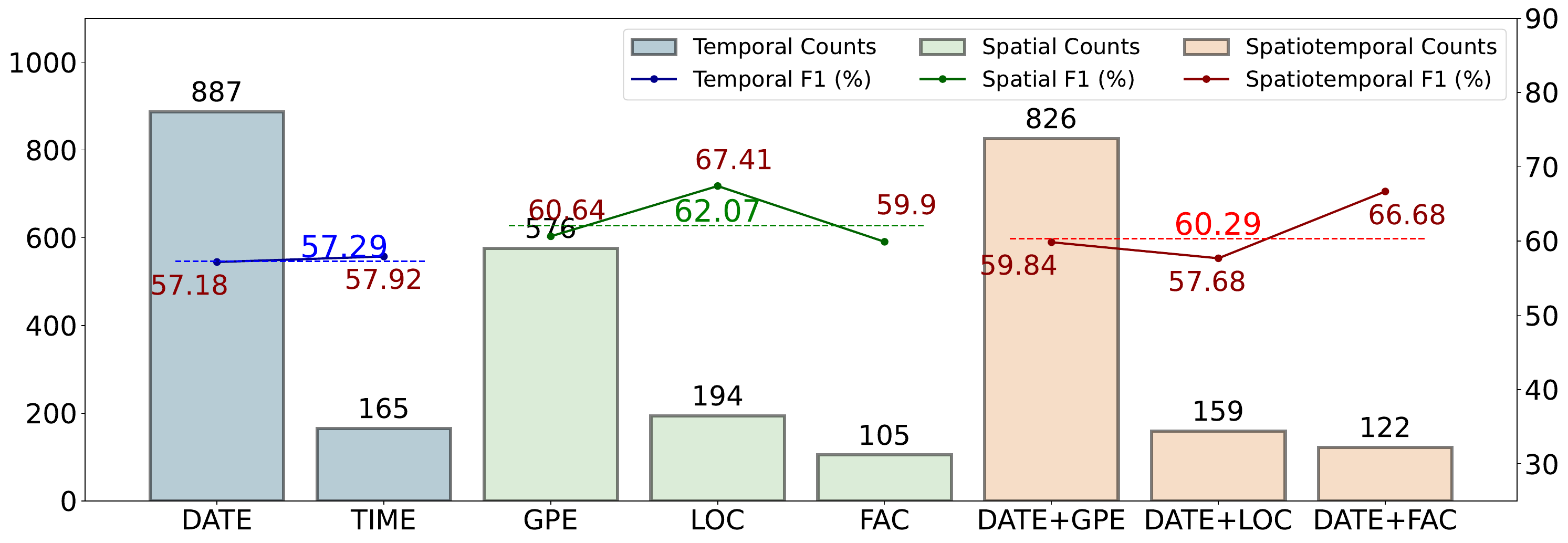}
    }
  \end{minipage}
  \\
  \begin{minipage}[t]{0.75\linewidth} %
    \subfigure[Gemma2-9B It]{
        \includegraphics[width=\linewidth]{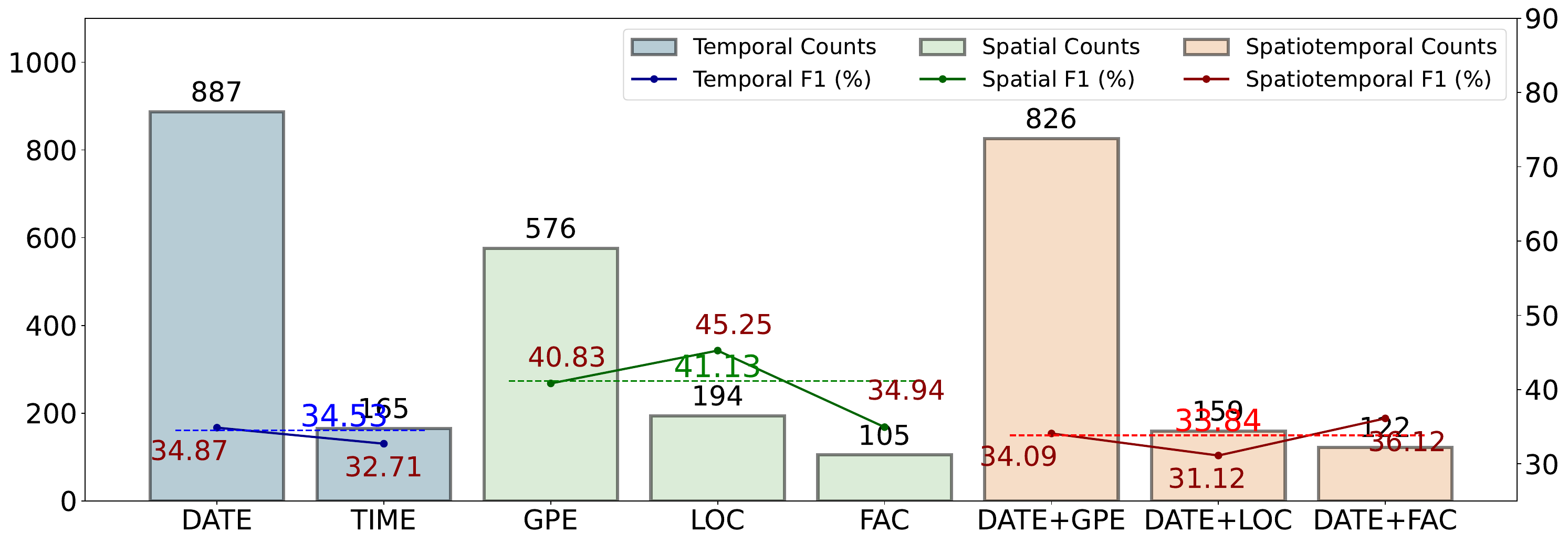}
    }  
  \end{minipage}
  \\
  \begin{minipage}[t]{0.75\linewidth} %
    \subfigure[Qwen2-7B-Instruct]{
        \includegraphics[width=\linewidth]{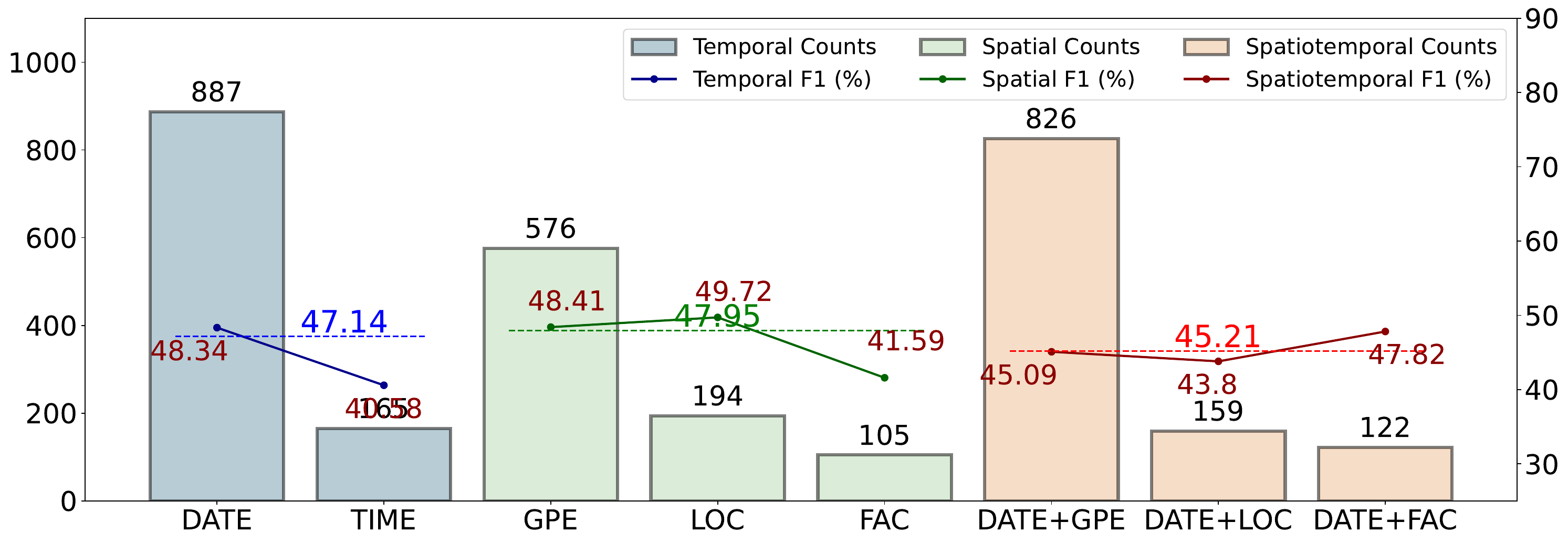}
    }  
  \end{minipage}
  \caption{Performance of Different Models on Spatiotemporal Cognition.}
  \label{apfig:spatiotemporal}
\end{figure*}

\noindent
Fine-grained Studies of Different Tasks: Please refer to Figures \ref{apfig:arithmetic} to \ref{apfig:spatiotemporal}.

\begin{figure*}[!t]
  \centering
  \begin{minipage}[t]{0.75\linewidth} %
    \subfigure[Llama3-8B Instruct]{
        \includegraphics[width=\linewidth]{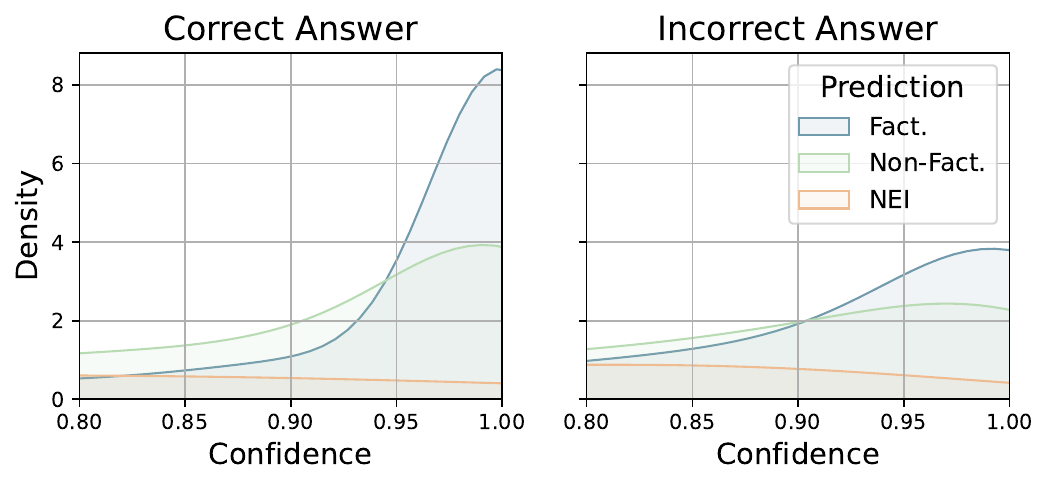}
    }
  \end{minipage}
  \\
  \begin{minipage}[t]{0.75\linewidth} %
    \subfigure[Gemma2-9B It]{
        \includegraphics[width=\linewidth]{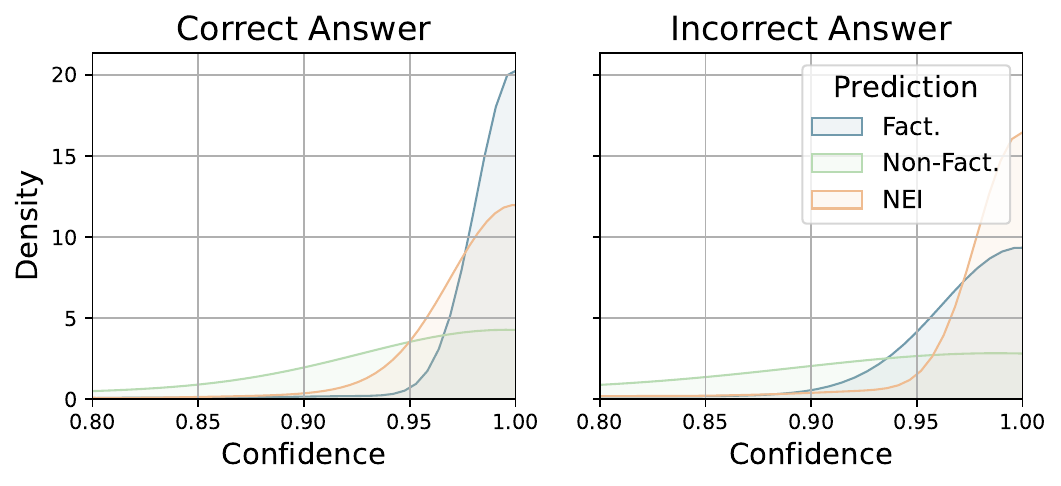}
    }  
  \end{minipage}
  \\
  \begin{minipage}[t]{0.75\linewidth} %
    \subfigure[Qwen2-7B-Instruct]{
        \includegraphics[width=\linewidth]{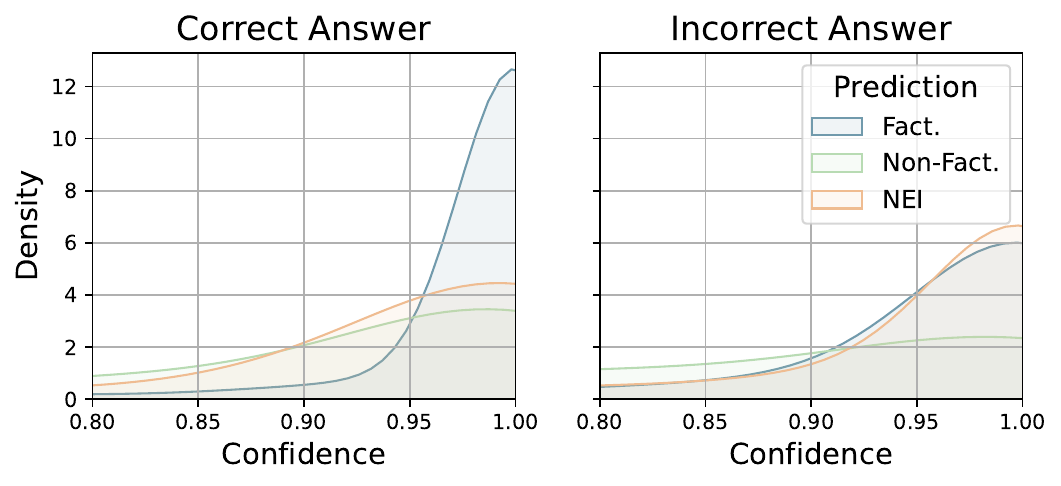}
    }  
  \end{minipage}
  \caption{Confidence of Different Models.}
\label{aptab:confidence}
\end{figure*}

\noindent
Model Confidence Analysis: Please refer to Figure \ref{aptab:confidence}.

\clearpage
\begin{table*}[!htbp]
  \centering
  \resizebox{\linewidth}{!}{
\begin{tabular}{cccccc}
\toprule
Dataset &Task&Source&Evidence/Data Type&Answer Type&Domain\\
\midrule
ToTTo~\shortcite{totto} &Table-to-Text Generation &WikiPedia  &Table & &General   \\
TaKG~\shortcite{qi-etal-2022-takg} &Table-to-Text Generation &WikiPedia &Table, Triplets, Text & &General \\
WebNLG~\shortcite{gardent-etal-2017-webnlg} &Table-to-Text Generation &DBPedia &Triplets & &General\\
DART~\shortcite{nan-etal-2021-dart} &Table-to-Text Generation &WikiPedia &Table, Triplets & &General\\
\midrule
SQA~\shortcite{sqa} &Question Answering &WikiPedia &Table &Span &General  \\
NQ-tables~\shortcite{herzig2021open} &Question Answering &WikiPedia &Table &Span &General \\
HybridQA~\shortcite{hybridqa} &Question Answering &WikiPedia &Table, Text &Span &General \\
WikiTableQuestion~\shortcite{wikitablequestion} &Question Answering &WikiPedia &Table, Text &Span &General   \\
FetaQA~\shortcite{fetaqa} &Question Answering &WikiPedia &Table, Text &Span &General \\
TAT-QA~\shortcite{zhu-etal-2021-tat} &Question Answering &WikiPedia &Table, Text &Span &General \\
Open-WikiTable~\shortcite{openwikitable} &Question Answering &WikiPedia &Table, Text, SQL &Span &General \\
WebSRC~\shortcite{websrc} &Question Answering &Web pages &HTML &Span/Boolean &General\\
OTTQA~\shortcite{chen2020open} &Question Answering &WikiPedia &Table, Text &Multiple Choice &General \\
MATH~\shortcite{math} &Question Answering &Exam &Text &Span &Mathematics  \\
GSM8K~\shortcite{gsm8k} &Question Answering &Exam &Text &Span &Mathematics\\
TSQA~\shortcite{li2021tsqa} & Question Answering &Exam &Table &Multiple Choice &Geography \\
\midrule
FEVEROUS~\shortcite{FEVEROUS} &Fact-checking &WikiPedia &Table, List, Text &Fact/Non-Fact/NEI &General  \\
TabFact~\shortcite{TabFact} &Fact-checking &WikiPedia &Table &Fact/Non-Fact &General \\
Infotabs~\shortcite{gupta-etal-2020-infotabs} &Fact-checking &WikiPedia &Table &Fact/Non-Fact/NEI & General \\
Fact-KG~\shortcite{kim2023factkg} &Fact-checking &WebNLG~\shortcite{gardent-etal-2017-webnlg},DBPedia &Triplets &Fact/Non-Fact &General \\
Semeval 2021 Task 9~\shortcite{wang-etal-2021-semeval} & Fact-checking &Scientific Articles &Table, Text &Fact/Non-Fact/NEI &Science\\
PubHealthTab~\shortcite{pubhealthtab} &Fact-checking &PubHealth~\shortcite{kotonya-toni-2020-explainable-automated}, WikiPedia &HTML &Fact/Non-Fact &Healthcare\\
\name\ &Fact-reasoning &WikiPedia &Table, List, Test & Fact/Non-Fact/NEI & General \\

\bottomrule
\end{tabular}
}
\caption{A comprehensive comparison of various datasets with structural facts.}
\label{tab:qa}
\vspace{-3mm}
\end{table*}

\begin{table*}[!htbp]
  \centering
  \resizebox{\linewidth}{!}{
\begin{tabular}{cccccc}
\toprule
\multirow{2}{*}{Dataset} &\multicolumn{5}{c}{Tasks}\\\cmidrule(lr){2-6}
&Arithmetic Calc. &Multi-hop Reas. &Composition Und. &Combining Struct. and Unstruct. &Spatiotemporal Cogn. \\
\midrule
ToTTo~\shortcite{totto} &\cmark &\cmark &\cmark &&\cmark\\
TaKG~\shortcite{qi-etal-2022-takg} & &\cmark &\cmark &&\cmark \\ 
WebNLG~\shortcite{gardent-etal-2017-webnlg} & &\cmark &\cmark & \\
DART~\shortcite{nan-etal-2021-dart} & &\cmark &\cmark & &\cmark \\
\midrule
SQA~\shortcite{sqa} &\cmark & & &\cmark &\cmark\\
NQ-tables~\shortcite{herzig2021open} & &\cmark &\cmark &  &\cmark\\
HybridQA~\shortcite{hybridqa} & &\cmark & &&\cmark \\
WikiTableQuestion~\shortcite{wikitablequestion} & &\cmark &\cmark &\cmark &\cmark  \\
FetaQA~\shortcite{fetaqa} & &\cmark &\cmark &\cmark &\cmark\\
TAT-QA~\shortcite{zhu-etal-2021-tat} &\cmark &\cmark & &\cmark \\
Open-WikiTable~\shortcite{openwikitable} & &\cmark &\cmark &  &\cmark\\
WebSRC~\shortcite{websrc} & & &\cmark &\cmark \\
OTTQA~\shortcite{chen2020open} &\cmark &\cmark & &\cmark &\cmark\\
MATH~\shortcite{math} &\cmark & & & \\
GSM8K~\shortcite{gsm8k} &\cmark & & & \\
TSQA~\shortcite{li2021tsqa} &\cmark &\cmark &\cmark & \\
\midrule
FEVEROUS~\shortcite{FEVEROUS} &\cmark & \cmark & \cmark & \cmark  \\
TabFact~\shortcite{TabFact} &\cmark & & \cmark &\cmark &\cmark\\
Infotabs~\shortcite{gupta-etal-2020-infotabs} &\cmark &\cmark & &\cmark &\cmark \\
Fact-KG~\shortcite{kim2023factkg} & &\cmark &\cmark &&\cmark \\
Semeval 2021 Task 9~\shortcite{wang-etal-2021-semeval} &\cmark & &\cmark &\cmark \\
PubHealthTab~\shortcite{pubhealthtab} &\cmark &\cmark &\cmark &\cmark\\
\name\ &\cmark &\cmark &\cmark &\cmark &\cmark \\

\bottomrule
\end{tabular}
}
\caption{Distribution of the five tasks across various public datasets containing structural facts.}
\label{tab:qa_tasks}
\vspace{-3mm}
\end{table*}

\begin{table*}[!htbp]
  \centering
\resizebox{\linewidth}{!}{
\begin{tabular}{llcccccccccccc}
\toprule
\multirow{2}{*}{Models}&\multirow{2}{*}{Methods} & \multicolumn{2}{c}{Arithmetic Calc.} & \multicolumn{2}{c}{Spatiotemporal Cogn.}& \multicolumn{2}{c}{Multi-hop Reas.}&\multicolumn{2}{c}{Composition Und.} & \multicolumn{2}{c}{Struct. \& Unstruct.}& \multicolumn{2}{c}{Overall} \\

\cmidrule(lr){3-4} \cmidrule(lr){5-6} \cmidrule(lr){7-8}  \cmidrule(lr){9-10} \cmidrule(lr){11-12} \cmidrule(lr){13-14} 
&& Acc. & F1 & Acc. & F1 & Acc. & F1 & Acc. & F1 & Acc. & F1 & Acc. & F1 \\ \midrule
\multirow{3}{*}{GPT-4o-mini}&Self-Refine. & 58.03 & 60.73 & 54.44 & 60.08 & 51.61 & 58.81 & 66.91 & 68.17 & 56.51 & 63.37 & 55.36 & 60.38 \\
&Self-Consis. & 62.80 & 63.49 & 59.84 & 63.02 & 58.30 & 63.92 & 66.91 & 67.35 & 62.30 & 67.06 & 60.83 & 63.69\\
&Format Instruct. & 62.82 & 63.44 & 58.53 & 61.95 & 57.52 & 63.10 & 66.67 & 67.16 & 61.37 & 66.38 &60.03 & 63.03 \\
\midrule
\multirow{3}{*}{LLaMA3-8B Instruct}&Self-Refine.& 58.05 & 57.74 & 54.87 & 55.40 & 64.71 & 66.75 & 61.20 & 59.46 & 69.83 & 70.94 & 60.23 & 60.44\\
&Self-Consis.& 62.52 & 56.96 & 55.33 & 51.38 & 71.00 & 68.54 & 61.19 & 57.51 & 73.30 & 70.86 & 63.26 & 58.95\\
&Format Instruct.& 61.95 & 55.85 & 51.35 & 46.00 & 68.41 & 66.38 & 60.20 & 54.04 & 71.21 & 68.72 & 60.75  & 55.76\\
\midrule
\multirow{3}{*}{Gemma-2-9B It}&Self-Refine.& 49.56 & 53.86 & 35.18 & 43.16 & 43.62 & 53.69 & 61.69 & 64.01 & 50.63 & 59.93 & 43.41 & 51.04\\
&Self-Consis.& 51.31 & 53.81 & 33.09 & 39.19 & 44.81 & 54.58 & 60.20 & 62.91 & 52.19 & 60.74 & 43.52 & 49.87\\
&Format Instruct.& 53.79 & 55.33 & 34.19 & 40.31 & 45.75 & 55.33 & 59.46 & 61.74 & 53.24 & 61.64&44.88   & 50.97\\
\midrule
\multirow{3}{*}{Qwen2-7B-Instruct}&Self-Refine. & 52.24 & 55.50 & 36.87 & 43.32 & 44.66 & 53.88 & 57.21 & 59.93 & 51.39 & 59.54& 44.95 & 51.34\\
&Self-Consis.& 55.18 & 57.62 & 40.95 & 47.56 & 47.61 & 56.30 & 57.46 & 61.66 & 53.88 & 61.58 & 48.20 & 54.24\\
&Format Instruct.& 56.50 & 58.43 & 37.48 & 43.72 & 48.61 & 57.07 & 57.46 & 61.28 & 54.55 & 62.15& 47.51   & 53.29\\
\bottomrule
\end{tabular}
}
\caption{Performance of GPT-4o-mini using different prompting strategies across five factual tasks.}
\label{aptab:other_prompts}
\end{table*}

\begin{table*}[htb]
  \centering
  \resizebox{\linewidth}{!}{
    \begin{tabular}{lccccccccccccc}
      \toprule
      \multirow{2}{*}{Methods} & \multicolumn{2}{c}{Zero-shot w/o CoT} & \multicolumn{2}{c}{Zero-shot w/ CoT} & \multicolumn{2}{c}{Few-shot w/o CoT} & \multicolumn{2}{c}{Few-shot w/ CoT} & \multicolumn{2}{c}{Overall} \\
      \cmidrule(lr){2-3} \cmidrule(lr){4-5} \cmidrule(lr){6-7} \cmidrule(lr){8-9} \cmidrule(lr){10-11}
      & BA  & MF1 & BA   & MF1 & BA   & MF1 & BA  & MF1 & BA   & MF1 \\ 
      \midrule
Qwen2-7B & 23.06 & 19.97 & 39.69 & 26.00 & 35.85 & 25.69 & 38.86 & 28.70 & 34.37 & 25.09 \\
LLaMA-3-8B & 20.65 & 17.57 & 19.17 & 17.94 & 25.29 & 19.66 & 35.16 & 26.76 & 25.07 & 20.48 \\
Gemma-2-9B & 14.97 & 12.87 & 32.50 & 24.51 & 14.55 & 11.51 & 39.56 & 29.92 & 25.40 & 19.70 \\
\midrule
Qwen2-7B Instruct & 43.51 & 27.29 & 40.87 & 24.53 & 43.81 & 29.25 & 41.45 & 24.66 & 42.41 & 26.43 \\
LLaMA-3-8B Instruct & 39.48 & 32.00 & 39.65 & 25.10 & 40.23 & 32.48 & 42.64 & 28.03 & 40.50 & 29.40 \\
Gemma-2-9B It & 44.34 & 27.74 & 44.58 & 25.94 & 45.11 & 34.52 & 44.51 & 26.81 & 44.64 & 28.75 \\
GLM-4-9B Chat & 42.88 & 38.49 & 43.25 & 26.78 & 42.47 & 38.17 & 44.65 & 28.41 & 43.31 & 32.96 \\
Mistral-7B Instruct & 39.77 & 26.97 & 41.31 & 24.16 & 42.46 & 30.52 & 43.20 & 27.17 & 41.69 & 27.20 \\
\midrule
GPT-4o-Mini & 46.96 & 44.92 & 46.46 & 42.86 & 47.08 & 42.90 & 46.89 & 43.71 & 46.85 & 43.60 \\
GPT-4-Turbo & 48.02 & 45.19 & 48.14 & 43.13 & 47.72 & 43.63 & 48.59 & 43.15 & 48.12 & 43.77 \\
      \bottomrule
    \end{tabular}
}
\caption{Balanced accuracy and Macro F1 of 10 LLMs on the \name\ benchmark using various prompts.}
\label{aptab:main_res_balanced}
\end{table*}

\begin{table*}[!htbp]
  \centering
  \resizebox{\linewidth}{!}{
    \begin{tabular}{lccccccccccccc}
      \toprule
      \multirow{2}{*}{Methods} & \multicolumn{2}{c}{Zero-shot w/o CoT} & \multicolumn{2}{c}{Zero-shot w/ CoT} & \multicolumn{2}{c}{Few-shot w/o CoT} & \multicolumn{2}{c}{Few-shot w/ CoT} & \multicolumn{2}{c}{Overall} \\
      \cmidrule(lr){2-3} \cmidrule(lr){4-5} \cmidrule(lr){6-7} \cmidrule(lr){8-9} \cmidrule(lr){10-11}
& Prec. & Recall & Prec. & Recall & Prec. & Recall & Prec. & Recall & Prec. & Recall \\   

\midrule
Qwen2-7B & 58.64 & 31.82 & 62.33 & 49.40 & 59.71 & 45.39 & 64.49 & 54.80 & 61.29 & 45.35 \\
LLaMA-3-8B & 52.78 & 29.72 & 55.78 & 27.65 & 53.25 & 32.13 & 58.83 & 55.64 & 55.16 & 36.28 \\
Gemma-2-9B & 51.77 & 22.67 & 58.18 & 42.76 & 53.72 & 17.31 & 61.93 & 61.14 & 56.40 & 35.97 \\
\midrule

Qwen2-7B Instruct & 65.11 & 47.85 & 66.63 & 41.27 & 66.79 & 44.88 & 66.67 & 41.01 & 66.30 & 43.75 \\
LLaMA-3-8B Instruct & 62.84 & 62.92 & 64.04 & 43.01 & 63.44 & 63.39 & 67.71 & 45.43 & 64.51 & 53.69 \\
Gemma-2-9B It & 70.37 & 43.53 & 69.03 & 41.08 & 71.10 & 44.81 & 70.67 & 43.03 & 70.29 & 43.11 \\
GLM-4-9B Chat & 64.82 & 52.56 & 68.24 & 42.58 & 65.38 & 52.97 & 68.44 & 47.10 & 66.72 & 48.80 \\
Mistral-7B Instruct & 62.68 & 50.90 & 65.58 & 37.33 & 63.29 & 60.13 & 66.40 & 43.80 & 64.49 & 48.04 \\
\midrule

GPT-4o-Mini & 68.00 & 60.80 & 70.27 & 54.20 & 68.82 & 55.06 & 70.43 & 56.35 & 69.38 & 56.60 \\
GPT-4-Turbo & 68.76 & 60.67 & 71.27 & 53.31 & 69.80 & 56.01 & 71.35 & 53.18 & 70.29 & 55.79 \\
      \bottomrule
    \end{tabular}
}
\caption{Precision and recall of 10 LLMs on the \name\ benchmark using various prompts.}
\vspace{-5mm}
\label{aptab:main_res_pr}
\end{table*}

\begin{table*}[!htbp]
\centering
\resizebox{\textwidth}{!}{
\begin{tabular}{lcccccccccccc}
\toprule
\multirow{2}{*}{Methods} & \multicolumn{2}{c}{Arithmetic Calc.} & \multicolumn{2}{c}{Spatiotemporal Cogn.}& \multicolumn{2}{c}{Multi-hop Reas.} & \multicolumn{2}{c}{Composition Und.}  & \multicolumn{2}{c}{Struct. \& Unstruct.} \\

\cmidrule(lr){2-3} \cmidrule(lr){4-5} \cmidrule(lr){6-7} \cmidrule(lr){8-9} \cmidrule(lr){10-11}
& Prec. & Recall & Prec. & Recall & Prec. & Recall & Prec. & Recall & Prec. & Recall \\ \midrule

Qwen2-7B & 58.11 & 28.30 & 55.32 & 28.24 & 68.87 & 34.05 & 65.94 & 38.31 & 69.78 & 41.26 \\
LLaMA-3-8B & 54.34 & 28.48 & 49.83 & 28.36 & 63.72 & 29.61 & 53.61 & 34.08 & 63.42 & 34.00 \\
Gemma-2-9B & 51.77 & 15.98 & 48.47 & 21.45 & 67.26 & 30.55 & 60.43 & 25.87 & 61.55 & 25.92 \\
\midrule
Qwen2-7B Instruct & 61.48 & 54.58 & 66.00 & 40.52 & 71.62 & 47.91 & 70.20 & 57.46 & 73.95 & 53.33 \\
LLaMA-3-8B Instruct & 57.41 & 62.28 & 62.58 & 54.78 & 69.43 & 70.61 & 61.54 & 60.94 & 72.15 & 73.28 \\
Gemma-2-9B It & 60.51 & 51.36 & 73.96 & 33.03 & 81.38 & 44.73 & 71.58 & 59.95 & 82.41 & 52.46 \\
GLM-4-9B Chat & 63.06 & 59.27 & 63.75 & 46.70 & 70.01 & 50.67 & 67.78 & 63.93 & 73.41 & 56.80 \\
Mistral-7B Instruct & 61.48 & 55.37 & 59.07 & 43.44 & 70.98 & 52.03 & 62.60 & 54.98 & 73.77 & 59.07 \\
\midrule
GPT-4o-Mini & 64.51 & 62.52 & 67.79 & 60.13 & 75.06 & 58.04 & 68.80 & 67.42 & 75.06 & 62.10 \\
GPT-4-Turbo & 64.16 & 61.76 & 68.71 & 61.93 & 75.03 & 54.90 & 70.78 & 70.15 & 77.18 & 61.59 \\
\midrule
Overall & 59.68 & 47.99 & 61.55 & 41.86 & 71.34 & 47.31 & 65.33 & 53.31 & 72.27 & 51.98 \\
\bottomrule
\end{tabular}
}
\caption{Precision and recall of 10 LLMs on the \name\ benchmark across five factual tasks under the zero-shot w/o CoT setting.}
\label{aptab:task_pr_zero_shot_wo_cot}
\end{table*}

\begin{table*}[!htbp]
\centering
\resizebox{\textwidth}{!}{
\begin{tabular}{lcccccccccccc}
\toprule
\multirow{2}{*}{Methods} & \multicolumn{2}{c}{Arithmetic Calc.} & \multicolumn{2}{c}{Spatiotemporal Cogn.}& \multicolumn{2}{c}{Multi-hop Reas.} & \multicolumn{2}{c}{Composition Und.}  & \multicolumn{2}{c}{Struct. \& Unstruct.} \\

\cmidrule(lr){2-3} \cmidrule(lr){4-5} \cmidrule(lr){6-7} \cmidrule(lr){8-9} \cmidrule(lr){10-11}
& BA  & MF1 & BA   & MF1 & BA   & MF1 & BA  & MF1 & BA   & MF1 \\ 
\midrule

Qwen2-7B & 19.42 & 18.07 & 23.62 & 19.35 & 22.30 & 19.48 & 27.85 & 23.99 & 27.13 & 21.99 \\
LLaMA-3-8B & 20.35 & 17.12 & 21.08 & 17.39 & 18.61 & 16.23 & 32.55 & 21.59 & 22.01 & 18.34 \\
Gemma-2-9B & 11.08 & 10.58 & 15.57 & 12.84 & 18.50 & 14.40 & 16.96 & 14.42 & 15.33 & 12.74 \\
\midrule
Qwen2-7B Instruct & 40.04 & 30.66 & 44.59 & 29.03 & 43.23 & 25.79 & 40.16 & 34.04 & 44.43 & 29.72 \\
LLaMA-3-8B Instruct & 37.14 & 35.27 & 39.92 & 31.00 & 38.34 & 37.48 & 38.26 & 37.28 & 40.23 & 39.96 \\
Gemma-2-9B It & 38.61 & 32.89 & 44.50 & 29.55 & 45.61 & 33.09 & 55.90 & 37.33 & 48.18 & 35.80 \\
GLM-4-9B Chat & 38.22 & 36.99 & 44.85 & 37.61 & 40.91 & 34.71 & 41.31 & 42.14 & 44.53 & 38.48 \\
Mistral-7B Instruct & 38.64 & 26.56 & 38.69 & 25.59 & 39.45 & 25.58 & 34.91 & 30.13 & 41.78 & 28.08 \\
\midrule
GPT-4o-Mini & 41.68 & 41.14 & 49.20 & 46.20 & 46.31 & 41.15 & 52.58 & 52.69 & 46.86 & 42.68 \\
GPT-4-Turbo & 41.05 & 40.17 & 50.43 & 47.09 & 47.20 & 39.42 & 58.31 & 60.05 & 49.42 & 43.67 \\
\midrule
Overall & 32.62 & 28.95 & 37.25 & 29.56 & 36.05 & 28.73 & 39.88 & 35.37 & 37.99 & 31.15 \\
\bottomrule
\end{tabular}
}
\caption{Balanced accuracy and Macro F1 of 10 LLMs on the \name\ benchmark across five factual tasks under the zero-shot w/o CoT setting.}
\end{table*}

\begin{table*}[!htbp]
\centering
\resizebox{\textwidth}{!}{
\begin{tabular}{lcccccccccccc}
\toprule
\multirow{2}{*}{Methods} & \multicolumn{2}{c}{Arithmetic Calc.} & \multicolumn{2}{c}{Spatiotemporal Cogn.}& \multicolumn{2}{c}{Multi-hop Reas.} & \multicolumn{2}{c}{Composition Und.}  & \multicolumn{2}{c}{Struct. \& Unstruct.} \\

\cmidrule(lr){2-3} \cmidrule(lr){4-5} \cmidrule(lr){6-7} \cmidrule(lr){8-9} \cmidrule(lr){10-11}
& Acc. & F1 & Acc. & F1 & Acc. & F1 & Acc. & F1 & Acc. & F1 \\ \midrule

Qwen2-7B & 57.65 & 57.56 & 39.63 & 42.68 & 52.11 & 59.05 & 59.21 & 59.50 & 55.22 & 61.00 \\
LLaMA-3-8B & 28.50 & 37.79 & 24.15 & 31.16 & 28.98 & 39.88 & 30.10 & 38.53 & 32.33 & 43.38 \\
Gemma-2-9B & 45.35 & 50.28 & 37.92 & 42.52 & 44.35 & 52.43 & 44.03 & 48.90 & 47.74 & 54.98 \\
\midrule
Qwen2-7B Instruct & 53.92 & 56.87 & 31.66 & 39.43 & 39.74 & 49.90 & 51.24 & 55.54 & 45.18 & 55.02 \\
LLaMA-3-8B Instruct & 50.37 & 54.01 & 36.13 & 42.62 & 40.61 & 49.98 & 51.49 & 54.39 & 49.30 & 57.93 \\
Gemma-2-9B It & 48.75 & 53.66 & 33.91 & 43.30 & 40.61 & 51.25 & 57.46 & 60.31 & 44.87 & 55.46 \\
GLM-4-9B Chat & 53.57 & 58.14 & 35.52 & 45.05 & 39.00 & 49.17 & 56.22 & 59.62 & 45.09 & 55.43 \\
Mistral-7B Instruct & 43.93 & 50.87 & 32.00 & 40.89 & 34.94 & 44.83 & 50.00 & 55.88 & 40.95 & 51.28 \\
\midrule
GPT-4o-Mini & 59.10 & 61.50 & 50.89 & 59.04 & 51.64 & 60.06 & 65.18 & 66.46 & 56.22 & 64.12 \\
GPT-4-Turbo & 58.44 & 61.04 & 51.71 & 60.26 & 49.48 & 57.78 & 64.93 & 65.93 & 52.64 & 61.56 \\
\midrule
Overall & 49.96 & 54.17 & 37.35 & 44.70 & 42.15 & 51.43 & 52.99 & 56.51 & 46.95 & 56.02 \\
\bottomrule
\end{tabular}
}
\caption{Accuracy and F1 score of 10 LLMs on the \name\ benchmark across five factual tasks under the zero-shot w/ CoT setting.}
\end{table*}

\begin{table*}[!htbp]
\centering
\resizebox{\textwidth}{!}{
\begin{tabular}{lcccccccccccc}
\toprule
\multirow{2}{*}{Methods} & \multicolumn{2}{c}{Arithmetic Calc.} & \multicolumn{2}{c}{Spatiotemporal Cogn.}& \multicolumn{2}{c}{Multi-hop Reas.} & \multicolumn{2}{c}{Composition Und.}  & \multicolumn{2}{c}{Struct. \& Unstruct.} \\

\cmidrule(lr){2-3} \cmidrule(lr){4-5} \cmidrule(lr){6-7} \cmidrule(lr){8-9} \cmidrule(lr){10-11}
& Prec. & Recall & Prec. & Recall & Prec. & Recall & Prec. & Recall & Prec. & Recall \\ \midrule

Qwen2-7B & 60.38 & 57.65 & 61.86 & 39.63 & 70.62 & 52.11 & 66.14 & 59.21 & 71.26 & 55.22 \\
LLaMA-3-8B & 57.51 & 28.50 & 52.43 & 24.15 & 66.28 & 28.98 & 57.02 & 30.10 & 67.36 & 32.33 \\
Gemma-2-9B & 58.46 & 45.35 & 55.55 & 37.92 & 67.74 & 44.35 & 57.02 & 44.03 & 69.66 & 47.74 \\
\midrule
Qwen2-7B Instruct & 61.87 & 53.92 & 69.06 & 31.66 & 73.41 & 39.74 & 67.39 & 51.24 & 75.92 & 45.18 \\
LLaMA-3-8B Instruct & 60.99 & 50.37 & 63.89 & 36.13 & 69.48 & 40.61 & 63.20 & 51.49 & 74.34 & 49.30 \\
Gemma-2-9B It & 61.11 & 48.75 & 71.34 & 33.91 & 77.50 & 40.61 & 66.14 & 57.46 & 78.25 & 44.87 \\
GLM-4-9B Chat & 64.25 & 53.57 & 69.49 & 35.52 & 74.36 & 39.00 & 65.64 & 56.22 & 76.37 & 45.09 \\
Mistral-7B Instruct & 62.24 & 43.93 & 64.62 & 32.00 & 72.72 & 34.94 & 65.04 & 50.00 & 74.37 & 40.95 \\
\midrule
GPT-4o-Mini & 64.24 & 59.10 & 72.71 & 50.89 & 76.01 & 51.64 & 68.24 & 65.18 & 76.66 & 56.22 \\
GPT-4-Turbo & 64.22 & 58.44 & 74.82 & 51.71 & 76.52 & 49.48 & 67.30 & 64.93 & 77.91 & 52.64 \\
\midrule
Overall & 61.53 & 49.96 & 65.58 & 37.35 & 72.46 & 42.15 & 64.31 & 52.99 & 74.21 & 46.95 \\
\bottomrule
\end{tabular}
}
\caption{Precision and recall of 10 LLMs on the \name\ benchmark across five factual tasks under the zero-shot w/ CoT setting.}
\end{table*}

\begin{table*}[!htbp]
\centering
\resizebox{\textwidth}{!}{
\begin{tabular}{lcccccccccccc}
\toprule
\multirow{2}{*}{Methods} & \multicolumn{2}{c}{Arithmetic Calc.} & \multicolumn{2}{c}{Spatiotemporal Cogn.}& \multicolumn{2}{c}{Multi-hop Reas.} & \multicolumn{2}{c}{Composition Und.}  & \multicolumn{2}{c}{Struct. \& Unstruct.} \\

\cmidrule(lr){2-3} \cmidrule(lr){4-5} \cmidrule(lr){6-7} \cmidrule(lr){8-9} \cmidrule(lr){10-11}
& BA  & MF1 & BA   & MF1 & BA   & MF1 & BA  & MF1 & BA   & MF1 \\ 
\midrule

Qwen2-7B & 36.98 & 26.80 & 40.23 & 23.88 & 39.31 & 26.03 & 37.19 & 28.95 & 42.11 & 25.92 \\
LLaMA-3-8B & 19.80 & 18.25 & 18.81 & 17.00 & 17.94 & 16.90 & 18.92 & 18.46 & 20.44 & 19.13 \\
Gemma-2-9B & 31.21 & 24.49 & 33.22 & 23.59 & 31.07 & 23.50 & 28.77 & 24.23 & 34.96 & 25.27 \\
\midrule
Qwen2-7B Instruct & 36.21 & 26.54 & 41.80 & 22.07 & 41.48 & 23.18 & 35.79 & 27.41 & 43.30 & 23.86 \\
LLaMA-3-8B Instruct & 34.25 & 25.21 & 41.18 & 23.77 & 37.35 & 22.84 & 32.38 & 26.36 & 45.42 & 26.09 \\
Gemma-2-9B It & 36.29 & 25.70 & 46.34 & 24.21 & 46.42 & 25.20 & 51.45 & 35.83 & 46.82 & 25.81 \\
GLM-4-9B Chat & 38.35 & 28.25 & 42.97 & 24.81 & 45.16 & 24.60 & 47.18 & 31.43 & 46.62 & 26.48 \\
Mistral-7B Instruct & 36.19 & 24.77 & 40.53 & 22.61 & 43.70 & 22.41 & 36.69 & 28.33 & 45.62 & 24.15 \\
\midrule
GPT-4o-Mini & 40.26 & 40.38 & 48.84 & 42.82 & 46.16 & 40.34 & 46.80 & 46.44 & 47.69 & 41.38 \\
GPT-4-Turbo & 42.68 & 41.29 & 50.38 & 43.69 & 48.17 & 39.82 & 53.91 & 49.44 & 48.18 & 41.04 \\
\midrule
Overall & 35.22 & 28.17 & 40.43 & 26.84 & 39.68 & 26.48 & 38.91 & 31.69 & 42.12 & 27.91 \\
\bottomrule
\end{tabular}
}
\caption{Balanced accuracy and Macro F1 of 10 LLMs on the \name\ benchmark across five factual tasks under the zero-shot w/ CoT setting.}
\end{table*}

\begin{table*}[!htbp]
\centering
\resizebox{\textwidth}{!}{
\begin{tabular}{lcccccccccccc}
\toprule
\multirow{2}{*}{Methods} & \multicolumn{2}{c}{Arithmetic Calc.} & \multicolumn{2}{c}{Spatiotemporal Cogn.}& \multicolumn{2}{c}{Multi-hop Reas.} & \multicolumn{2}{c}{Composition Und.}  & \multicolumn{2}{c}{Struct. \& Unstruct.} \\

\cmidrule(lr){2-3} \cmidrule(lr){4-5} \cmidrule(lr){6-7} \cmidrule(lr){8-9} \cmidrule(lr){10-11}
& Acc. & F1 & Acc. & F1 & Acc. & F1 & Acc. & F1 & Acc. & F1 \\ \midrule

Qwen2-7B & 50.19 & 54.03 & 39.36 & 44.78 & 46.28 & 54.34 & 50.75 & 53.96 & 50.23 & 57.83 \\
LLaMA-3-8B & 30.30 & 37.95 & 29.45 & 34.63 & 36.58 & 46.16 & 30.35 & 37.67 & 35.85 & 45.21 \\
Gemma-2-9B & 17.43 & 23.45 & 15.98 & 20.19 & 18.65 & 25.42 & 18.41 & 25.58 & 18.46 & 24.95 \\
\midrule
Qwen2-7B Instruct & 54.35 & 57.82 & 36.45 & 43.11 & 44.58 & 53.43 & 57.21 & 60.90 & 49.14 & 57.50 \\
LLaMA-3-8B Instruct & 63.37 & 58.22 & 55.42 & 53.08 & 70.17 & 68.47 & 63.68 & 60.77 & 73.30 & 72.09 \\
Gemma-2-9B It & 55.12 & 56.73 & 34.22 & 41.49 & 45.62 & 55.37 & 63.19 & 64.72 & 50.75 & 60.11 \\
GLM-4-9B Chat & 60.39 & 59.73 & 44.69 & 49.17 & 53.96 & 60.11 & 63.93 & 63.60 & 58.41 & 64.52 \\
Mistral-7B Instruct & 61.35 & 60.78 & 52.10 & 52.54 & 66.71 & 68.99 & 63.68 & 61.93 & 68.38 & 70.15 \\
\midrule
GPT-4o-Mini & 60.38 & 62.24 & 52.50 & 58.28 & 52.55 & 59.80 & 66.42 & 67.15 & 54.84 & 62.19 \\
GPT-4-Turbo & 60.38 & 61.88 & 55.38 & 61.06 & 51.63 & 58.24 & 66.42 & 66.91 & 54.98 & 63.15 \\
\midrule
Overall & 51.33 & 53.28 & 41.55 & 45.83 & 48.67 & 55.03 & 54.40 & 56.32 & 51.43 & 57.77 \\
\bottomrule
\end{tabular}
}
\caption{Accuracy and F1 score of 10 LLMs on the \name\ benchmark across five factual tasks under the few-shot w/o CoT setting.}

\end{table*}

\begin{table*}[!htbp]
\centering
\resizebox{\textwidth}{!}{
\begin{tabular}{lcccccccccccc}
\toprule
\multirow{2}{*}{Methods} & \multicolumn{2}{c}{Arithmetic Calc.} & \multicolumn{2}{c}{Spatiotemporal Cogn.}& \multicolumn{2}{c}{Multi-hop Reas.} & \multicolumn{2}{c}{Composition Und.}  & \multicolumn{2}{c}{Struct. \& Unstruct.} \\

\cmidrule(lr){2-3} \cmidrule(lr){4-5} \cmidrule(lr){6-7} \cmidrule(lr){8-9} \cmidrule(lr){10-11}
& Prec. & Recall & Prec. & Recall & Prec. & Recall & Prec. & Recall & Prec. & Recall \\ \midrule

Qwen2-7B & 59.91 & 50.19 & 57.40 & 39.36 & 68.94 & 46.28 & 61.30 & 50.75 & 69.95 & 50.23 \\
LLaMA-3-8B & 53.94 & 30.30 & 49.07 & 29.45 & 66.43 & 36.58 & 55.05 & 30.35 & 65.39 & 35.85 \\
Gemma-2-9B & 54.89 & 17.43 & 48.64 & 15.98 & 64.74 & 18.65 & 65.32 & 18.41 & 66.72 & 18.46 \\
\midrule
Qwen2-7B Instruct & 64.18 & 54.35 & 67.43 & 36.45 & 73.35 & 44.58 & 71.91 & 57.21 & 74.64 & 49.14 \\
LLaMA-3-8B Instruct & 58.19 & 63.37 & 63.33 & 55.42 & 68.70 & 70.17 & 64.32 & 63.68 & 73.42 & 73.30 \\
Gemma-2-9B It & 62.17 & 55.12 & 73.25 & 34.22 & 80.08 & 45.62 & 71.86 & 63.19 & 81.23 & 50.75 \\
GLM-4-9B Chat & 62.74 & 60.39 & 65.13 & 44.69 & 70.68 & 53.96 & 70.20 & 63.93 & 74.36 & 58.41 \\
Mistral-7B Instruct & 61.39 & 61.35 & 61.66 & 52.10 & 71.83 & 66.71 & 65.70 & 63.68 & 73.17 & 68.38 \\
\midrule
GPT-4o-Mini & 65.42 & 60.38 & 67.54 & 52.50 & 76.16 & 52.55 & 68.91 & 66.42 & 76.72 & 54.84 \\
GPT-4-Turbo & 64.26 & 60.38 & 70.50 & 55.38 & 76.40 & 51.63 & 67.73 & 66.42 & 78.20 & 54.98 \\
Overall & 60.71 & 51.33 & 62.40 & 41.55 & 71.73 & 48.67 & 66.23 & 54.40 & 73.38 & 51.43 \\

\bottomrule
\end{tabular}
}
\caption{Precision and recall of 10 LLMs on the \name\ benchmark across five factual tasks under the few-shot w/o CoT setting.}
\end{table*}

\begin{table*}[!htbp]
\centering
\resizebox{\textwidth}{!}{
\begin{tabular}{lcccccccccccc}
\toprule
\multirow{2}{*}{Methods} & \multicolumn{2}{c}{Arithmetic Calc.} & \multicolumn{2}{c}{Spatiotemporal Cogn.}& \multicolumn{2}{c}{Multi-hop Reas.} & \multicolumn{2}{c}{Composition Und.}  & \multicolumn{2}{c}{Struct. \& Unstruct.} \\

\cmidrule(lr){2-3} \cmidrule(lr){4-5} \cmidrule(lr){6-7} \cmidrule(lr){8-9} \cmidrule(lr){10-11}
& BA  & MF1 & BA   & MF1 & BA   & MF1 & BA  & MF1 & BA   & MF1 \\ 
\midrule
Qwen2-7B & 36.48 & 26.20 & 34.27 & 24.48 & 35.19 & 24.32 & 57.21 & 30.75 & 36.01 & 25.16 \\
LLaMA-3-8B & 23.39 & 18.58 & 26.25 & 19.28 & 26.55 & 20.46 & 22.82 & 18.56 & 25.00 & 19.54 \\
Gemma-2-9B & 14.01 & 11.48 & 14.54 & 10.91 & 14.74 & 11.74 & 12.36 & 12.65 & 15.24 & 11.31 \\
\midrule
Qwen2-7B Instruct & 42.22 & 31.10 & 43.03 & 32.05 & 43.52 & 32.24 & 54.20 & 42.54 & 45.12 & 32.96 \\
LLaMA-3-8B Instruct & 36.66 & 35.01 & 41.74 & 31.90 & 37.46 & 36.98 & 40.16 & 39.54 & 41.87 & 42.13 \\
Gemma-2-9B It & 40.02 & 35.09 & 45.79 & 31.18 & 46.99 & 34.38 & 61.71 & 47.96 & 44.61 & 35.25 \\
GLM-4-9B Chat & 38.28 & 37.16 & 44.50 & 36.40 & 39.18 & 35.47 & 44.21 & 43.11 & 45.03 & 39.22 \\
Mistral-7B Instruct & 39.92 & 35.92 & 43.07 & 29.12 & 42.08 & 34.00 & 43.79 & 43.40 & 43.94 & 34.62 \\
\midrule
GPT-4o-Mini & 42.76 & 41.79 & 47.27 & 42.27 & 46.73 & 39.86 & 55.97 & 53.10 & 49.30 & 40.46 \\
GPT-4-Turbo & 41.90 & 41.22 & 49.54 & 44.33 & 48.96 & 39.64 & 45.09 & 44.99 & 47.29 & 41.07 \\
\midrule
Overall & 35.56 & 31.36 & 39.00 & 30.19 & 38.14 & 30.91 & 43.75 & 37.66 & 39.34 & 32.17 \\

\bottomrule
\end{tabular}
}
\caption{Balanced accuracy and Macro F1 of 10 LLMs on the \name\ benchmark across five factual tasks under the few-shot w/o CoT setting.}
\end{table*}

\begin{table*}[!htbp]
\centering
\resizebox{\textwidth}{!}{
\begin{tabular}{lcccccccccccc}
\toprule
\multirow{2}{*}{Methods} & \multicolumn{2}{c}{Arithmetic Calc.} & \multicolumn{2}{c}{Spatiotemporal Cogn.}& \multicolumn{2}{c}{Multi-hop Reas.} & \multicolumn{2}{c}{Composition Und.}  & \multicolumn{2}{c}{Struct. \& Unstruct.} \\

\cmidrule(lr){2-3} \cmidrule(lr){4-5} \cmidrule(lr){6-7} \cmidrule(lr){8-9} \cmidrule(lr){10-11}
& Acc. & F1 & Acc. & F1 & Acc. & F1 & Acc. & F1 & Acc. & F1 \\ \midrule

Qwen2-7B & 57.98 & 58.50 & 47.25 & 49.93 & 58.91 & 64.52 & 61.19 & 62.20 & 61.66 & 66.44 \\
LLaMA-3-8B & 56.03 & 55.56 & 48.52 & 46.85 & 60.50 & 63.71 & 54.97 & 54.24 & 65.22 & 66.91 \\
Gemma-2-9B & 60.02 & 58.96 & 53.72 & 52.45 & 68.17 & 69.11 & 62.94 & 62.50 & 70.94 & 71.31 \\
\midrule
Qwen2-7B Instruct & 51.68 & 55.78 & 31.88 & 39.99 & 41.24 & 51.29 & 52.74 & 58.31 & 44.62 & 54.55 \\
LLaMA-3-8B Instruct & 52.21 & 57.44 & 40.64 & 49.51 & 42.44 & 52.08 & 53.23 & 57.34 & 48.74 & 58.37 \\
Gemma-2-9B It & 53.86 & 57.52 & 34.86 & 44.45 & 42.03 & 52.95 & 61.44 & 63.33 & 45.07 & 55.99 \\
GLM-4-9B Chat & 56.36 & 59.70 & 39.08 & 47.86 & 46.19 & 55.65 & 60.45 & 63.31 & 51.28 & 60.48 \\
Mistral-7B Instruct & 48.49 & 54.83 & 38.30 & 46.84 & 43.53 & 52.71 & 58.21 & 61.53 & 48.05 & 57.49 \\
\midrule
GPT-4o-Mini & 62.36 & 63.52 & 52.88 & 60.49 & 53.66 & 61.60 & 70.15 & 70.40 & 57.11 & 64.72 \\
GPT-4-Turbo & 60.03 & 62.38 & 50.98 & 59.20 & 48.95 & 57.88 & 66.42 & 67.97 & 51.70 & 61.01 \\
\midrule
Overall & 55.90 & 58.42 & 43.81 & 49.76 & 50.56 & 58.15 & 60.17 & 62.11 & 54.44 & 61.73 \\
\bottomrule
\end{tabular}
}
\caption{Accuracy and F1 score of 10 LLMs on the \name\ benchmark across five factual tasks under the few-shot w/ CoT setting.}
\vspace{-5mm}
\end{table*}

\begin{table*}[!htbp]
\centering
\resizebox{\textwidth}{!}{
\begin{tabular}{lcccccccccccc}
\toprule
\multirow{2}{*}{Methods} & \multicolumn{2}{c}{Arithmetic Calc.} & \multicolumn{2}{c}{Spatiotemporal Cogn.}& \multicolumn{2}{c}{Multi-hop Reas.} & \multicolumn{2}{c}{Composition Und.}  & \multicolumn{2}{c}{Struct. \& Unstruct.} \\

\cmidrule(lr){2-3} \cmidrule(lr){4-5} \cmidrule(lr){6-7} \cmidrule(lr){8-9} \cmidrule(lr){10-11}
& Prec. & Recall & Prec. & Recall & Prec. & Recall & Prec. & Recall & Prec. & Recall \\ \midrule

Qwen2-7B & 61.01 & 57.98 & 63.75 & 47.25 & 72.32 & 58.91 & 69.20 & 61.19 & 73.67 & 61.66 \\
LLaMA-3-8B & 56.82 & 56.03 & 59.04 & 48.52 & 67.47 & 60.50 & 60.35 & 54.97 & 69.66 & 65.22 \\
Gemma-2-9B & 58.77 & 60.02 & 61.26 & 53.72 & 70.30 & 68.17 & 65.10 & 62.94 & 72.25 & 70.94 \\
\midrule
Qwen2-7B Instruct & 62.31 & 51.68 & 68.47 & 31.88 & 73.24 & 41.24 & 67.83 & 52.74 & 75.82 & 44.62 \\
LLaMA-3-8B Instruct & 64.28 & 52.21 & 68.66 & 40.64 & 73.38 & 42.44 & 63.49 & 53.23 & 75.91 & 48.74 \\
Gemma-2-9B It & 63.13 & 53.86 & 72.82 & 34.86 & 78.73 & 42.03 & 67.84 & 61.44 & 79.02 & 45.07 \\
GLM-4-9B Chat & 63.87 & 56.36 & 68.98 & 39.08 & 75.07 & 46.19 & 69.23 & 60.45 & 76.90 & 51.28 \\
Mistral-7B Instruct & 64.16 & 48.49 & 65.57 & 38.30 & 72.79 & 43.53 & 65.79 & 58.21 & 74.63 & 48.05 \\
\midrule
GPT-4o-Mini & 64.77 & 62.36 & 72.79 & 52.88 & 76.08 & 53.66 & 71.00 & 70.15 & 76.38 & 57.11 \\
GPT-4-Turbo & 65.18 & 60.03 & 72.94 & 50.98 & 78.59 & 48.95 & 70.02 & 66.42 & 78.80 & 51.70 \\
\midrule
Overall & 62.43 & 55.90 & 67.43 & 43.81 & 73.80 & 50.56 & 66.99 & 60.17 & 75.30 & 54.44 \\
\bottomrule
\end{tabular}
}
\caption{Precision and recall of 10 LLMs on the \name\ benchmark across five factual tasks under the few-shot w/ CoT setting.}
\end{table*}

\begin{table*}[!htbp]
\centering
\resizebox{\textwidth}{!}{
\begin{tabular}{lcccccccccccc}
\toprule
\multirow{2}{*}{Methods} & \multicolumn{2}{c}{Arithmetic Calc.} & \multicolumn{2}{c}{Spatiotemporal Cogn.}& \multicolumn{2}{c}{Multi-hop Reas.} & \multicolumn{2}{c}{Composition Und.}  & \multicolumn{2}{c}{Struct. \& Unstruct.} \\

\cmidrule(lr){2-3} \cmidrule(lr){4-5} \cmidrule(lr){6-7} \cmidrule(lr){8-9} \cmidrule(lr){10-11}
& BA  & MF1 & BA   & MF1 & BA   & MF1 & BA  & MF1 & BA   & MF1 \\ 
\midrule
Qwen2-7B & 36.38 & 27.26 & 39.40 & 27.52 & 38.25 & 28.76 & 38.74 & 30.34 & 41.33 & 29.97 \\
LLaMA-3-8B & 33.82 & 25.66 & 36.34 & 25.64 & 33.19 & 26.29 & 44.65 & 29.37 & 36.41 & 28.65 \\
Gemma-2-9B & 36.45 & 27.61 & 40.59 & 28.89 & 38.58 & 29.72 & 47.59 & 33.37 & 41.84 & 32.09 \\
\midrule
Qwen2-7B Instruct & 37.95 & 26.50 & 40.97 & 22.27 & 41.58 & 23.80 & 37.52 & 29.18 & 46.30 & 24.26 \\
LLaMA-3-8B Instruct & 36.26 & 27.89 & 45.20 & 27.38 & 42.27 & 25.74 & 34.38 & 28.20 & 45.90 & 27.54 \\
Gemma-2-9B It & 38.08 & 27.32 & 45.93 & 24.73 & 46.63 & 26.56 & 50.37 & 37.23 & 45.33 & 26.06 \\
GLM-4-9B Chat & 37.39 & 28.48 & 44.61 & 26.39 & 48.13 & 27.40 & 46.04 & 36.53 & 48.93 & 31.90 \\
Mistral-7B Instruct & 38.88 & 27.36 & 43.15 & 25.80 & 45.75 & 25.82 & 38.37 & 34.12 & 44.68 & 26.94 \\
\midrule
GPT-4o-Mini & 41.51 & 41.51 & 49.39 & 43.89 & 47.84 & 41.21 & 46.62 & 47.22 & 45.37 & 41.44 \\
GPT-4-Turbo & 43.31 & 42.05 & 49.59 & 42.99 & 49.79 & 40.50 & 55.17 & 50.34 & 50.68 & 41.24 \\
\midrule
Overall & 38.00 & 30.16 & 43.52 & 29.55 & 43.20 & 29.58 & 43.95 & 35.59 & 44.68 & 31.01 \\
\bottomrule
\end{tabular}
}
\caption{Balanced accuracy and Macro F1 of 10 LLMs on the \name\ benchmark across five factual tasks under the few-shot w/ CoT setting.}
\label{aptab:task_balanced_few_shot_w_cot}
\end{table*}

\renewcommand{\arraystretch}{1.15}

\begin{table*}[t!]
\centering
\resizebox{\textwidth}{!}{
\begin{tabular}{llcccccccccc}
\toprule
\multirow{2}{*}{Settings} &\multirow{2}{*}{Metrics}  & \multicolumn{2}{c}{Arithmetic Calc.} & \multicolumn{2}{c}{Spatiotemporal Cogn.}& \multicolumn{2}{c}{Multi-hop Reas.} & \multicolumn{2}{c}{Composition Und.}  & \multicolumn{2}{c}{Struct. \& Unstruct.} \\

\cmidrule(lr){3-4} \cmidrule(lr){5-6} \cmidrule(lr){7-8} \cmidrule(lr){9-10} \cmidrule(lr){11-12} 
&& Score (\%) & Decr.(\%) & Score & Decr.(\%) & Score & Decr.(\%) & Score & Decr.(\%) & Score & Decr.(\%) \\ \midrule

\multirow{2}{*}{w/ data} & Acc & 62.52 & - & 60.13 & - & 58.04 & - & 67.42 & - & 62.10 & - \\
& F1 & 63.23 & - & 63.33 & - & 63.72 & - & 67.96 & - & 66.88 & - \\
& Prec. & 64.51 & - & 67.79 & - & 75.06 & - & 68.80 & - & 75.06 & - \\
& Recall & 62.52 & - & 60.13 & - & 58.04 & - & 67.42 & - & 62.10 & - \\
\midrule
\multirow{2}{*}{\parbox{3cm}{w/ shuffled data\\(rows)}} & Acc & 60.53 & (-1.99) & 57.56 & (-2.57) & 53.92 & (-4.12) & 65.67 & (-1.75) & 57.85 & (-4.25) \\
& F1 & 62.06 & (-1.17) & 61.37 & (-1.96) & 60.06 & (-3.66) & 66.19 & (-1.77) & 63.84 & (-3.04) \\
& Prec. & 65.09 & (+0.58) & 66.70 & (-1.09) & 74.36 & (-0.70) & 67.14 & (-1.66) & 75.63 & (+0.57) \\
& Recall & 60.53 & (-1.99) & 57.56 & (-2.57) & 53.92 & (-4.12) & 65.67 & (-1.75) & 57.85 & (-4.25) \\
\midrule
\multirow{2}{*}{\parbox{3cm}{w/ shuffled data\\(columns)}} & Acc & 58.30 & (-4.22) & 57.50 & (-2.63) & 54.31 & (-3.73) & 62.69 & (-4.73) & 58.98 & (-3.12) \\
& F1 & 59.82 & (-3.41) & 61.25 & (-2.08) & 60.69 & (-3.03) & 63.77 & (-4.19) & 64.42 & (-2.46) \\
& Prec. & 63.58 & (-0.93) & 66.51 & (-1.28) & 74.83 & (-0.23) & 65.41 & (-3.39) & 74.71 & (-0.35) \\
& Recall & 58.30 & (-4.22) & 57.50 & (-2.63) & 54.31 & (-3.73) & 62.69 & (-4.73) & 58.98 & (-3.12) \\
\midrule
\multirow{4}{*}{\parbox{3cm}{w/ shuffled data\\(rows and columns)}} 
& Acc & 54.78 & (-7.74) & 56.30 & (-3.83) & 53.86 & (-4.18) & 61.19 & (-6.23) & 57.25 & (-4.85) \\
& F1 & 57.01 & (-6.22) & 60.21 & (-3.12) & 60.18 & (-3.54) & 62.26 & (-5.70) & 63.14 & (-3.74) \\
& Prec. & 62.98 & (-1.53) & 66.07 & (-1.72) & 75.09 & (+0.03) & 64.33 & (-4.47) & 74.64 & (-0.42) \\
& Recall & 54.78 & (-7.74) & 56.30 & (-3.83) & 53.86 & (-4.18) & 61.19 & (-6.23) & 57.25 & (-4.85) \\
\midrule
\multirow{2}{*}{w/o data}  & Acc & 25.76 & (-36.76) & 49.97 & (-10.16) & 57.91 & (-0.13) & 41.79 & (-25.63) & 59.12 & (-2.98) \\
& F1 & 31.74 & (-31.49) & 53.60 & (-9.73) & 63.83 & (+0.11) & 47.15 & (-20.81) & 64.14 & (-2.74) \\
& Prec. & 53.56 & (-10.95) & 60.19 & (-7.60) & 72.95 & (-2.11) & 57.64 & (-11.16) & 71.45 & (-3.61) \\
& Recall & 25.76 & (-36.76) & 49.97 & (-10.16) & 57.91 & (-0.13) & 41.79 & (-25.63) & 59.12 & (-2.98) \\
\bottomrule
\end{tabular}
}
\caption{GPT-4o-mini's evidence resilience across different factual tasks under zero-shot settings without CoT prompts. The percentage of decrease with respect to the setting with original structured data (w/ data) is shown in brackets.}
\label{aptab:radar}
\end{table*}

\begin{figure*}[!t]
  \centering
  \begin{minipage}[t]{0.3\linewidth} %
    \subfigure[Llama3-8B Instruct]{
        \includegraphics[width=\linewidth]{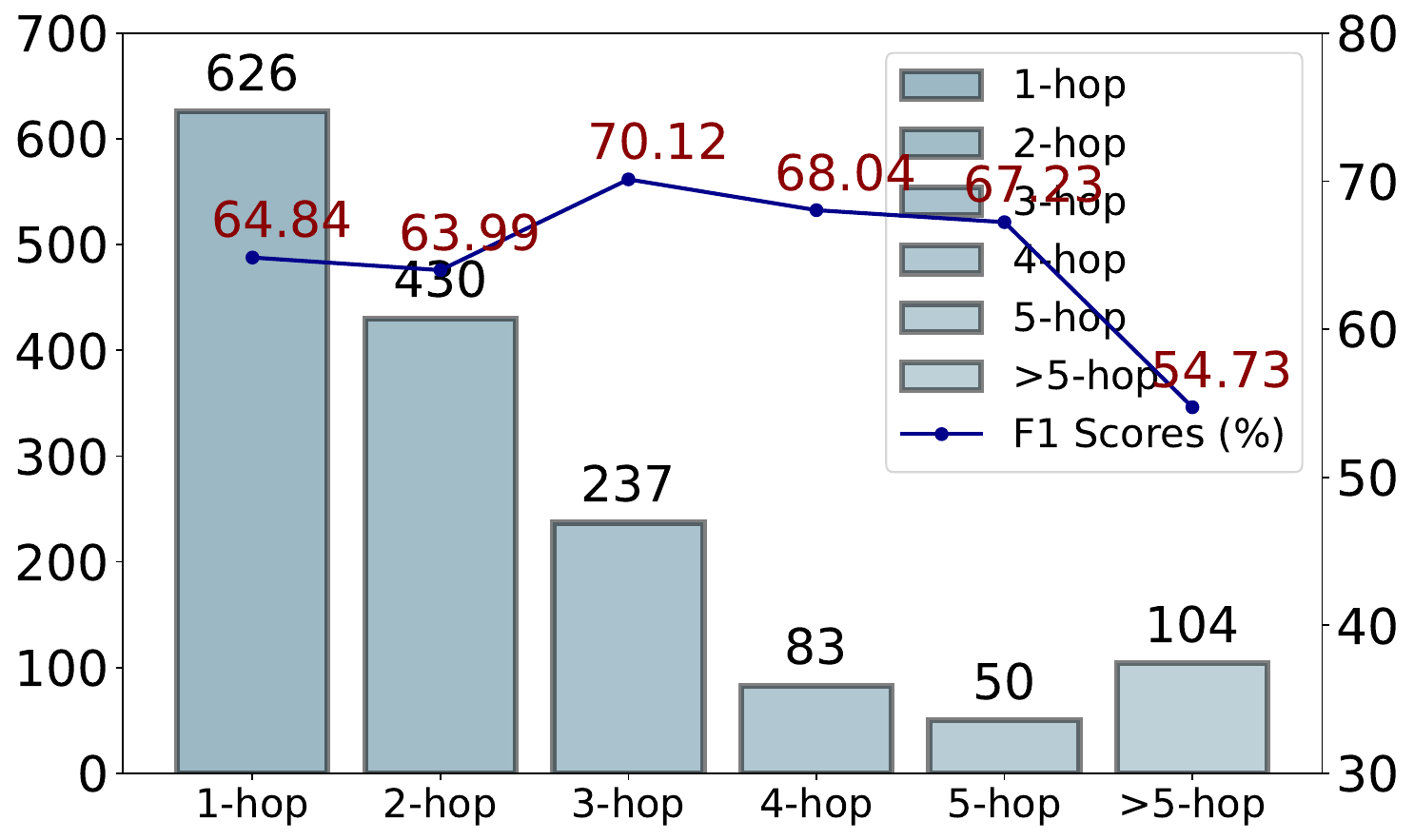}
    }
  \end{minipage}
  \begin{minipage}[t]{0.3\linewidth} %
    \subfigure[Gemma2-9B It]{
        \includegraphics[width=\linewidth]{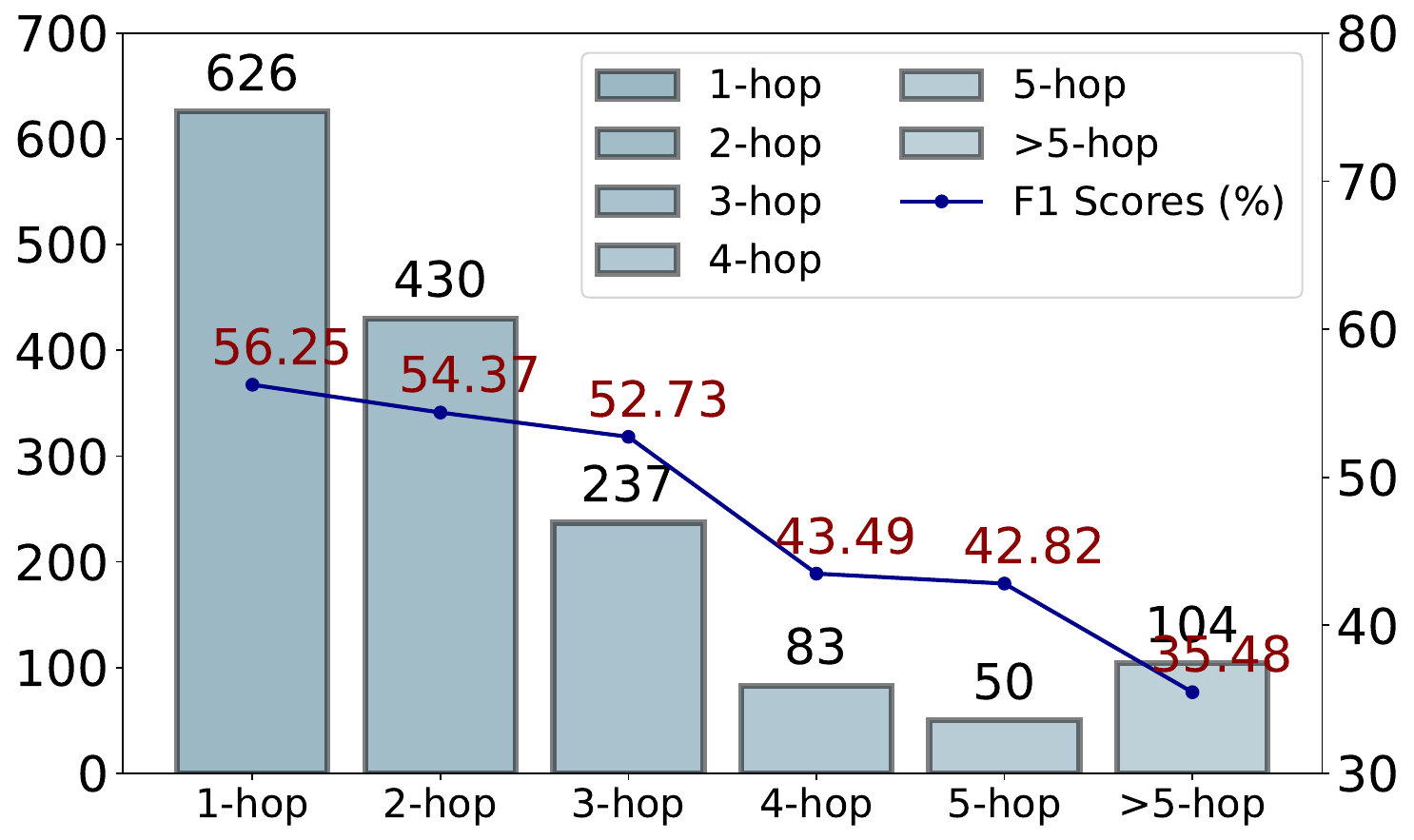}
    }  
  \end{minipage}
  \begin{minipage}[t]{0.3\linewidth} %
    \subfigure[Qwen2-7B-Instruct]{
        \includegraphics[width=\linewidth]{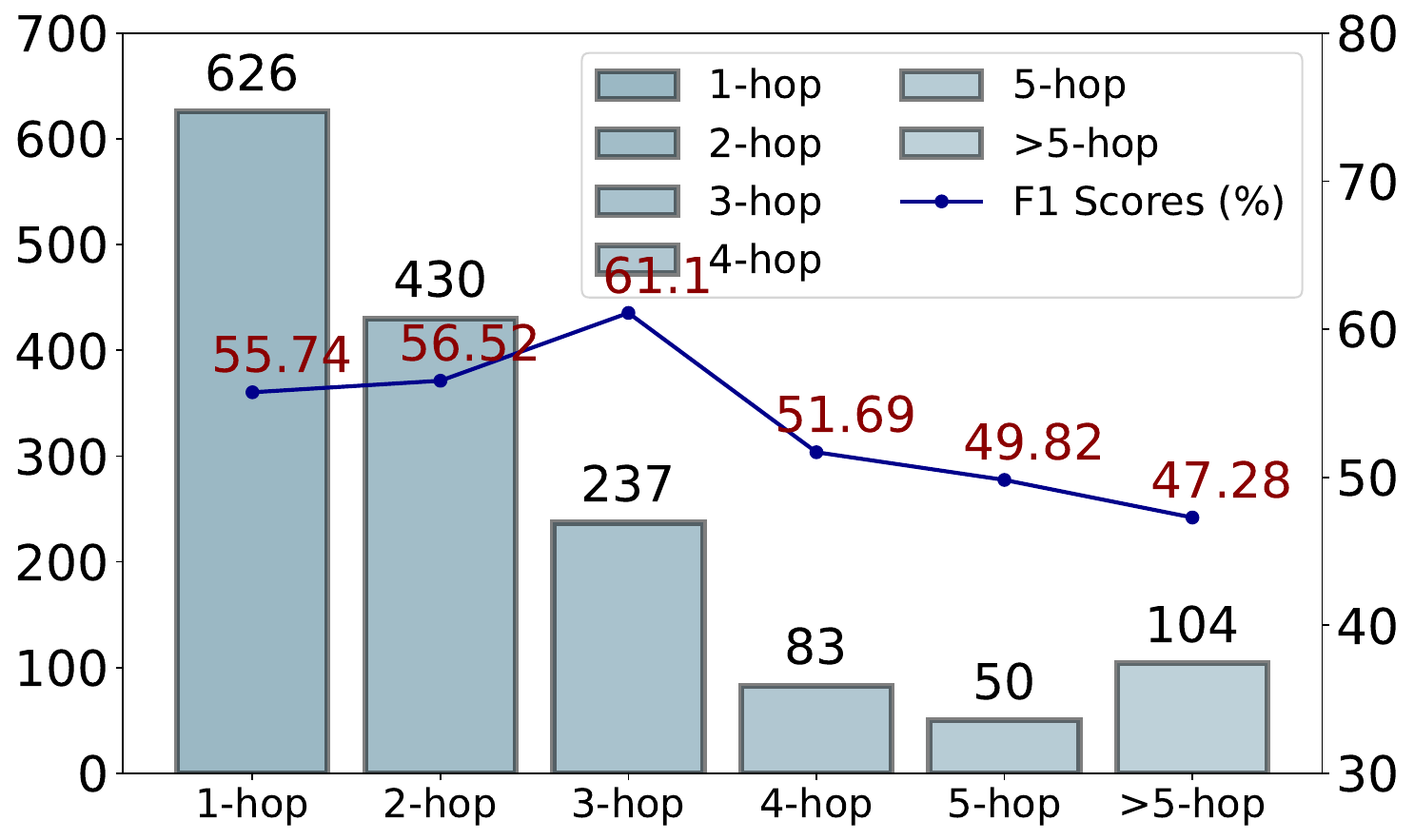}
    }  
  \end{minipage}
  \caption{Performance of Different Models on Multi-hop Reasoning.}
\end{figure*}

\begin{figure*}[!t]
  \centering
  \begin{minipage}[t]{0.3\linewidth} %
    \subfigure[Llama3-8B Instruct]{
        \includegraphics[width=\linewidth]{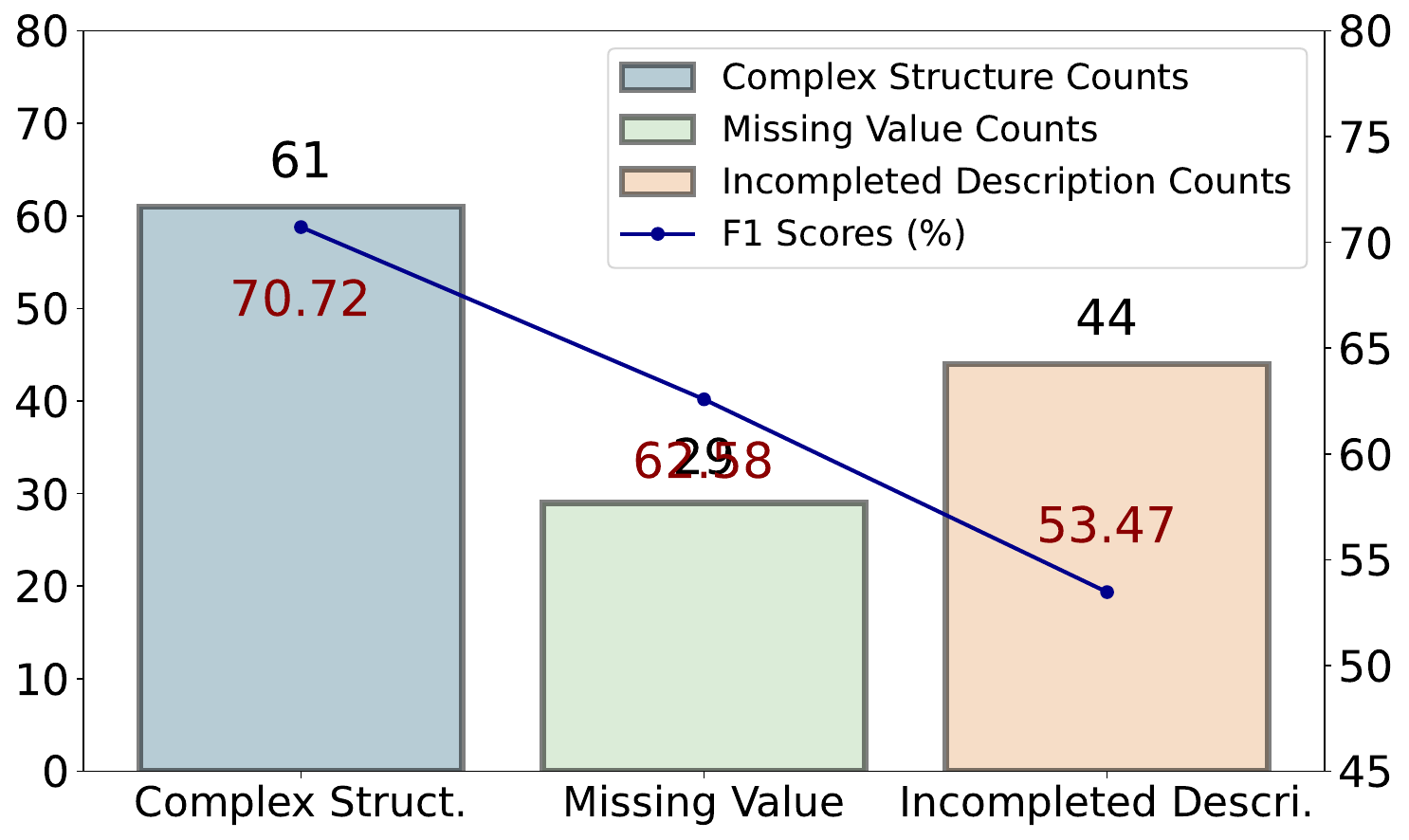}
    }
  \end{minipage}
  \begin{minipage}[t]{0.3\linewidth} %
    \subfigure[Gemma2-9B It]{
        \includegraphics[width=\linewidth]{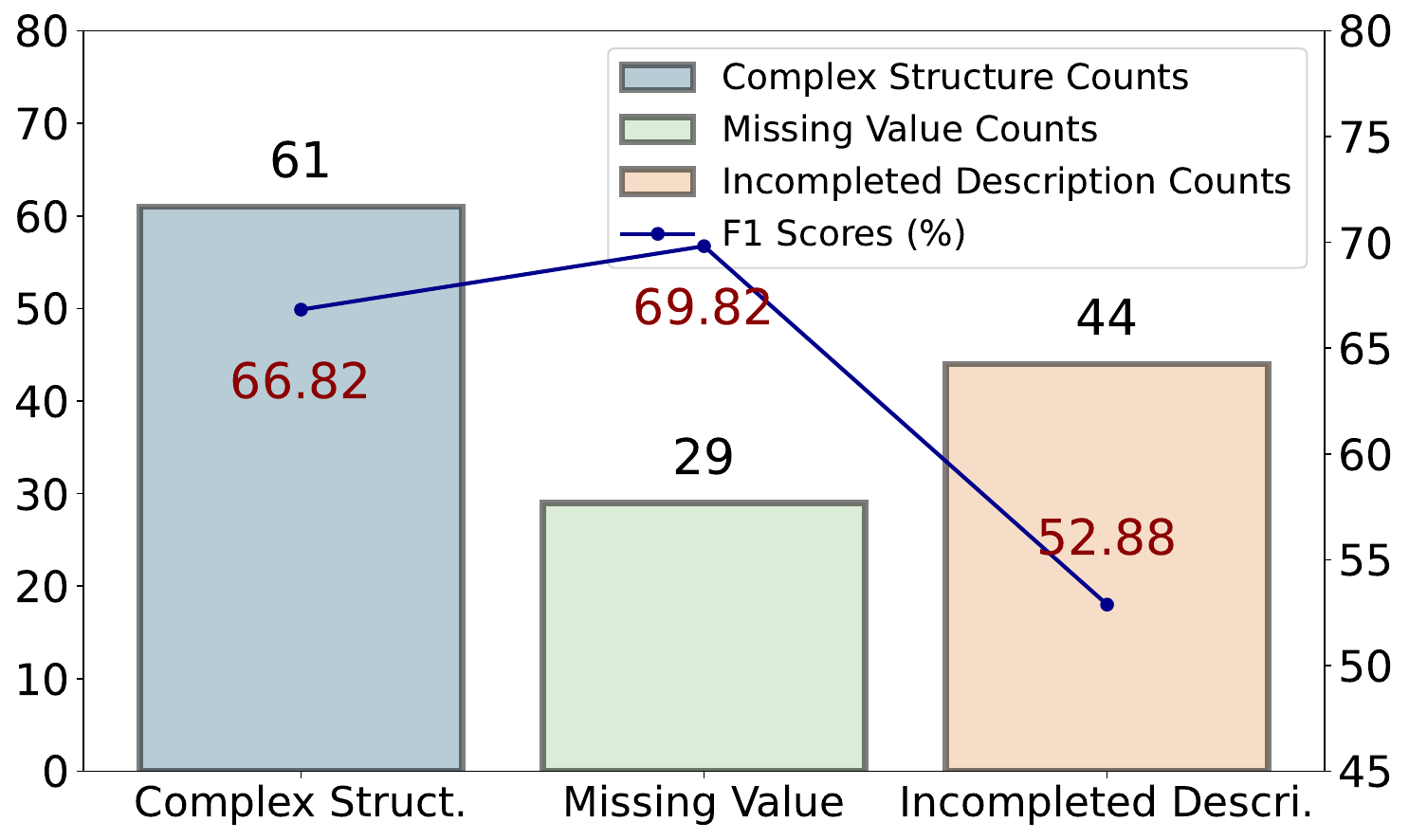}
    }  
  \end{minipage}
  \begin{minipage}[t]{0.3\linewidth} %
    \subfigure[Qwen2-7B-Instruct]{
        \includegraphics[width=\linewidth]{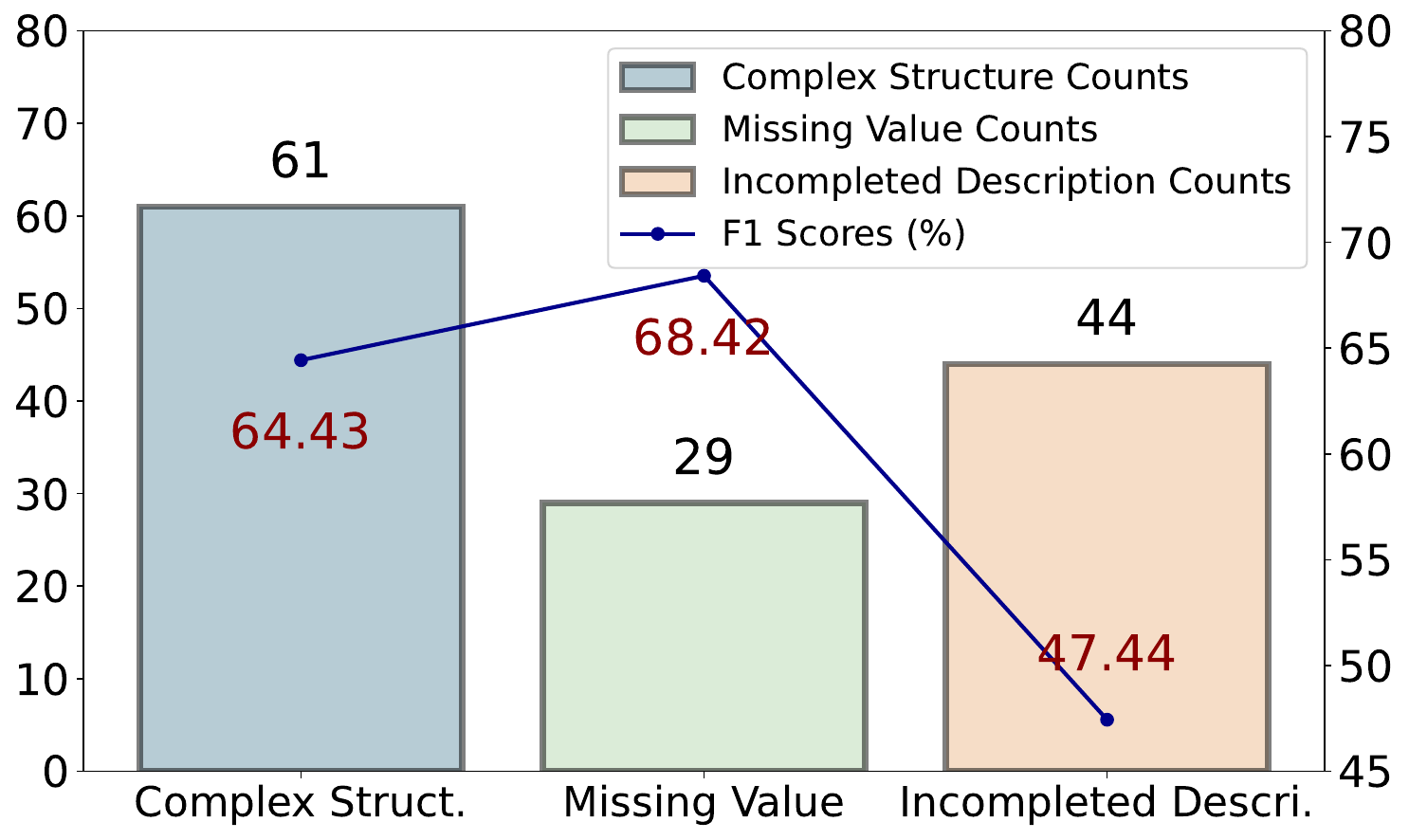}
    }  
  \end{minipage}
  \caption{Performance of Different Models on Composition Understanding.}
\end{figure*}

\end{document}